\documentclass{article}

\usepackage{iclr2027_conference,times}

\usepackage{amsmath,amsfonts,bm}

\def\eqref#1{equation~\ref{#1}}

\def\1{\bm{1}}

\DeclareMathAlphabet{\mathsfit}{\encodingdefault}{\sfdefault}{m}{sl}
\SetMathAlphabet{\mathsfit}{bold}{\encodingdefault}{\sfdefault}{bx}{n}

\usepackage{url}
\usepackage{graphicx}
\usepackage{booktabs}
\usepackage{multirow}
\usepackage{enumitem}
\usepackage[table]{xcolor}
\usepackage{colortbl}
\usepackage{wrapfig}
\usepackage{float}
\usepackage{framed}
\usepackage{fvextra}

\usepackage[pagebackref=true,breaklinks=true,colorlinks=true,bookmarks=false]{hyperref}
\definecolor{deepred}{HTML}{940000}
\hypersetup{linkcolor=deepred}

\definecolor{linkpurple}{rgb}{0.40,0.15,0.95}
\definecolor{lightblue}{rgb}{0.30,0.60,1.00}

\colorlet{citeblue}{linkpurple!60!lightblue} 
\colorlet{citemuted}{citeblue!90!gray} 

\hypersetup{
  citecolor = citemuted,
  urlcolor  = citemuted,
}

\definecolor{promptborder}{RGB}{176,190,210}
\definecolor{promptbackground}{RGB}{247,249,252}
\definecolor{algorithmborder}{RGB}{142,166,194}
\definecolor{algorithmbackground}{RGB}{245,248,252}
\definecolor{tierzerohead}{RGB}{229,238,247}
\definecolor{tierzerobody}{RGB}{248,250,252}
\definecolor{tieronehead}{RGB}{238,233,246}
\definecolor{tieronebody}{RGB}{251,249,252}
\definecolor{tiertwohead}{RGB}{231,242,235}
\definecolor{tiertwobody}{RGB}{249,252,250}
\definecolor{tierthreehead}{RGB}{248,238,224}
\definecolor{tierthreebody}{RGB}{253,251,247}
\definecolor{gptcasegreen}{RGB}{55,112,25}
\definecolor{geminicaseblue}{RGB}{0,94,168}
\definecolor{opuscaseorange}{RGB}{220,105,0}
\definecolor{astracasepurple}{RGB}{92,38,150}
\definecolor{seedcasepurple}{RGB}{92,38,150}
\definecolor{piacasegray}{RGB}{85,85,85}
\definecolor{deltapos}{RGB}{76,175,80}
\definecolor{deltaneg}{RGB}{220,80,80}
\definecolor{t2pblue}{RGB}{145,190,220}
\definecolor{t2porange}{RGB}{244,192,143}
\definecolor{citationblue}{RGB}{25,65,115}
\definecolor{crossrefred}{RGB}{145,35,45}
\definecolor{conceptanimal}{RGB}{232,241,249}
\definecolor{conceptfood}{RGB}{252,239,226}
\definecolor{concepthousehold}{RGB}{241,236,248}
\definecolor{concepttransport}{RGB}{232,244,235}
\definecolor{conceptnature}{RGB}{249,245,220}
\newenvironment{promptbox}{%
    \def\FrameCommand##1{{%
        \setlength{\fboxsep}{5pt}%
        \fcolorbox{promptborder}{promptbackground}{##1}}}%
    \MakeFramed{\advance\hsize-\width\FrameRestore}%
    \vspace{-4pt}
}{%
    \endMakeFramed
}
\newenvironment{algorithmbox}[1]{%
    \def\FrameCommand##1{{%
        \setlength{\fboxsep}{6pt}%
        \fcolorbox{algorithmborder}{algorithmbackground}{##1}}}%
    \MakeFramed{\advance\hsize-\width\FrameRestore}%
    \small\noindent\textbf{#1}\par\vspace{2pt}
}{%
    \endMakeFramed
    \vspace{2pt}
}

\usepackage[toc,page,header]{appendix}
\usepackage{minitoc}

\renewcommand \thepart{}
\renewcommand \partname{}

\title{PixelBench: Can LLMs Understand and\\Generate Images Pixel by Pixel?}
\newcommand{\authoritemsep}{\hspace{0.8em}}
\newcommand{\authornamesep}{\hspace{0.85em}}
\newcommand{\subsetmark}{\raisebox{0.45ex}[0pt][0pt]{\scriptsize *}}
\newcommand{\animalconcept}[1]{\cellcolor{conceptanimal}\strut#1}
\newcommand{\foodconcept}[1]{\cellcolor{conceptfood}\strut#1}
\newcommand{\householdconcept}[1]{\cellcolor{concepthousehold}\strut#1}
\newcommand{\transportconcept}[1]{\cellcolor{concepttransport}\strut#1}
\newcommand{\natureconcept}[1]{\cellcolor{conceptnature}\strut#1}
\newcommand{\scoreblue}[2]{\cellcolor{t2pblue!#1}\strut #2}

\newcommand{\scoreborder}[1]{\cellcolor{white}\strut #1}

\newcommand{\tablecaptionfont}{\scriptsize}
\makeatletter
\renewcommand{\fnum@figure}{\footnotesize\textbf{\figurename~\thefigure}}
\renewcommand{\fnum@table}{\tablecaptionfont\textbf{\tablename~\thetable}}
\makeatother
\newcommand{\setpositionedtablecaption}[2]{%
    \def\storedtablecaption{#1}%
    \def\storedtablelabel{#2}%
    \caption{\storedtablecaption}%
    \label{\storedtablelabel}%
}
\newcommand{\placepositionedtablecaptionbottom}{}
\AtBeginEnvironment{table}{\setlength{\belowcaptionskip}{5pt}}
\AtBeginEnvironment{wraptable}{\setlength{\belowcaptionskip}{5pt}}

\author{\small
Haoquan Zhang$^{1}$\authornamesep
Feiyang Liu$^{2}$\authornamesep
Juanxi Tian$^{3}$\authornamesep
Wanhua Li$^{3}$\authornamesep
Haodong Duan$^{1,\dagger}$\authornamesep
Weiyang Liu$^{1}$
\\[0.25em]
{\normalfont\small $^{1}$CUHK\authoritemsep
$^{2}$SCUT\authoritemsep
$^{3}$NTU Singapore\authoritemsep
$^{\dagger}$Corresponding author\authoritemsep\quad
\href{https://haoquanzhang.github.io/pixelbench/}{\textbf{Project Page}}}
\vspace{-5mm}
}

\iclrfinalcopy

\begin{document}

\maketitle
\doparttoc
\faketableofcontents

\lhead{~\raisebox{-0.47ex}{\includegraphics[height=.38cm]{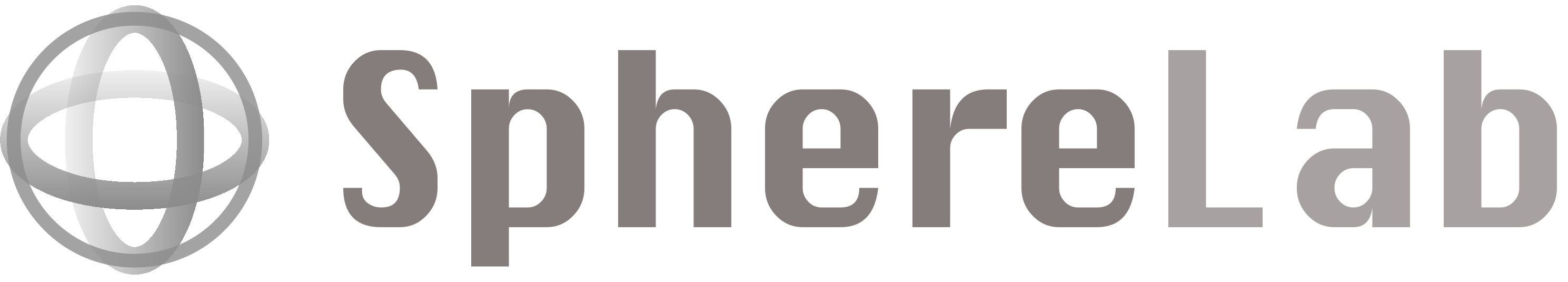}}~~~Technical Report}  

\begin{figure}[h]
\centering
\includegraphics[width=\linewidth]{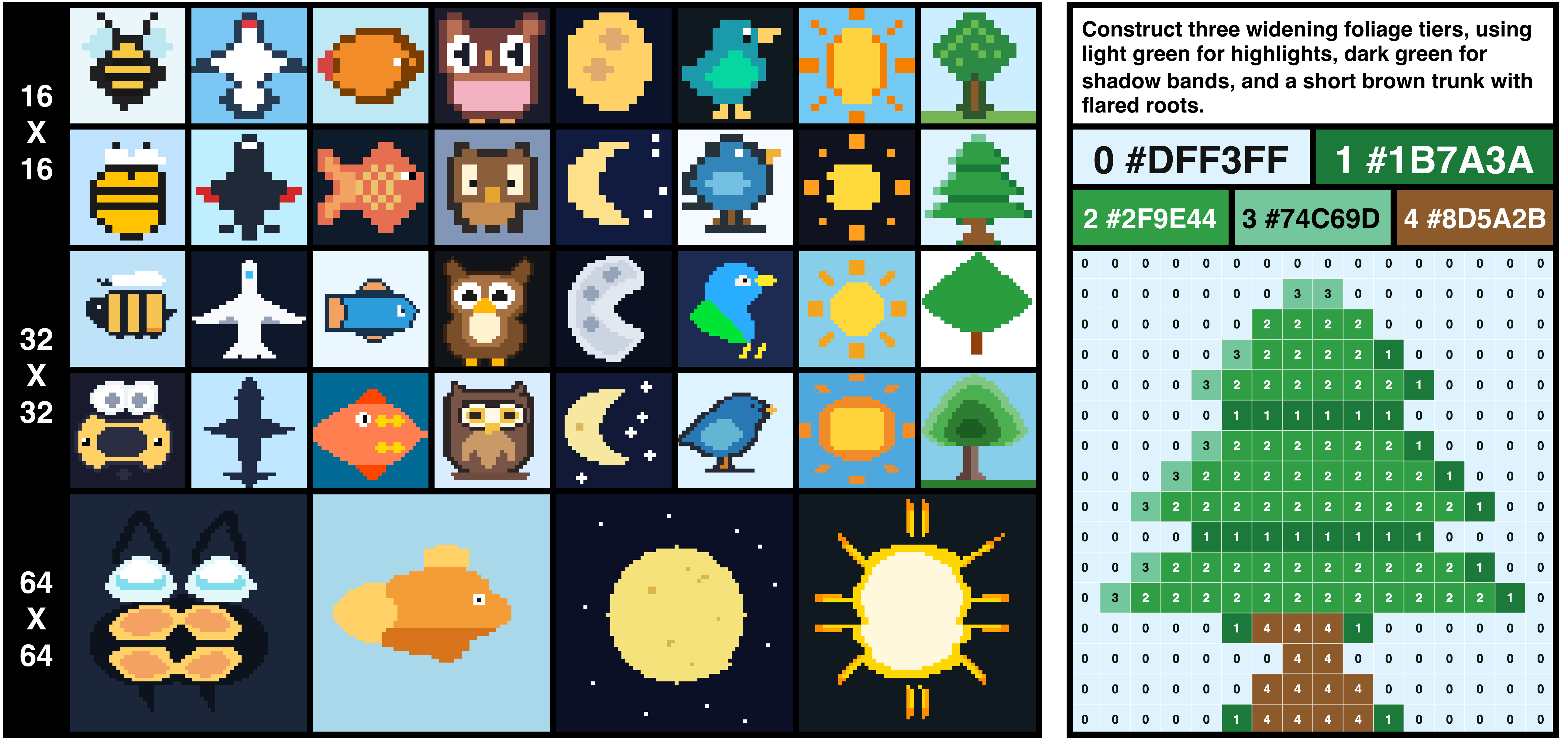}
\vspace{-7mm}
\caption{\footnotesize\textbf{Zero-shot text-to-pixel generation by large
language models.} Left: pixel images generated at different resolutions.
Right: our generation protocol, in which the model defines a color palette and
outputs a numerical grid whose entries index the palette colors. See
Figure~\ref{fig:i2p-case-studies} for image-to-pixel generation.}
\label{fig:teaser}
\vspace{1.5mm}
\end{figure}

\begin{abstract}
\vspace{-1mm}
    Large language models (LLMs) can generate visual assets through text-based formats such as SVG and CAD, but these symbolic representations describe images using high-level primitives rather than explicit raster values. We ask whether LLMs can instead generate pixel images directly through their existing vocabulary, without a learned image tokenizer or decoder. We introduce PixelBench, a benchmark for text-to-pixel (T2P) and image-to-pixel (I2P) generation across multiple resolutions, together with a complementary pixel-understanding evaluation. Models generate an RGB color palette followed by a grid of indices specifying the palette color of each output pixel. In T2P, the strongest models produce recognizable images directly from text. In I2P, the strongest models outperform palette-matched downsampling and specialized pixelization algorithms at low resolutions. Pixel-understanding performance correlates strongly with generation performance. A resolution-scaling study shows that pixel-generation performance often improves initially before declining at high resolutions. Finally, we introduce Pixel Reconstruction Tuning (PRT), which augments visual instruction tuning with caption-palette-grid trajectories. We post-train Molmo2-4B from its pretraining checkpoint by mixing general instruction data with our curated PRT data, improving performance across 16 visual-centric benchmarks, indicating that explicit pixel reconstruction can provide useful auxiliary supervision.
\end{abstract}

\vspace{-1mm}
\section{Introduction}
\vspace{-1mm}

Large language models (LLMs) were originally developed to predict the next token in
natural-language sequences. The same autoregressive training paradigm has
since produced strong code-generation capabilities. Some generated code,
including SVG code and CAD programs, can be rendered into images, providing an
interface for probing how LLMs express visual information
\citep{sharma2024vision,qiu2024symbolic,chen2025symbolicgraphics,rodriguez2025starvector,xing2025llm4svg,khan2024text2cad}.
However, these formats describe images using semantically meaningful,
high-level primitives and are therefore one step away from actual visual
appearance: a renderer expands each primitive into many pixels. By contrast,
RGB rasters provide a more direct and ubiquitous representation of digital
images, with each spatial location in a two-dimensional grid storing a color
value. Most modern image generators map learned visual tokens or latent
representations to pixels through dedicated decoders
\citep{chameleon2024,wang2024emu3}. {Prior work has also trained specialized
autoregressive image models to generate images directly in raster space
\citep{oord2016pixelrnn,oord2016pixelcnn,parmar2018image,chen2020igpt}.
}
These studies demonstrate the feasibility of direct pixel generation, but rely on models specifically designed and trained for image modeling. To better investigate pixel-based visual understanding and generation, we ask the following question:

\begin{center}
\vspace{-2pt}
\setlength{\fboxsep}{5.5pt}
\setlength{\fboxrule}{0.5pt}
\fcolorbox{promptborder}{promptbackground}{%
\parbox{\dimexpr\linewidth-2\fboxsep-2\fboxrule\relax}{\centering\small
Can LLMs directly generate and understand visual content pixel by pixel?}}
\end{center}
\vspace{-2pt}

To address this research question, we introduce \textbf{PixelBench}, a
benchmark for evaluating LLMs on direct pixel generation and understanding.
PixelBench uses a
palette-indexed raster representation: the model first defines a palette of up
to eight RGB colors and then generates a grid of palette indices, with each
grid position specifying the color of one pixel. This interface needs no
learned tokenizer, image decoder, or architectural change.
PixelBench evaluates three tasks: \emph{text-to-pixel generation} (T2P), where a model generates a recognizable pixel image from an object label, \emph{image-to-pixel generation} (I2P), where the model reconstructs a given image in pixel format, and \emph{pixel understanding}, where the model identifies the underlying concept from a textual palette-and-grid representation generated by either an algorithmic or a model-based pixel oracle.

Our evaluation shows that frontier LLMs already exhibit substantial pixel-generation capabilities, with leading models producing recognizable T2P outputs and competitive I2P reconstructions.
Their behavior, however, depends strongly on target resolution. For many models, both visual quality and quantitative performance peak at an intermediate resolution, with $64\times64$ often performing worse than $32\times32$, while GPT-6 Astra exhibits notably better robustness.
We therefore conduct a fine-grained resolution-scaling study and find that, except for Astra's consistently strong performance, most evaluated models initially improve but eventually decline as resolution increases.
Viewing resolution as a measure of visual abstraction, these model-specific performance peaks suggest that different models are most effective at different levels of abstraction. Pixel understanding exhibits a similar ranking across models and correlates strongly with generation performance, suggesting that the connection between visual understanding and generation persists for pixel-based interface.

The observed connection between pixel generation and visual understanding raises a natural question of whether reconstructing an input image can serve as a training objective to improve a model's visual understanding.
Reconstruction Alignment (RecA)~\citep{xie2025reconstruction} shows that reconstruction-aligned visual
representations can improve image generation, but whether reconstruction can also enhance visual understanding remains underexplored.
Motivated by this, we propose Pixel Reconstruction Tuning (PRT), which augments post-training with trajectories that describe an image, explain an automatically constructed color palette, and reconstruct the image as a pixel grid.
In a controlled experiment with Molmo2-4B, we start from its pretraining checkpoint and post-train it on 500k LLaVA-OneVision-1.5 instructions mixed with PRT examples. Incorporating PRT data improves average performance across 16 visual-centric benchmarks. Ablations varying the number of PRT examples from 1k to 8k show consistent improvements across all tested data scales.
A complementary study with Qwen3-VL further shows that Pixel data can regularize post-training when incorporated into a general instruction mixture.
Our contributions are summarized below:
\begin{itemize}[leftmargin=1.2em,labelsep=0.4em]
    \item {\textbf{A unified interface for evaluating pixel capabilities.} We introduce PixelBench, a shared text-based pixel interface for evaluating frontier LLMs on both pixel generation and understanding.}
    \item {\textbf{Capability limits at increasing
    resolutions.} We characterize model-specific performance peaks and failure modes. GPT-6 Astra remains robust as resolution increases, while other evaluated models generally improve and then decline.}
    \item {\textbf{Pixels as training supervision.} We propose Pixel Reconstruction Tuning (PRT), which uses pixel reconstruction as a post-training objective. Starting from the Molmo2-4B pretraining checkpoint, incorporating PRT data into general instruction tuning consistently improves performance.}
\end{itemize}

\begin{figure}[t]
    \centering
    \includegraphics[width=\linewidth]{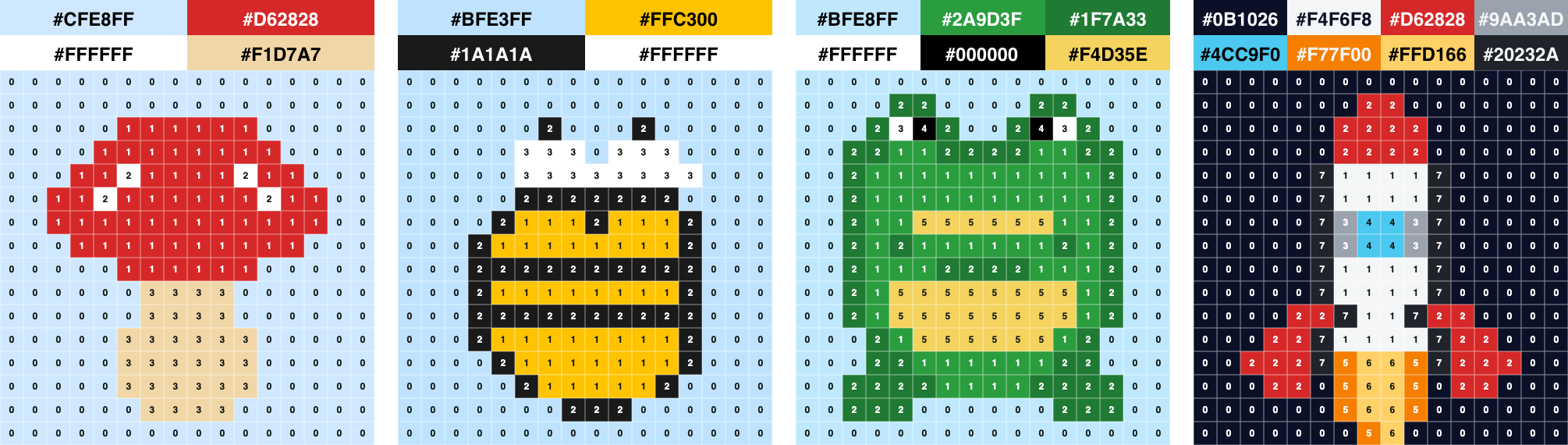}
    \vspace{-6mm}
    \caption{\footnotesize\textbf{Examples of palette-indexed pixel representations.}
    Each example pairs a compact color palette (top) with a $16\times16$ grid
    of palette indices (bottom). The mushroom, bee, frog, and rocket remain
    recognizable despite the low resolution and restricted palettes.}
    \label{fig:pixel-representation}
    \vspace{-1mm}
\end{figure}

\vspace{-1mm}
\section{\underline{PixelBench}: Evaluating Pixel Understanding and Generation}
\label{sec:pixelbench}
\vspace{-.75mm}

\subsection{Pixels as a Visual Representation}
\label{sec:pixel-representation}
\vspace{-.5mm}

Pixel art is a compact visual output format. Its restricted palette and raster
structure can preserve recognizable colors and shapes while limiting the
number of symbols that a model must generate. Related representations are
commonly used in icons, pictograms, low-resolution displays, and digital
graphics. In this work, \emph{pixel art} refers specifically to a
palette-indexed raster with two components: a palette of RGB colors and an
$S\times S$ grid of palette indices. Each grid entry selects the color of one
pixel. The representation is human-interpretable and deterministic to render,
and can be expressed using the model's existing text vocabulary. We serialize
the palette and index grid as JSON. The \texttt{palette} field
contains two to eight opaque RGB colors in \texttt{\#RRGGBB} format, and the
\texttt{pixels} field contains the $S$ rows of palette indices. Rendering maps
each index to its corresponding RGB value. Figure~\ref{fig:pixel-representation}
shows rendered examples and their palettes.

\vspace{-.5mm}
\subsection{Benchmark Introduction and Design}
\label{sec:pixelbench-design}
\vspace{-.5mm}

We introduce PixelBench to probe how much image-generation capability a
language model can express through its native token interface. Unlike SVG and
CAD, which describe images through high-level geometric primitives, our
palette-indexed grids specify raster values directly: each generated grid entry
corresponds to one rendered pixel, and rendering requires only a deterministic
palette lookup. PixelBench therefore evaluates visual generation through a
representation closer to the native form of raster images, without relying on
a learned image tokenizer or decoder.

Section~\ref{sec:pixelbench-design} defines the generation and understanding
tasks, models, baselines, and protocol; Section~\ref{sec:pixelbench-metrics}
specifies the metrics; and Section~\ref{sec:pixelbench-results} presents the main results.
We find that current models already exhibit substantial
pixel-image generation ability through their text interface, while their
pixel-understanding performance closely tracks their generation performance.
Generation quality does not consistently improve with output resolution, as
further analyzed in Section~\ref{sec:resolution}.

\textbf{Tasks and datasets.}
\phantomsection\label{sec:t2p}
In \emph{text-to-pixel generation} (T2P), a model maps an English object label
to a recognizable icon without a reference image. We use 100 concrete concepts
spanning five categories (Appendix~\ref{app:t2p-concepts}).
\phantomsection\label{sec:i2p}
In \emph{image-to-pixel generation} (I2P), a vision encoder supplies a natural
image and the model reconstructs it while preserving subject identity, pose,
composition, and dominant colors. We sample one validation image from each of
100 ImageNet-1K classes \citep{deng2009imagenet} and center-crop it to
$256\times256$ (Appendix~\ref{app:i2p-sources}); the category is withheld.
Every item in both generation tasks is evaluated at
$16^2$, $32^2$, and $64^2$, yielding 300 instances per task and model.
\phantomsection\label{sec:pixel-understanding}
In \emph{pixel understanding}, a model receives only the textual serialization
of an RGB palette and index grid and selects the concept from ten candidates.
Inputs come from deterministic PIA-reduced reconstructions and GPT-6 Astra I2P
generations (Astra-I2P), each with 100 questions at each resolution.

\textbf{Models and baselines.}
We evaluate Gemini~3.6 Flash~\citep{google2026gemini36flash},
GPT-5.6 Sol~\citep{openai2026gpt56},
GPT-6 Astra~\citep{openai2026gpt6astra},
Claude Opus~5~\citep{anthropic2026claudeopus5},
Seed-2.1 Pro~\citep{bytedance2026seed21}, and
Kimi~K3~\citep{kimiteam2026k3} on both generation tasks and pixel understanding
in a zero-shot setting.
DeepSeek~V4 Pro~\citep{deepseekai2026v4} is included as a supplementary model
for T2P and pixel understanding, but not for I2P generation.
Detailed task-specific inference settings, including temperature, reasoning
configuration, output-token limits, and samples per item, are reported in
Appendix~\ref{app:inference-settings}.
For I2P, we additionally compare against three representative palette-matched
algorithms: nearest-neighbor sampling followed by median-cut quantization
\citep{heckbert1982color}; SSIM-optimal downscaling followed by median cut
\citep{oztireli2015perceptual}; and a reduced Pixelated Image Abstraction
pipeline \citep{gerstner2012pixelated} that combines SLIC regions
\citep{achanta2012slic} with an eight-color weighted $k$-means palette. The
complete ten-method comparison and implementation details are in
Appendix~\ref{app:baselines}.

\textbf{Generation protocol.}
Both generation tasks use the palette-indexed representation in
Figure~\ref{fig:pixel-representation}: two to eight opaque RGB colors and an
exactly sized index grid, with task-specific prompts. Successfully returned
outputs with invalid JSON, palette references, or grid sizes receive zero;
rate limits and token-limit truncations are retried, raising the limit when
needed. Pixel understanding directly presents the serialized grid and ten
choices, with no rendered image. Prompts are in
Appendix~\ref{app:full-prompts}.

\vspace{-.5mm}
\subsection{Evaluation Metrics}
\label{sec:pixelbench-metrics}
\vspace{-.5mm}

\textbf{Text-to-Pixel metrics.}
Using the paired SigLIP-SO400M text and vision encoders
\citep{zhai2023sigmoid}, we retrieve text candidates represented as ``a pixel
art icon depicting \{label\}.'' The main table reports 500-way Top-1 and a
complementary VLM-as-judge score from Gemini~3.5
Flash~\citep{google2026gemini35flash}, whose 0-100 rubric prompt is in
Appendix~\ref{app:t2p-judge-prompt}. The 100-way Top-1 and MRR
diagnostics are in Appendix~\ref{app:t2p-full-results}; retrieval definitions
and vocabulary are in Appendices~\ref{app:siglip-retrieval}
and~\ref{app:t2p-500way}.

\textbf{Image-to-Pixel metrics.}
We report \emph{100-way Top-1} by embedding each generated output and the 100 candidate source images with the same SigLIP vision encoder. We additionally compute paired cosine similarity using pooled SigLIP features and DINOv2-Large CLS features \citep{oquab2024dinov2}, which measure semantic and appearance-level similarity to the source, respectively.
For both T2P and I2P, invalid outputs count as retrieval misses or zero
similarity. We also report the Valid and Flat rates. Valid is the percentage
of outputs with well-formed JSON, valid palette references, and the required
grid size; Flat is the percentage of outputs that are valid but render to a
single color.

\textbf{Pixel-understanding metric.}
We report 10-way recognition accuracy from the serialized palette-and-grid
inputs. Each resolution contains 100 questions, so each source average covers
300 questions and the overall average covers all 600. Random-choice accuracy
is 10\%.

\vspace{-.5mm}
\subsection{Experimental Results}
\label{sec:pixelbench-results}
\vspace{-.5mm}

\begin{wraptable}[12]{r}{0.63\linewidth}
\vspace{-34pt}
\centering
\begin{minipage}[t]{\linewidth}
\caption{\tablecaptionfont{\textbf{Text-to-Pixel generation results.}
We report 500-way Top-1 retrieval accuracy and a VLM-as-judge score at $16^2$,
$32^2$, and $64^2$. The Aggregated score is the mean of these six values. Valid and Flat report
output validity and single-color generation rates. Gemini~3.5 Flash serves as
the judge; 100-way Top-1 and MRR are in the appendix. Bold and underline mark
the top two results per column. Agg.\ abbreviates Aggregated score.}}
\label{tab:t2p-main-results}
\centering
\scriptsize
\setlength{\tabcolsep}{2.2pt}
\renewcommand{\arraystretch}{0.9}
\resizebox{\linewidth}{!}{%
\begin{tabular}{@{}l
!{\color{white}\vrule width 1.2pt}c
!{\color{white}\vrule width 1.2pt}c
!{\color{white}\vrule width 1.2pt}c
!{\color{white}\vrule width 1.2pt}c
!{\color{white}\vrule width 1.2pt}c
!{\color{white}\vrule width 1.2pt}c
!{\color{white}\vrule width 1.2pt}c
!{\color{white}\vrule width 1.2pt}c
!{\color{white}\vrule width 1.2pt}c
!{\color{white}\vrule width 1.2pt}c
!{\color{white}\vrule width 1.2pt}c@{\hspace{\tabcolsep}}}
\toprule
& & & \multicolumn{4}{c}{500-way Top-1} &
\multicolumn{4}{c}{{VLM Judge}} & \\
\cmidrule(lr){4-7}\cmidrule(lr){8-11}
Model & Valid & Flat &
$16$ & $32$ & $64$ & Avg &
{$16$} & {$32$} &
{$64$} & {Avg} & Agg. \\
\midrule
GPT-6 Astra & 100.0 & 0.0 &
\scoreblue{22}{\textbf{87}} & \scoreblue{33}{\textbf{92}} &
\scoreblue{39}{\textbf{94}} & \scoreblue{31}{\textbf{91.0}} &
\scoreblue{30}{\textbf{90.8}} & \scoreblue{32}{\textbf{91.6}} &
\scoreblue{28}{\textbf{89.7}} & \scoreblue{30}{\textbf{90.7}} &
\scoreblue{31}{\textbf{90.8}} \\
Claude Opus 5 & 100.0 & 0.0 &
\scoreblue{12}{81} & \scoreblue{22}{87} &
\scoreblue{26}{\underline{89}} & \scoreblue{20}{\underline{85.7}} &
\scoreblue{5}{75.1} & \scoreblue{4}{74.1} &
\scoreblue{1}{70.1} & \scoreblue{3}{73.1} &
\scoreblue{10}{79.4} \\
GPT-5.6 Sol & 100.0 & 0.0 &
\scoreblue{17}{\underline{84}} & \scoreblue{26}{\underline{89}} &
\scoreblue{15}{83} & \scoreblue{19}{85.3} &
\scoreblue{8}{\underline{78.1}} & \scoreblue{9}{\underline{78.9}} &
\scoreblue{3}{\underline{73.0}} & \scoreblue{6}{\underline{76.7}} &
\scoreblue{12}{\underline{81.0}} \\
Kimi K3 & 99.7 & 0.0 &
\scoreblue{12}{81} & \scoreblue{22}{87} &
\scoreblue{7}{77} & \scoreblue{13}{81.7} &
\scoreblue{4}{74.5} & \scoreblue{4}{74.7} &
\scoreborder{61.5} & \scoreblue{1}{70.3} &
\scoreblue{6}{76.0} \\
Gemini 3.6 Flash & 100.0 & 1.3 &
\scoreblue{22}{\textbf{87}} & \scoreblue{8}{78} &
\scoreblue{2}{72} & \scoreblue{9}{79.0} &
\scoreblue{8}{77.8} & \scoreblue{2}{71.2} &
\scoreborder{60.4} & \scoreblue{1}{69.8} &
\scoreblue{4}{74.4} \\
Seed-2.1 Pro & 100.0 & 0.0 &
\scoreblue{1}{70} & \scoreblue{2}{72} &
\scoreblue{4}{74} & \scoreblue{2}{72.0} &
\scoreblue{1}{68.6} & \scoreborder{64.6} &
\scoreborder{59.3} & \scoreborder{64.2} &
\scoreblue{1}{68.1} \\
DeepSeek V4 Pro & 98.7 & 1.3 &
\scoreborder{49} & \scoreborder{59} &
\scoreborder{51} & \scoreborder{53.0} &
\scoreborder{53.7} & \scoreborder{55.0} &
\scoreborder{45.7} & \scoreborder{51.5} &
\scoreborder{52.2} \\
\bottomrule
\end{tabular}%
}
\end{minipage}
\vspace{-4pt}
\end{wraptable}

\begin{figure}[t]
\centering
\vspace{-1mm}
\includegraphics[width=\linewidth]{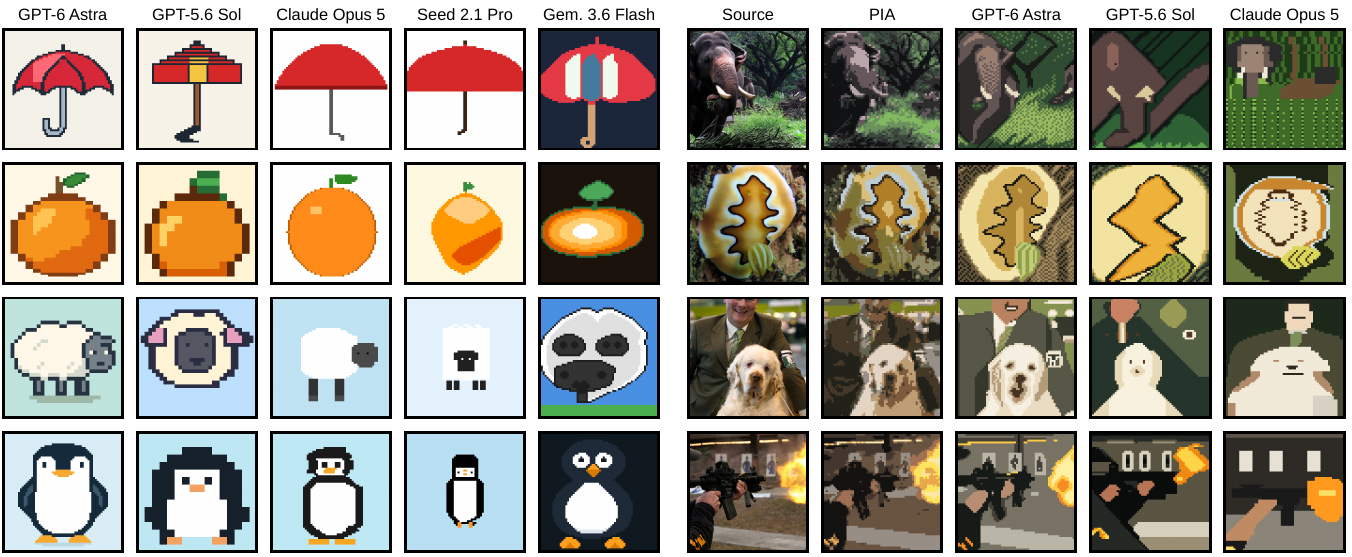}
\vspace{-6mm}
\caption{\footnotesize\textbf{Qualitative PixelBench generations.}
\textbf{Left:} T2P outputs from GPT-6 Astra, GPT-5.6 Sol, Claude Opus 5,
Seed-2.1 Pro, and Gemini 3.6 Flash. \textbf{Right:} four I2P sources and
reconstructions from PIA, Astra, Sol, and Claude. Some models preserve subjects
more clearly at low resolutions, whereas PIA better tracks local appearance at
higher ones. PIA is the reduced variant (Appendix~\ref{app:baselines});
Appendix~\ref{app:qualitative-cases} adds 20 cases per task.}
\label{fig:i2p-case-studies}
\vspace{1.5mm}
\end{figure}

\textbf{Text-to-Pixel Generation.}
Table~\ref{tab:t2p-main-results} summarizes the main results for T2P
generation.  GPT-6 Astra leads throughout: its 500-way Top-1 rises
from 87\% to 94\% across resolutions and its Agg.\ score reaches 90.8.
GPT-5.6 Sol and Claude Opus 5
follow at 81.0 and 79.4, with Kimi K3 at 76.0. The stricter judge spans
90.66-51.45 overall and leaves Astra 14.01 points ahead of Sol, providing
clearer discrimination while preserving the same leader.
Figure~\ref{fig:i2p-case-studies} and
Appendix~\ref{app:qualitative-cases} further show that model-generated
pixelizations can preserve semantic structure; together with Astra's leading
I2P Aggregated score, this motivates Astra-I2P as one understanding oracle.
Full 100-way Top-1 and MRR results are in
Appendix~\ref{app:t2p-full-results}. Most models decline from $32^2$ to $64^2$,
motivating Section~\ref{sec:resolution}.

\textbf{Image-to-Pixel Generation.}
Table~\ref{tab:i2p-main-results} compares six multimodal
language models with three palette-matched algorithms. Five models are fully
valid; Kimi K3 reaches 96.3\%, and only Gemini 3.6 Flash produces flat fills
(1.0\%). GPT-5.6 Sol leads model Top-1 Avg (56.0\%), while GPT-6 Astra leads
model SigLIP (.654) and DINO (.175) averages. The best model exceeds all
algorithms at $16^2$ and $32^2$, but algorithms lead at $64^2$: PIA reaches
78\% Top-1 and .697 SigLIP, and SSIM-optimized pixelization reaches .543 DINO.
Four models also decline from $32^2$ to $64^2$. GPT-6 Astra achieves the highest Aggregated score
(80.45), ahead of PIA (77.50). See Appendix~\ref{app:baselines} for all
algorithms.

\begin{table}[t]
\vspace{-3pt}
\centering
\setpositionedtablecaption{\tablecaptionfont\textbf{Image-to-Pixel results.}
Valid and Flat are percentages over 300 model--resolution items for models and
100 images per resolution for algorithms. Top-1 uses the 100 source images;
SigLIP and DINO are paired cosines, and invalid generations score zero. Agg.\
denotes the Aggregated score: the mean of the min--max-normalized Top-1,
SigLIP, and DINO average columns. Colors are normalized within each metric
family; bold and underline mark the top two results per column.}
{tab:i2p-main-results}
\scriptsize
\setlength{\tabcolsep}{4pt}
\resizebox{\linewidth}{!}{%
\begin{tabular}{@{}lrr
!{\color{white}\vrule width 1.2pt}r
!{\color{white}\vrule width 1.2pt}r
!{\color{white}\vrule width 1.2pt}r
!{\color{white}\vrule width 1.2pt}r
!{\color{white}\vrule width 1.2pt}r
!{\color{white}\vrule width 1.2pt}r
!{\color{white}\vrule width 1.2pt}r
!{\color{white}\vrule width 1.2pt}r
!{\color{white}\vrule width 1.2pt}r
!{\color{white}\vrule width 1.2pt}r
!{\color{white}\vrule width 1.2pt}r
!{\color{white}\vrule width 1.2pt}r
!{\color{white}\vrule width 1.2pt}r@{\hspace{\tabcolsep}}}
\toprule
& & & \multicolumn{4}{c}{100-way Top-1} &
\multicolumn{4}{c}{SigLIP cosine} &
\multicolumn{4}{c}{DINO cosine} & \\
\cmidrule(lr){4-7}\cmidrule(lr){8-11}\cmidrule(lr){12-15}
Model / method & Valid & Flat &
$16$ & $32$ & $64$ & Avg &
$16$ & $32$ & $64$ & Avg &
$16$ & $32$ & $64$ & Avg & Agg. \\
\midrule
\multicolumn{16}{@{}l}{\textcolor{black!50}{\emph{Multimodal language models}}} \\
GPT-6 Astra & 100.0 & 0.0 &
\scoreblue{7}{18} & \scoreblue{30}{\underline{50}} &
\scoreblue{49}{\underline{76}} & \scoreblue{28}{48.0} &
\scoreblue{29}{\textbf{.615}} & \scoreblue{39}{\textbf{.652}} &
\scoreblue{49}{\underline{.695}} & \scoreblue{39}{\textbf{.654}} &
\scoreblue{1}{\textbf{.023}} & \scoreblue{16}{\textbf{.177}} &
\scoreblue{30}{.326} & \scoreblue{15}{.175} &
\scoreblue{50}{\textbf{80.45}} \\
GPT-5.6 Sol & 100.0 & 0.0 &
\scoreblue{25}{\textbf{44}} & \scoreblue{38}{\textbf{61}} &
\scoreblue{39}{63} & \scoreblue{34}{\textbf{56.0}} &
\scoreblue{17}{\underline{.569}} & \scoreblue{24}{.597} &
\scoreblue{25}{.599} & \scoreblue{22}{.588} &
\scoreblue{1}{\underline{.022}} & \scoreblue{9}{.112} &
\scoreblue{13}{.147} & \scoreblue{8}{.094} &
\scoreblue{35}{57.27} \\
Claude Opus 5 & 100.0 & 0.0 &
\scoreblue{12}{26} & \scoreblue{29}{49} &
\scoreblue{28}{47} & \scoreblue{23}{40.7} &
\scoreblue{9}{.535} & \scoreblue{17}{.567} &
\scoreblue{21}{.582} & \scoreblue{15}{.561} &
\scoreblue{1}{.021} & \scoreblue{5}{.062} &
\scoreblue{8}{.098} & \scoreblue{5}{.060} &
\scoreblue{14}{23.16} \\
Kimi K3 & 96.3 & 0.0 &
\scoreblue{18}{\underline{34}} & \scoreblue{27}{46} &
\scoreblue{15}{30} & \scoreblue{20}{36.7} &
\scoreblue{15}{.558} & \scoreblue{16}{.562} &
\scoreborder{.501} & \scoreblue{10}{.541} &
\scoreblue{1}{.020} & \scoreblue{5}{.064} &
\scoreblue{2}{.038} & \scoreblue{3}{.040} &
\scoreblue{4}{8.48} \\
Seed-2.1 Pro & 100.0 & 0.0 &
\scoreblue{17}{33} & \scoreblue{24}{42} &
\scoreblue{15}{30} & \scoreblue{19}{35.0} &
\scoreblue{9}{.536} & \scoreblue{13}{.553} &
\scoreblue{13}{.553} & \scoreblue{12}{.547} &
\scoreblue{1}{.020} & \scoreblue{4}{.056} &
\scoreblue{4}{.054} & \scoreblue{3}{.043} &
\scoreblue{4}{8.52} \\
Gemini 3.6 Flash & 100.0 & 1.0 &
\scoreblue{14}{28} & \scoreblue{20}{37} &
\scoreblue{13}{27} & \scoreblue{16}{30.7} &
\scoreblue{9}{.535} & \scoreblue{11}{.546} &
\scoreblue{9}{.536} & \scoreblue{10}{.539} &
\scoreblue{1}{.018} & \scoreblue{5}{.065} &
\scoreblue{5}{.070} & \scoreblue{4}{.051} &
\scoreborder{1.98} \\
\midrule
\multicolumn{16}{@{}l}{\textcolor{black!50}{\emph{Algorithmic baselines}}} \\
Nearest + median cut & 100.0 & 0.0 &
\scoreborder{9} & \scoreblue{14}{28} &
\scoreblue{38}{62} & \scoreblue{17}{33.0} &
\scoreblue{6}{.523} & \scoreblue{17}{.568} &
\scoreblue{42}{.665} & \scoreblue{21}{.585} &
\scoreborder{.012} & \scoreblue{7}{.090} &
\scoreblue{44}{.480} & \scoreblue{17}{.194} &
\scoreblue{27}{44.11} \\
SSIM-opt. + median cut & 100.0 & 0.0 &
\scoreborder{9} & \scoreblue{18}{34} &
\scoreblue{48}{75} & \scoreblue{22}{39.3} &
\scoreblue{6}{.524} & \scoreblue{20}{.579} &
\scoreblue{47}{.686} & \scoreblue{24}{.596} &
\scoreblue{0.1}{.013} & \scoreblue{10}{.119} &
\scoreblue{50}{\textbf{.543}} & \scoreblue{20}{\textbf{.225}} &
\scoreblue{38}{61.19} \\
PIA (reduced) & 100.0 & 0.0 &
\scoreblue{7}{18} & \scoreblue{30}{\underline{50}} &
\scoreblue{50}{\textbf{78}} & \scoreblue{29}{\underline{48.7}} &
\scoreblue{11}{.546} & \scoreblue{27}{\underline{.608}} &
\scoreblue{50}{\textbf{.697}} & \scoreblue{30}{\underline{.617}} &
\scoreblue{0.2}{.014} & \scoreblue{12}{\underline{.135}} &
\scoreblue{45}{\underline{.490}} & \scoreblue{19}{\underline{.213}} &
\scoreblue{48}{\underline{77.50}} \\
\bottomrule
\end{tabular}%
}
\vspace{-3mm}
\end{table}

\def\pixelcorrelationsummary{%
\centering
\scriptsize
\setlength{\tabcolsep}{2.2pt}
\resizebox{\linewidth}{!}{%
\begin{tabular}{@{}lcccc@{}}
\toprule
Comparison & $n$ & Pearson $r$ & Spearman $\rho$ & Kendall $\tau$ \\
\midrule
Understanding--T2P & 7 & .871 & 1.000 & 1.000 \\
Understanding--I2P (w/o DeepSeek) & 6 & .930 & .829 & .733 \\
\bottomrule
\end{tabular}}}

\begin{wraptable}{r}{0.60\linewidth}
\vspace{-10pt}
\centering
\begin{minipage}[t]{\linewidth}
\caption{\tablecaptionfont\textbf{Pixel understanding.}
Each source contributes 100 ten-way questions per resolution. Overall averages
600 questions; each source average covers 300. Chance accuracy is 10\%.
Background color is globally normalized from 8 (white) to 85 (blue); bold
and underline mark the best and second-best values in each column.}
\label{tab:pixel-understanding-results}
\vspace{0pt}
\centering
\scriptsize
\setlength{\tabcolsep}{4.5pt}
\resizebox{\linewidth}{!}{%
\begin{tabular}{@{}l
!{\color{white}\vrule width 1.2pt}c
!{\color{white}\vrule width 1.2pt}cccc
!{\color{white}\vrule width 1.2pt}cccc@{\hspace{\tabcolsep}}}
\toprule
& & \multicolumn{4}{c}{PIA source} &
\multicolumn{4}{c}{Astra-I2P source} \\
\cmidrule(lr){3-6}\cmidrule(lr){7-10}
Model & Overall &
$16$ & $32$ & $64$ & Avg &
$16$ & $32$ & $64$ & Avg \\
\midrule
GPT-6 Astra & \scoreblue{27}{\textbf{67.5}} &
\scoreblue{18}{\textbf{48}} & \scoreblue{22}{\textbf{57}} &
\scoreblue{27}{\textbf{67}} & \scoreblue{22}{\textbf{57.3}} &
\scoreblue{29}{\textbf{72}} & \scoreblue{31}{\textbf{76}} &
\scoreblue{35}{\textbf{85}} & \scoreblue{32}{\textbf{77.7}} \\
GPT-5.6 Sol & \scoreblue{14}{\underline{39.2}} &
\scoreblue{12}{\underline{34}} & \scoreblue{10}{\underline{29}} &
\scoreblue{12}{\underline{34}} & \scoreblue{11}{\underline{32.3}} &
\scoreblue{17}{\underline{45}} & \scoreblue{15}{42} &
\scoreblue{20}{\underline{51}} & \scoreblue{17}{46.0} \\
Claude Opus 5 & \scoreblue{12}{35.3} &
\scoreblue{6}{22} & \scoreblue{8}{26} & \scoreblue{7}{24} &
\scoreblue{7}{24.0} & \scoreblue{15}{40} &
\scoreblue{19}{\underline{50}} & \scoreblue{19}{50} &
\scoreblue{18}{\underline{46.7}} \\
Kimi K3 & \scoreblue{9}{27.3} &
\scoreblue{5}{20} & \scoreblue{5}{19} & \scoreblue{4}{16} &
\scoreblue{5}{18.3} & \scoreblue{14}{39} & \scoreblue{12}{34} &
\scoreblue{13}{36} & \scoreblue{13}{36.3} \\
Gemini 3.6 Flash & \scoreblue{6}{21.8} &
\scoreblue{6}{21} & \scoreblue{1}{11} & \scoreblue{4}{17} &
\scoreblue{4}{16.3} & \scoreblue{9}{28} & \scoreblue{7}{24} &
\scoreblue{10}{30} & \scoreblue{9}{27.3} \\
Seed-2.1 Pro & \scoreblue{4}{16.0} &
\scoreblue{3}{15} & \scoreblue{3}{14} & \scoreblue{2}{12} &
\scoreblue{3}{13.7} & \scoreblue{5}{20} & \scoreblue{3}{15} &
\scoreblue{5}{20} & \scoreblue{5}{18.3} \\
DeepSeek V4 Pro & \scoreblue{2}{12.5} &
\scoreblue{2}{13} & \scoreblue{1}{10} & \scoreborder{8} &
\scoreblue{1}{10.3} & \scoreblue{3}{15} & \scoreblue{3}{14} &
\scoreblue{3}{15} & \scoreblue{3}{14.7} \\
\bottomrule
\end{tabular}%
}
\end{minipage}
\vspace{12pt}
\refstepcounter{table}
\label{tab:pixel-understanding-correlations}
\parbox{\linewidth}{\tablecaptionfont\textbf{Table~\thetable: Correlation
summary.} Pearson, Spearman, and Kendall correlations between pixel
understanding and generation performance across models, using the T2P and I2P
Aggregated scores as the respective generation summaries. The I2P comparison
includes only models with available I2P generation results.}
\par\vspace{4pt}
\pixelcorrelationsummary
\vspace{-7pt}
\end{wraptable}

\textbf{Pixel Understanding.} We construct understanding inputs from two complementary sources. PIA is
deterministic but not uniformly optimal. GPT-6 Astra
achieves the highest I2P Aggregated score (80.45), ahead of PIA (77.50), and at $32^2$ also leads PIA
on SigLIP cosine (.652 vs.\ .608) and DINO cosine (.177 vs.\ .135).
Figure~\ref{fig:i2p-case-studies} and
Appendix~\ref{app:qualitative-cases} further show cases in which
model-generated outputs match or surpass deterministic pixelization, so we
also use Astra-I2P as a model-generated oracle.

Table~\ref{tab:pixel-understanding-results} reports recognition accuracy for
both sources. GPT-6 Astra leads overall accuracy (67.5\%), followed by GPT-5.6
Sol (39.2\%). On PIA alone it already leads by a wide margin (57.3\% vs.\
32.3\% for GPT-5.6 Sol), and the two sources largely preserve the same
ranking. Switching to Astra-I2P raises every model's source average, with
Claude Opus~5 gaining more than Astra ($+22.7$ vs.\ $+20.4$), so the
model-based oracle helps broadly rather than rewarding Astra for recognizing
its own generations. Overall accuracy correlates strongly with generation
performance: the Pearson correlation is 0.871 for T2P and 0.930 for I2P.
Table~\ref{tab:pixel-understanding-correlations} also reports Spearman and
Kendall coefficients, which support the same strong relationship. The T2P
comparison uses all evaluated models, whereas the I2P comparison uses those
with available generation results. Since the comparison covers only a small
set of models, we treat these correlations as descriptive rather than general
and use them to motivate the training experiment in
Section~\ref{sec:prt}.

\begin{figure}[t]
\centering
\includegraphics[width=\linewidth]{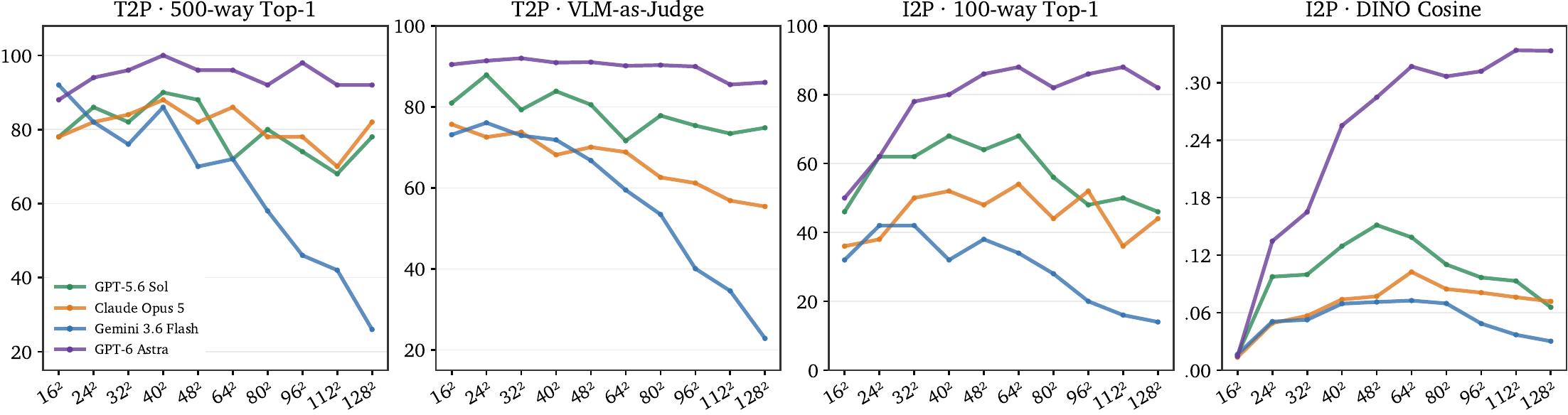}
\vspace{-6mm}
\caption{\footnotesize\textbf{Fine-grained resolution scaling on PixelBench.}
From left to right: T2P 500-way Top-1, T2P VLM-as-Judge, I2P 100-way
Top-1, and paired I2P DINOv2 CLS cosine from $16^2$ to $128^2$. Retrieval
and VLM judging agree that higher resolution does not consistently improve
T2P: Astra remains robust, while the other models fluctuate or decline. In
I2P, Astra continues to benefit from larger grids, whereas the other models
generally peak at intermediate resolutions and then decline.}
\label{fig:resolution-sweep}
\vspace{-2mm}
\end{figure}

\vspace{-1mm}
\section{Resolution Scaling and Failure Modes}
\label{sec:resolution}
\vspace{-1mm}

The main results suggest that higher resolution is not
uniformly better: for many models in both T2P and I2P, $64^2$ underperforms
$32^2$ (Tables~\ref{tab:t2p-main-results}
and~\ref{tab:i2p-main-results}). To examine this behavior more closely, we
sweep {GPT-6 Astra},
{GPT-5.6 Sol}, {Claude Opus
5}, and {Gemini 3.6 Flash} over ten resolutions from
$16^2$ to $128^2$. The number of predicted grid positions grows quadratically
with side length: relative to $16^2$, $64^2$ and $128^2$ require 16$\times$ and
64$\times$ as many pixel indices. Resolution scaling therefore increases both
representational capacity and the burden of coordinating a long spatial
output.

\textbf{Text-to-Pixel Generation.} 
The left two panels of Figure~\ref{fig:resolution-sweep} report 500-way retrieval and VLM-as-judge quality, showing that increasing resolution does not consistently improve performance under either metric.
GPT-6 Astra maintains strong performance across resolutions, with its judge score remaining near its peak through $96^2$ and declining only modestly at the two highest resolutions. GPT-5.6 Sol fluctuates after an early peak, while Claude Opus 5 exhibits oscillating retrieval performance but a clearer downward trend under VLM judging. Gemini 3.6 Flash shows the sharpest degradation at high resolutions, particularly beyond $64^2$.
The consistent agreement between retrieval and judge-based evaluations also suggests that the lack of monotonic improvement with resolution is not an artifact of the retrieval metric.

\textbf{Image-to-Pixel Generation.} 
The right two panels reveal a clearer scaling pattern. GPT-6 Astra increasingly benefits from larger grids, with both retrieval accuracy and DINO similarity  remaining strong at the highest resolutions. In contrast, the other models generally peak at intermediate resolutions and then decline. These results suggest that larger grids offer additional expressive capacity, but only GPT-6 Astra can translate this capacity into more faithful reconstructions.

\begin{figure}[t]
\centering
\includegraphics[width=\linewidth]{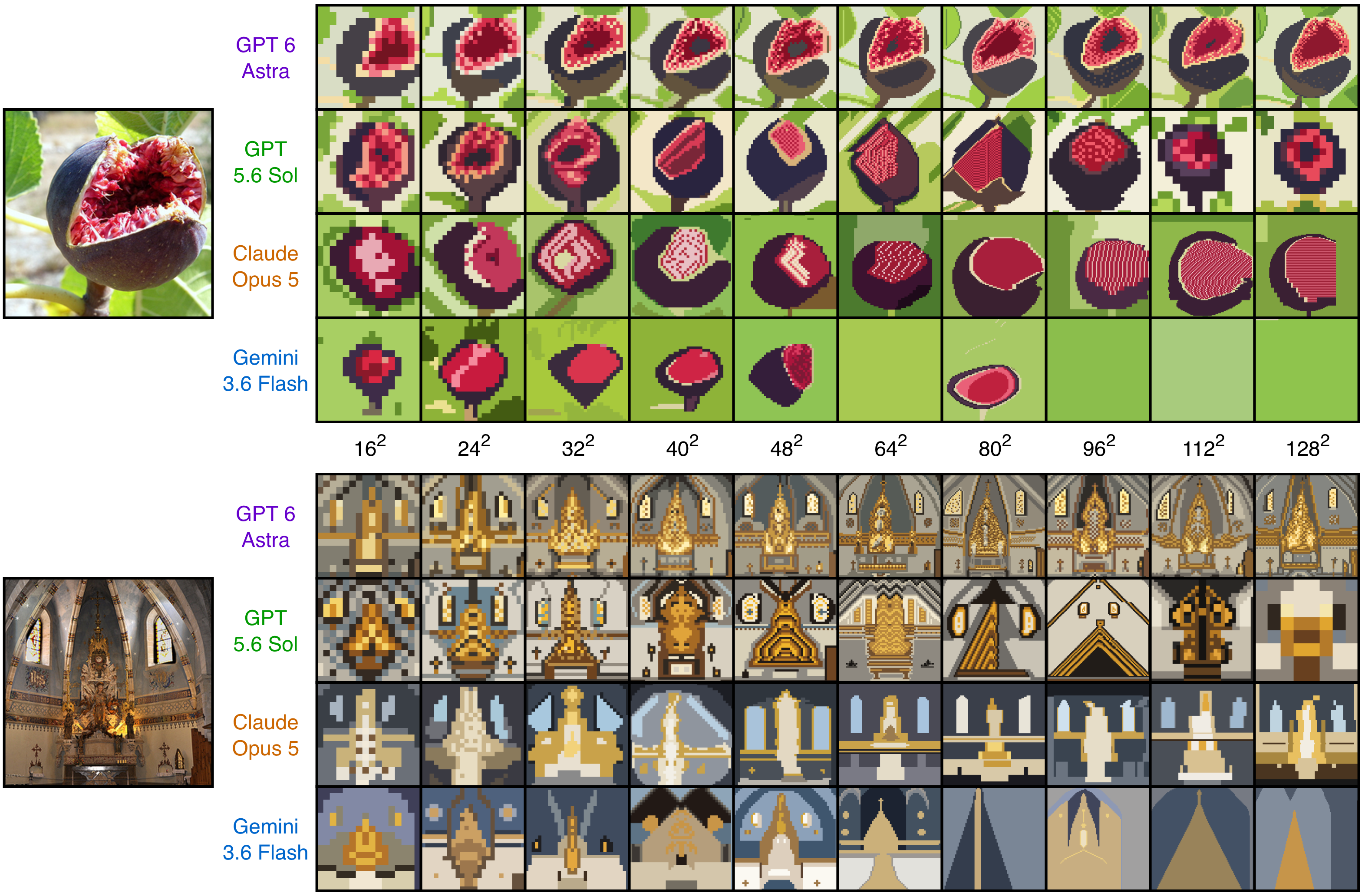}
\vspace{-6mm}
\caption{\footnotesize\textbf{Qualitative I2P resolution sweep.}
Two source images (left) are reconstructed from $16^2$ to $128^2$; rows show
GPT-6 Astra, GPT-5.6 Sol, Claude Opus 5, and Gemini 3.6 Flash. Astra remains strongest and adds coherent detail
as resolution grows; GPT-5.6 Sol collapses to a lower effective resolution,
Claude mechanically repeats central details in one case and fails to recover
fine architectural structure in the other,
and Gemini increasingly
omits the subject or other key structures.}
\label{fig:i2p-resolution-sweep-examples}
\vspace{1mm}
\end{figure}

To inspect these differences directly,
Figure~\ref{fig:i2p-resolution-sweep-examples} shows two I2P examples across
the same resolution sweep. {GPT-6 Astra}
progressively adds texture, sharper boundaries, and architectural detail while
preserving the subject and overall layout.
{GPT-5.6 Sol} instead exhibits an
\emph{effective-resolution collapse}: despite the increasing nominal grid
size, its outputs become dominated by large color blocks. This behavior can
preserve a coarse semantic layout, but it underuses the available grid and
reduces fidelity. {Claude Opus 5} shows
different failures across the two inputs. In the upper example, it
mechanically repeats details from the central region, sacrificing the fidelity
of the global reconstruction. In the lower indoor-architecture example, it
preserves only a coarse layout and fails to recover fine structural details.
{Gemini 3.6 Flash} degrades most severely, with the
subject or key structures gradually vanishing at high resolutions.

Together, these examples expose three high-resolution
failure modes: effective-resolution collapse, repetitive local patterns that
disrupt global structure, and omission of the subject or other key structures.
They also reveal a broader consistency gap across resolutions. Beyond how
effectively models use additional grid positions, other models can produce
substantially different reconstructions as
resolution increases, whereas Astra preserves subject identity, composition,
and overall appearance across scales. Astra's advantage is therefore not
merely its ability to produce longer outputs, but its ability to organize
additional grid positions into coherent details while maintaining a stable
reconstruction. Representative T2P cases in
Appendix~\ref{app:resolution-examples} exhibit the same failure modes.

\vspace{-1.25mm}
\section{Pixel Reconstruction Tuning}
\label{sec:prt}
\vspace{-1.25mm}

The strong correlation between pixel generation and
pixel understanding in PixelBench raises a training question: can pixel
generation itself serve as an objective that improves a VLM's general visual
understanding? RecA~\citep{xie2025reconstruction} explores a related idea for unified multimodal models,
training them to reconstruct input images from their visual representations
and thereby improving general image generation and editing. For visual understanding, however, conventional
image-text and VQA supervision provides sparse, non-unique textual targets
rather than dense supervision over visual content. 
Pixel generation provides such a target by requiring a VLM to reconstruct its input as a palette-indexed pixel grid, turning the image itself into deterministic, spatially aligned training supervision. We term this approach Pixel Reconstruction Tuning (PRT).

\vspace{-.75mm}
\subsection{Data and Training}
\label{sec:prt-training}
\vspace{-.75mm}

\textbf{Pixel Reconstruction Tuning Data Construction.}
We sample images from a 20k-image PixMo-Cap pool \citep{deitke2025molmo} after removing exact matches with evaluation images. To avoid distortion from square reconstruction targets, we discard images with aspect ratios above 1.33. We also exclude text-heavy and multi-panel images that cannot be faithfully represented at resolutions of $16^2$ to $32^2$.
We render both resolutions with the
PIA-reduced procedure (Appendix~\ref{app:baselines}) and compute the SigLIP2
cosine similarity between each rendered pixel image and its source image. We
retain the top 40\% at $16^2$ and the top 80\% at $32^2$. For each retained
image, we query Gemini 3.8 Flash to write an image description and summarize
the colors in the algorithmically constructed palette. We package these with
the grid using \texttt{<CAPTION>}, \texttt{<PALETTE>}, and \texttt{<GRID>}, while keeping the palette and grid values deterministic. 
The retained examples are our PRT data (see Appendix~\ref{app:prt-additional-results}).

\begin{table}[t]
\centering
\caption{\tablecaptionfont{\textbf{PRT evaluation suite.}
The suite comprises 16 benchmarks spanning four capability groups. For each
benchmark, we summarize the tested content and number of evaluated examples
($N$).}}
\label{tab:prt-benchmark-suite}
\fontsize{7.4}{8.6}\selectfont
\renewcommand{\arraystretch}{.95}
\newlength{\prtbenchwidth}\newlength{\prtnwidth}
\settowidth{\prtbenchwidth}{BlindTest \citep{rahmanzadehgervi2024vision}}
\settowidth{\prtnwidth}{3{,}434}
\begin{tabular}{@{}
>{\raggedright\arraybackslash}p{\prtbenchwidth}
@{\hspace{8pt}}
>{\raggedright\arraybackslash}p{\dimexpr\linewidth-\prtbenchwidth-\prtnwidth-16pt\relax}
@{\hspace{8pt}}
>{\raggedleft\arraybackslash}p{\prtnwidth}
@{}}
\toprule
Benchmark & Task content & $N$ \\
\midrule
\multicolumn{3}{@{}l}{\textcolor{black!50}{\emph{General perception and reasoning}}} \\
MMStar \citep{chen2024mmstar}
& Vision-essential perception and reasoning across six capability areas & 1{,}500 \\
MMBench \citep{liu2024mmbench}
& Multiple-choice perception and reasoning over diverse visual content & 1{,}292 \\
BLINK \citep{fu2024blink}
& Fourteen classic computer-vision tasks recast as visual question answering & 1{,}901 \\
BabyVision \citep{chen2026babyvision}
& Foundational visual perception and reasoning over simple scenes & 388 \\
PerceptionBench \citep{lin2026perceptionbench}
& Atomic perception of existence, count, position, color, and relations & 3{,}000 \\
\arrayrulecolor{black!40}\midrule\arrayrulecolor{black}
\multicolumn{3}{@{}l}{\textcolor{black!50}{\emph{Fine-grained visual and spatial detail}}} \\
MMVP \citep{tong2024eyes}
& Fine-grained visual details missed by CLIP-style vision encoders & 150 \\
Q-Bench \citep{wu2024qbench}
& Low-level visual attributes, image quality, and visual perception & 1{,}495 \\
HallusionBench \citep{guan2024hallusionbench}
& Image-context reasoning that exposes visual illusion and hallucination & 1{,}129 \\
V-STAR Bench \citep{wu2023vstar}
& High-resolution visual search and grounding for small or hidden targets & 191 \\
CV-Bench \citep{tong2024cambrian}
& 2D spatial relations and counting; 3D depth and distance & 2{,}638 \\
BlindTest \citep{rahmanzadehgervi2024vision}
& Geometric primitives, spatial relations, and precise attribute reading & 3{,}434 \\
VLMs Are Biased \citep{vo2026biased}
& Counting in counterfactual images that conflict with learned priors & 2{,}784 \\
\arrayrulecolor{black!40}\midrule\arrayrulecolor{black}
\multicolumn{3}{@{}l}{\textcolor{black!50}{\emph{Color}}} \\
ColorBench \citep{liang2025colorbench}
& Color recognition, extraction, comparison, counting, and reasoning & 955 \\
ColorBench-Robust \citep{liang2025colorbench}
& Answer consistency across each original image and nine recolored variants & 493 \\
\arrayrulecolor{black!40}\midrule\arrayrulecolor{black}
\multicolumn{3}{@{}l}{\textcolor{black!50}{\emph{Charts and mathematics}}} \\
ChartQA \citep{masry2022chartqa}
& Visual extraction and logical or arithmetic reasoning over real-world charts & 2{,}500 \\
MathVista \citep{lu2024mathvista}
& Mathematical reasoning over diagrams, plots, puzzles, and scientific figures & 1{,}000 \\
\bottomrule
\end{tabular}
\vspace{-3mm}
\end{table}

\begin{wraptable}{r}{0.45\linewidth}
\vspace{-16pt}
\centering
\caption{\tablecaptionfont\textbf{Molmo2-4B post-training
with Pixel trajectories.} Four-seed means for OV 500k and $+$2k Pixel;
$\Delta$ is $+$Pixel minus OV. Example avg.\ pools 24,850 official
evaluation units; Benchmark avg.\ weights the 16 benchmarks equally.}
\label{tab:molmo-main}
\vspace{0pt}
\scriptsize
\setlength{\tabcolsep}{3.5pt}
\renewcommand{\arraystretch}{0.92}
\begin{tabular*}{\linewidth}{@{\extracolsep{\fill}}lcccc}
\toprule
Benchmark & $N$ & OV & $+$Pixel & $\Delta$ \\
\midrule
BlindTest          & 3{,}434 & 58.3 & 59.7 & \cellcolor{deltapos!30}$+1.4$ \\
VLM Bias           & 2{,}784 & 26.1 & 27.2 & \cellcolor{deltapos!25}$+1.1$ \\
CV-Bench           & 2{,}638 & 75.8 & 76.7 & \cellcolor{deltapos!21}$+0.9$ \\
ColorBench-R.      &   493 & 61.8 & 62.6 & \cellcolor{deltapos!19}$+0.8$ \\
BLINK              & 1{,}901 & 52.8 & 53.6 & \cellcolor{deltapos!19}$+0.8$ \\
MMStar             & 1{,}500 & 55.3 & 56.0 & \cellcolor{deltapos!19}$+0.8$ \\
MathVista          & 1{,}000 & 56.2 & 56.7 & \cellcolor{deltapos!13}$+0.5$ \\
ColorBench         &   955 & 32.1 & 32.5 & \cellcolor{deltapos!11}$+0.4$ \\
Q-Bench            & 1{,}495 & 71.9 & 72.2 & \cellcolor{deltapos!9}$+0.3$ \\
Hallusion          & 1{,}129 & 59.6 & 59.9 & \cellcolor{deltapos!7}$+0.2$ \\
MMVP               &   150 & 58.7 & 58.8 & \cellcolor{deltapos!7}$+0.2$ \\
ChartQA            & 2{,}500 & 77.9 & 77.9 & \cellcolor{white}$+0.0$ \\
Perception         & 3{,}000 & 25.9 & 25.9 & \cellcolor{white}$+0.0$ \\
MMBench            & 1{,}292 & 78.6 & 78.6 & \cellcolor{white}$+0.0$ \\
BabyVision         &   388 & 10.2 & 10.1 & \cellcolor{deltaneg!7}$-0.1$ \\
V-STAR             &   191 & 65.2 & 65.1 & \cellcolor{deltaneg!7}$-0.1$ \\
\arrayrulecolor{black!50}\midrule\arrayrulecolor{black}
Example avg.   & 24{,}850 & 54.2 & 54.8 & \cellcolor{deltapos!15}$+0.6$ \\
Benchmark avg. & 16 & 54.1 & 54.6 & \cellcolor{deltapos!13}$+0.5$ \\
\bottomrule
\end{tabular*}

\vspace{0pt}
\caption{\tablecaptionfont\textbf{Pixel-quantity
ablation.} Changes relative to OV-only post-training. The 2k setting averages
four seeds; the others average two. Bold and underline mark the best and
second-best values in each row.}
\label{tab:molmo-quantity-ablation}
\vspace{0pt}
\scriptsize
\setlength{\tabcolsep}{3.5pt}
\begin{tabular*}{\linewidth}{@{\extracolsep{\fill}}lcccc}
\toprule
Avg. & $\Delta_{\mathrm{1k}}$ & $\Delta_{\mathrm{2k}}$ & $\Delta_{\mathrm{4k}}$ & $\Delta_{\mathrm{8k}}$ \\
\midrule
Example
& \cellcolor{white}+0.40
& \cellcolor{deltapos!30}\textbf{+0.61}
& \cellcolor{white}+0.36
& \cellcolor{deltapos!15}\underline{+0.50} \\
Benchmark
& \cellcolor{deltapos!30}\textbf{+0.56}
& \cellcolor{white}+0.45
& \cellcolor{white}+0.39
& \cellcolor{deltapos!15}\underline{+0.51} \\
\bottomrule
\end{tabular*}
\vspace{-30pt}
\end{wraptable}

\textbf{Post-training setup.} Starting from the
Molmo2-4B \citep{clark2026molmo2} step-32k pretraining checkpoint, we
post-train all modules for one pass on 500k LLaVA-OneVision-1.5 instructions
\citep{an2025llavaonevision}, adding PRT data only in treatment runs. The main
2k experiment (0.4\%; Table~\ref{tab:molmo-main}) uses four matched seed
pairs. The quantity ablation tests 1k, 2k, 4k, and 8k examples
(Table~\ref{tab:molmo-quantity-ablation}) on two matched seeds shared across
amounts; 2k includes two additional seeds and is reported over four pairs.
Within each pair, OV-only and OV$+$Pixel share the same random seed, isolating
the effect of PRT data.

\textbf{Evaluation suite.} We evaluate 16 benchmarks
across four groups (Table~\ref{tab:prt-benchmark-suite}): general
perception and reasoning, fine-grained visual and spatial detail, color, and
charts and mathematics. All use their full evaluation sets except BlindTest,
for which we sample a category-balanced subset.

\vspace{-.5mm}
\subsection{Main Results}
\vspace{-.5mm}

\textbf{PRT Improves VLM Performance.} In the controlled 2k comparison
(Table~\ref{tab:molmo-main}), Pixel trajectories raise Example averaging from
54.2 to 54.8 ($+0.6$) and Benchmark averaging from 54.1 to 54.6 ($+0.5$); the
seed-paired Benchmark gain is positive in three of four pairs
($+0.8/+0.5/+0.7/-0.1$). Across quantities (Table~\ref{tab:molmo-quantity-ablation}),
both aggregate deltas stay positive from 1k to 8k ($+0.36$-$+0.61$) without a
monotonic scaling trend; 2k is the only four-seed setting. At the benchmark
level, the largest gains are on BlindTest ($+1.4$), VLMs Are Biased ($+1.1$),
and CV-Bench ($+0.9$). ChartQA, PerceptionBench, and MMBench are unchanged,
while BabyVision and V-STAR decrease by only $0.1$ point. Example averaging
peaks at 2k ($+0.61$), whereas Benchmark averaging peaks at 1k ($+0.56$), further
indicating that the benefit does not scale monotonically with quantity. Full
results and caveats are in Appendix~\ref{app:prt-additional-results}.\par

\textbf{Pixel Reconstruction Serves as Regularization.}
We study Qwen3-VL-Instruct at 2B, 4B, and 8B. FV uses 64k FineVision examples;
FV$+$Pixel adds 3k examples that directly pair an input image with its
deterministic palette-indexed reconstruction. Here, the target contains only
the reconstructed pixels, with no accompanying image description or palette
explanation.
Table~\ref{tab:qwen-scale-summary-main} shows positive
gains at every scale: 2B rises from 48.1/47.3 to 50.1/50.3
($+2.0/+3.0$), 4B from 53.9/53.6 to 56.8/57.3 ($+2.9/+3.7$), and 8B from
55.1/56.3 to 57.2/57.9 ($+2.1/+1.6$); 4B gains the most.
Resolution preferences vary by scale (Table~\ref{tab:qwen-resolution-summary-main}):
2B peaks at $32^2$ ($+2.50/+3.37$), 4B at $16^2$ ($+3.14/+4.07$), and
8B at Mix for Example averaging ($+2.07$) but $64^2$ for Benchmark averaging
($+1.73$). Mix nevertheless ranks first or second in four of six rows, making
it a robust choice without committing to a single resolution.\par
\WFclear

\begin{table}[t]
\noindent
\begin{minipage}[t]{0.40\linewidth}
\vspace{0pt}
\centering
\makeatletter\def\@captype{table}\makeatother
\caption{\tablecaptionfont\textbf{Qwen3-VL across
model scales.} FV$+$Pixel adds 3k Pixel examples to 64k FV examples.}
\label{tab:qwen-scale-summary-main}
\vspace{2pt}
\scriptsize
\setlength{\tabcolsep}{3pt}
\renewcommand{\arraystretch}{1.05}
\begin{tabular*}{\linewidth}{@{\extracolsep{\fill}}clccc}
\toprule
\rule[-0.7ex]{0pt}{2.6ex}Model & Avg. & FV & FV$+$Pixel & $\Delta$ \\
\midrule
\multirow{2}{*}{2B}
& Example   & 48.1 & 50.1 & \cellcolor{deltapos!18}+2.0 \\
& Benchmark & 47.3 & 50.3 & \cellcolor{deltapos!25}+3.0 \\
\arrayrulecolor{black!50}\midrule\arrayrulecolor{black}
\multirow{2}{*}{4B}
& Example   & 53.9 & 56.8 & \cellcolor{deltapos!24}+2.9 \\
& Benchmark & 53.6 & 57.3 & \cellcolor{deltapos!30}+3.7 \\
\arrayrulecolor{black!50}\midrule\arrayrulecolor{black}
\multirow{2}{*}{8B}
& Example   & 55.1 & 57.2 & \cellcolor{deltapos!18}+2.1 \\
& Benchmark & 56.3 & 57.9 & \cellcolor{deltapos!14}+1.6 \\
\bottomrule
\end{tabular*}
\end{minipage}\hfill
\begin{minipage}[t]{0.56\linewidth}
\vspace{0pt}
\centering
\makeatletter\def\@captype{table}\makeatother
\caption{\tablecaptionfont\textbf{Target-resolution
ablation.} Gains over FV; Mix spans \smash{$16^2$}, \smash{$32^2$}, and \smash{$64^2$}. Bold and
underline mark the best and second-best values per row.}
\label{tab:qwen-resolution-summary-main}
\vspace{2pt}
\scriptsize
\setlength{\tabcolsep}{3pt}
\renewcommand{\arraystretch}{1.05}
\begin{tabular*}{\linewidth}{@{\extracolsep{\fill}}clcccc}
\toprule
\rule[-0.7ex]{0pt}{2.6ex}Model & Avg. & $\Delta_{16}$ & $\Delta_{32}$ & $\Delta_{64}$ & $\Delta_{\mathrm{Mix}}$ \\
\midrule
\multirow{2}{*}{2B}
& Example   & \cellcolor{deltapos!8}+2.14 & \cellcolor{deltapos!30}\textbf{+2.50} & \cellcolor{deltapos!16}\underline{+2.27} & \cellcolor{white}+2.01 \\
& Benchmark & \cellcolor{white}+2.66 & \cellcolor{deltapos!30}\textbf{+3.37} & \cellcolor{deltapos!14}+2.99 & \cellcolor{deltapos!15}\underline{+3.02} \\
\arrayrulecolor{black!50}\midrule\arrayrulecolor{black}
\multirow{2}{*}{4B}
& Example   & \cellcolor{deltapos!30}\textbf{+3.14} & \cellcolor{deltapos!22}+2.87 & \cellcolor{white}+2.18 & \cellcolor{deltapos!24}\underline{+2.94} \\
& Benchmark & \cellcolor{deltapos!30}\textbf{+4.07} & \cellcolor{deltapos!21}\underline{+3.90} & \cellcolor{white}+3.48 & \cellcolor{deltapos!11}+3.70 \\
\arrayrulecolor{black!50}\midrule\arrayrulecolor{black}
\multirow{2}{*}{8B}
& Example   & \cellcolor{white}+1.88 & \cellcolor{deltapos!6}\underline{+1.92} & \cellcolor{deltapos!3}+1.90 & \cellcolor{deltapos!30}\textbf{+2.07} \\
& Benchmark & \cellcolor{deltapos!9}+1.38 & \cellcolor{white}+1.23 & \cellcolor{deltapos!30}\textbf{+1.73} & \cellcolor{deltapos!22}\underline{+1.60} \\
\bottomrule
\end{tabular*}
\end{minipage}
\end{table}

\vspace{-1mm}
\section{Concluding Remarks}
\vspace{-1mm}

We introduce PixelBench, a text-based interface for direct pixel generation and understanding without a learned image tokenizer or decoder. Across T2P, I2P, and understanding tasks, frontier models exhibit substantial yet uneven capabilities, with model-specific resolution peaks. GPT-6 Astra remains robust as grid resolution increases, whereas most other models initially improve before declining. Beyond evaluation, we introduce PRT, which transforms deterministic palettes and grids into instruction trajectories. By combining general instruction data with curated PRT data, we post-train Molmo2-4B from its pretraining checkpoint, improving performance across 16 visual-centric benchmarks. The performance gain can also be observed in Qwen3-VL across different scales. These findings highlight the dual role of explicit pixels as a behavioral probe for evaluating visual capabilities and a practical training signal for improving them.

\subsubsection*{Acknowledgments}

HZ thanks Ji Xie for a careful review of the paper and helpful suggestions. HZ
also thanks Yangyi Huang, Sihan Yang, Zhou Jiang, Yi Xie, and Hongrui Zhang for
useful comments.

\bibliography{iclr2027_conference}
\bibliographystyle{iclr2027_conference}

\clearpage
\newpage
\appendix
\onecolumn
\addcontentsline{toc}{section}{Appendix}
\renewcommand \thepart{}
\renewcommand \partname{}
\part{\Large{\centerline{Appendix}}}
\parttoc

\newpage

\section{Related Work}
\label{app:related-work}

\subsection{Symbolic Visual Representations}
\label{app:symbolic-representations}

A growing line of work represents visual content as a discrete symbol sequence
that a renderer executes, rather than as a dense array of pixels. On the
understanding side, SGP-Bench evaluates whether language models can infer the
semantics of SVG and CAD programs without seeing their renderings and improves
this ability through Symbolic Instruction Tuning \citep{qiu2024symbolic}, while
VGBench and SVGenius broaden evaluation to editing and generation across many
formats, domains, and complexity levels \citep{zou2024vgbench,chen2025svgenius}.
On the generation side, models emit SVG code or executable CAD programs directly
\citep{chen2025symbolicgraphics,rodriguez2025starvector,xing2025llm4svg,wu2025chat2svg,khan2024text2cad,rukhovich2025cadrecode,zhang2026funcad},
and \citet{sharma2024vision} use rendering code as a stand-in for pixels,
demonstrating that text-only models encode substantial visual knowledge.
Together, these studies show that general-purpose models encode usable
geometric knowledge in symbolic form, although their performance degrades when
precise spatial and structural fidelity is required.

Two properties separate that setting from ours. A program mediates between model
and image, so the renderer supplies geometry and resolution independence and
each primitive covers many pixels; token-based generative models instead keep
the raster form but substitute a learned codebook whose vocabulary is opaque and
whose decoding requires the paired detokenizer
\citep{chen2020igpt,chameleon2024,wang2024emu3}. PixelBench asks for an explicit
palette and per-pixel indices, so rendering is a table lookup and every emitted
symbol is one spatial location. The closest prior evaluation of this format is
Pixel Art Bench, a web leaderboard that ranks $24\times24$ grids for seven
subjects by LLM-judged Elo \citep{emergentmind2026pixelart}. Color-index
grids also appear in abstract reasoning \citep{chollet2019arc}.
PixelBench broadens this to text-to-pixel and image-to-pixel generation over
100 concepts and 100 natural images at three resolutions, adds a complementary
pixel-understanding evaluation, replaces judge-based ranking with
deterministic retrieval and fidelity metrics, and compares against
palette-matched pixelization and downscaling algorithms
\citep{oztireli2015perceptual,weber2016dpid,wu2022pixelization}.

\subsection{Vision-Language Training and Visual Representations}
\label{app:vl-training}

Contemporary VLMs pair a pretrained visual encoder with a language model through
a connector, aligned by image-text pretraining and visual instruction tuning
\citep{radford2021clip,zhai2023sigmoid,alayrac2022flamingo,li2023blip2,liu2023llava,dai2023instructblip,liu2024llava15}.
Controlled ablations attribute much of the reported progress to the unimodal
backbones and to training data rather than to the connector
\citep{karamcheti2024prismatic,laurencon2024matters}, and a parallel body of work
stresses the composition, quality, and cleanliness of that data: vision-centric
mixtures sharpen fine-grained perception \citep{tong2024cambrian}, curated
non-distilled and decontaminated corpora improve generalization
\citep{deitke2025molmo,wiedmann2025finevision}, and benchmark audits show that
apparent gains can stem from leakage or from questions answerable without the
image \citep{chen2024mmstar}. Across these pipelines the supervision target is
language, whatever the visual input.

That choice has consequences: distinctions lost by the encoder propagate into the
assembled model \citep{tong2024eyes}, which remains unreliable on low-level
spatial queries \citep{rahmanzadehgervi2024vision,fu2024blink}. Objectives that
require reproducing visual detail instead yield transferable representations
\citep{he2022mae,chen2020igpt}, and MetaMorph finds understanding and visual
generation mutually beneficial \citep{tong2025metamorph}.
Reconstruction Alignment (RecA) conditions unified multimodal models on their
own visual encoder representations and trains them to reconstruct the input
image, thereby connecting visual understanding and generation
\citep{xie2025reconstruction}. PRT studies an analogous objective for VLMs
without a dedicated image generator: predicting the palette-index grid itself
through the existing vocabulary, with no learned image tokenizer or decoder.
Our primary protocol starts from a Molmo2 pretraining checkpoint and adds
caption-palette-grid trajectories to a fixed 500k-example instruction
mixture. Appendix~\ref{app:prt-additional-results} additionally reports the
earlier Qwen3-VL experiments, where pixel targets are added to FineVision
post-training as a regularizer.

\section{Additional Resolution-Scaling Examples}
\label{app:resolution-examples}

\begin{figure}[H]
\centering
\includegraphics[width=\linewidth]{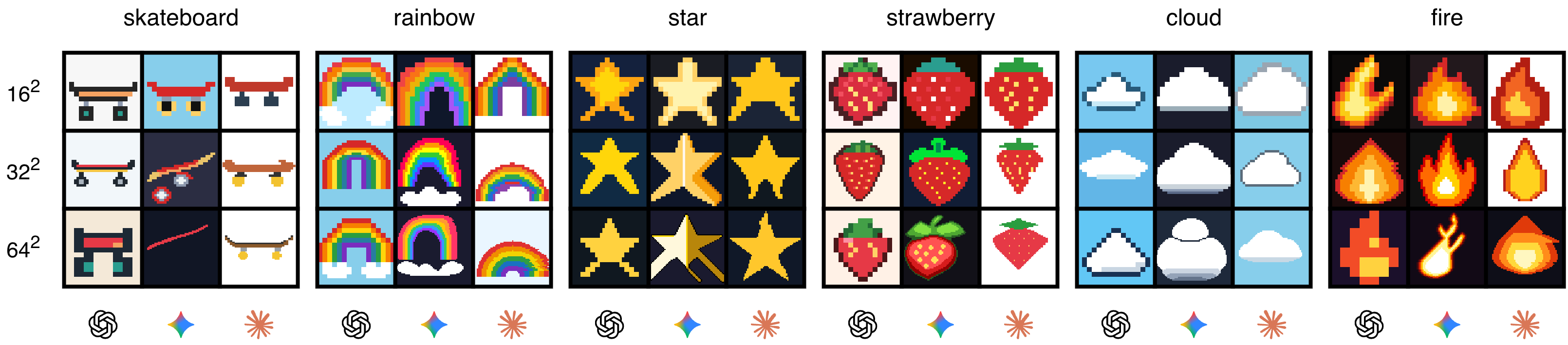}
\caption{\footnotesize\textbf{Model-dependent resolution behavior in T2P.}
Each case group shows GPT-5.6 Sol, Gemini 3.6 Flash, and Claude Opus 5 at
$16^2$, $32^2$, and $64^2$. Higher nominal resolution can yield finer detail,
coarse effective pixels, or degraded global structure depending on the model
and prompt.}
\label{fig:t2p-case-study}
\end{figure}

These cases instantiate the same three failure modes observed in the
qualitative I2P sweep. GPT-5.6 Sol exhibits
\emph{effective-resolution underutilization}:
its $64^2$ skateboard and fire are dominated by blocks roughly one-eighth of
the canvas width, giving them a granularity closer to an $8\times8$ grid. The
outputs remain recognizable, but the additional positions contribute little
detail. Gemini 3.6 Flash uses finer boundaries and local variation, but these
local patterns can disrupt global structure: at $64^2$, its skateboard
collapses to a diagonal stroke, its star loses the five-point structure, and
its fire loses a coherent silhouette. The latter cases also omit
identity-bearing structures. Claude Opus 5 varies by prompt: it preserves the
depicted objects in these examples while using the higher-resolution grid, but
its aggregate T2P performance fluctuates across resolutions and does not
sustain its intermediate-resolution peak.

\section{Additional Qualitative PixelBench Cases}
\label{app:qualitative-cases}

The following pages expand Figure~\ref{fig:i2p-case-studies} with 20
additional T2P cases and 20 additional I2P cases. Each T2P case compares all
seven evaluated models at $16^2$, $32^2$, and $64^2$. Each I2P case compares
the six evaluated models and three algorithmic baselines at the same
resolutions. The I2P panels provide additional qualitative evidence that some
model-based reconstructions preserve semantic structure more clearly than the
algorithmic baselines at low resolutions, complementing the aggregate results
in Table~\ref{tab:i2p-main-results}. Pixel images use nearest-neighbor
enlargement for display.

\newcommand{\nativecasepair}[2]{%
  \clearpage
  \vspace*{\fill}
  \begin{center}
  \includegraphics[width=\linewidth,height=0.42\textheight,keepaspectratio]{#1}\\[0.6em]
  \includegraphics[width=\linewidth,height=0.42\textheight,keepaspectratio]{#2}
  \end{center}
  \vspace*{\fill}}
\nativecasepair{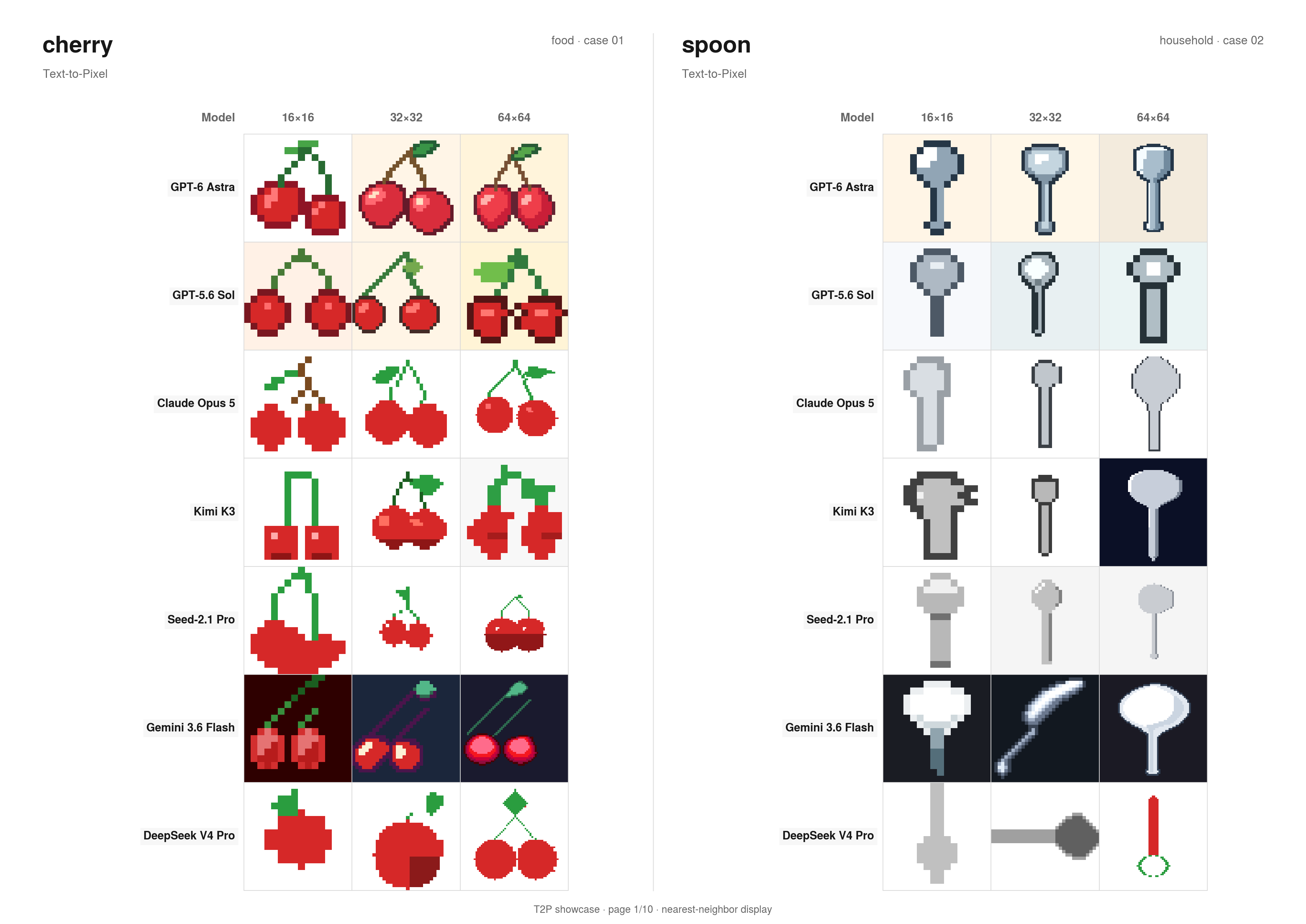}{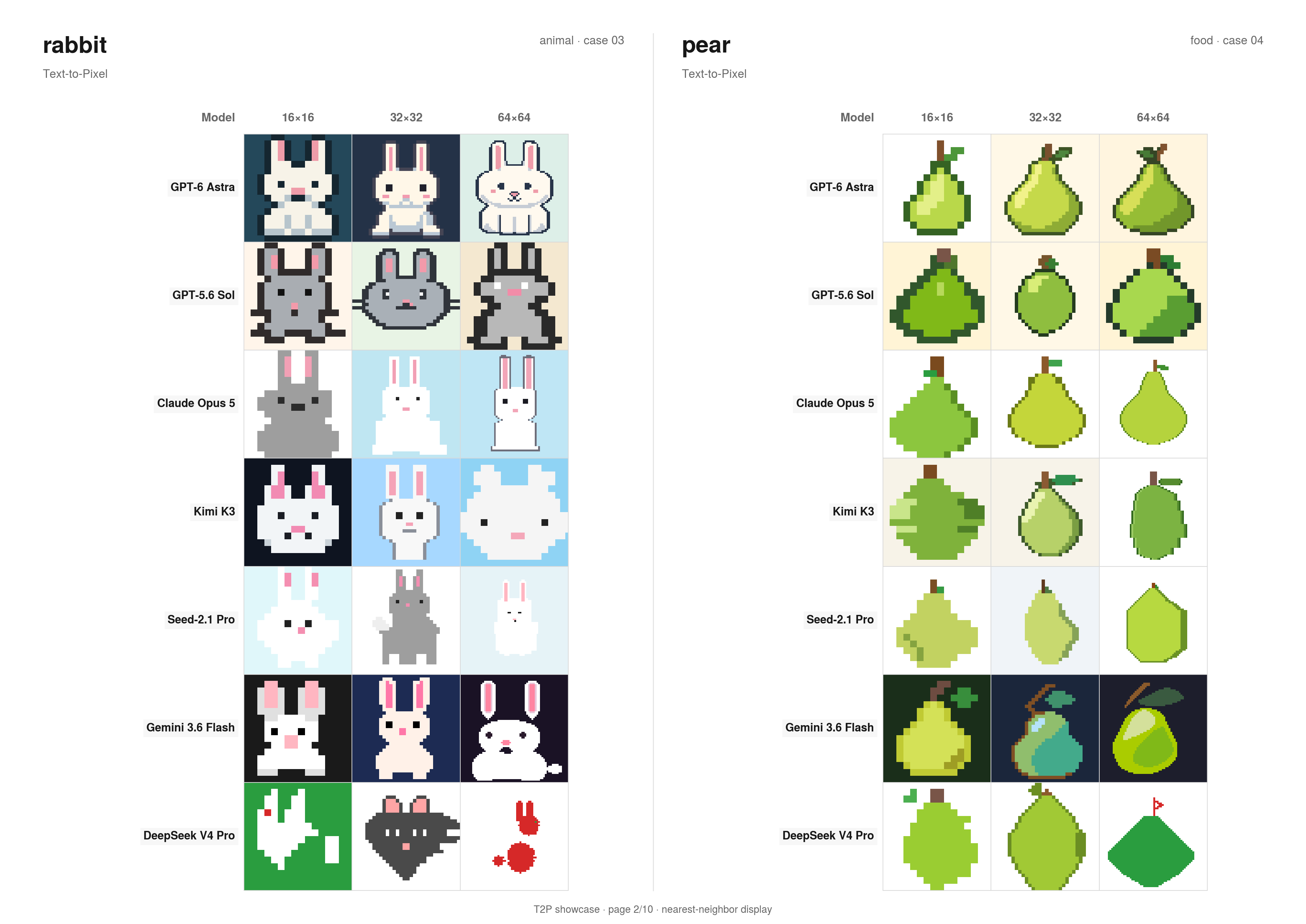}
\nativecasepair{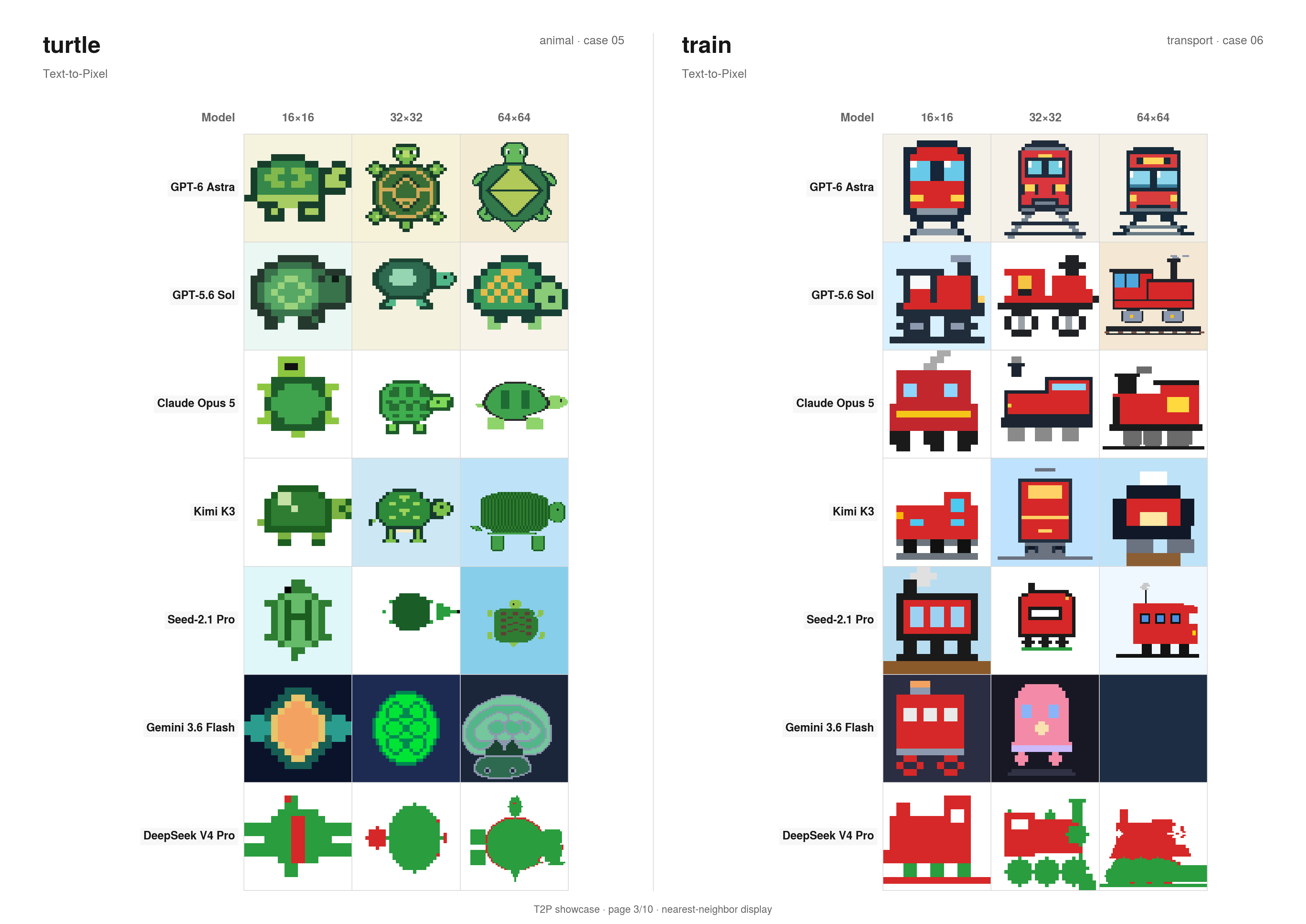}{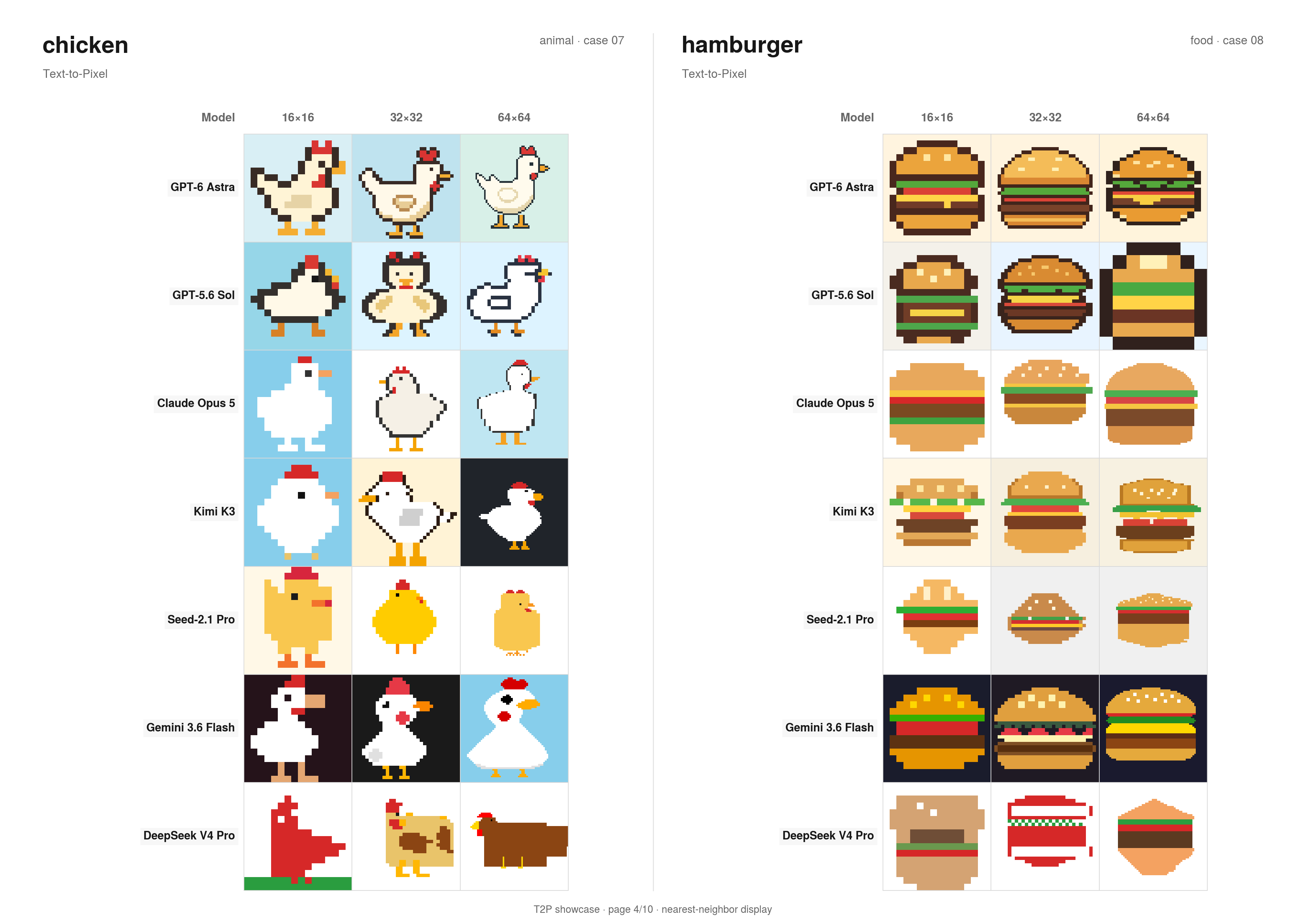}
\nativecasepair{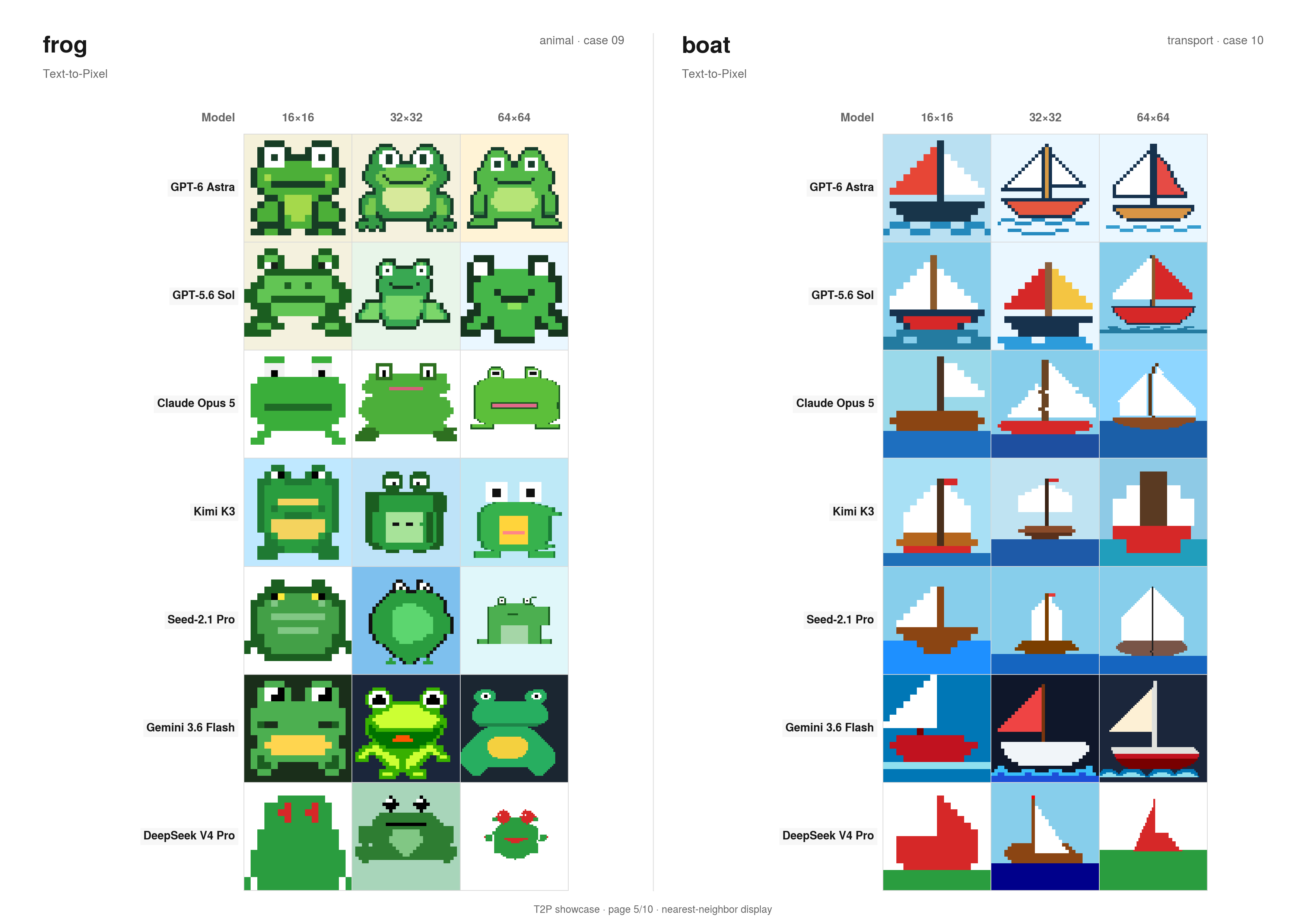}{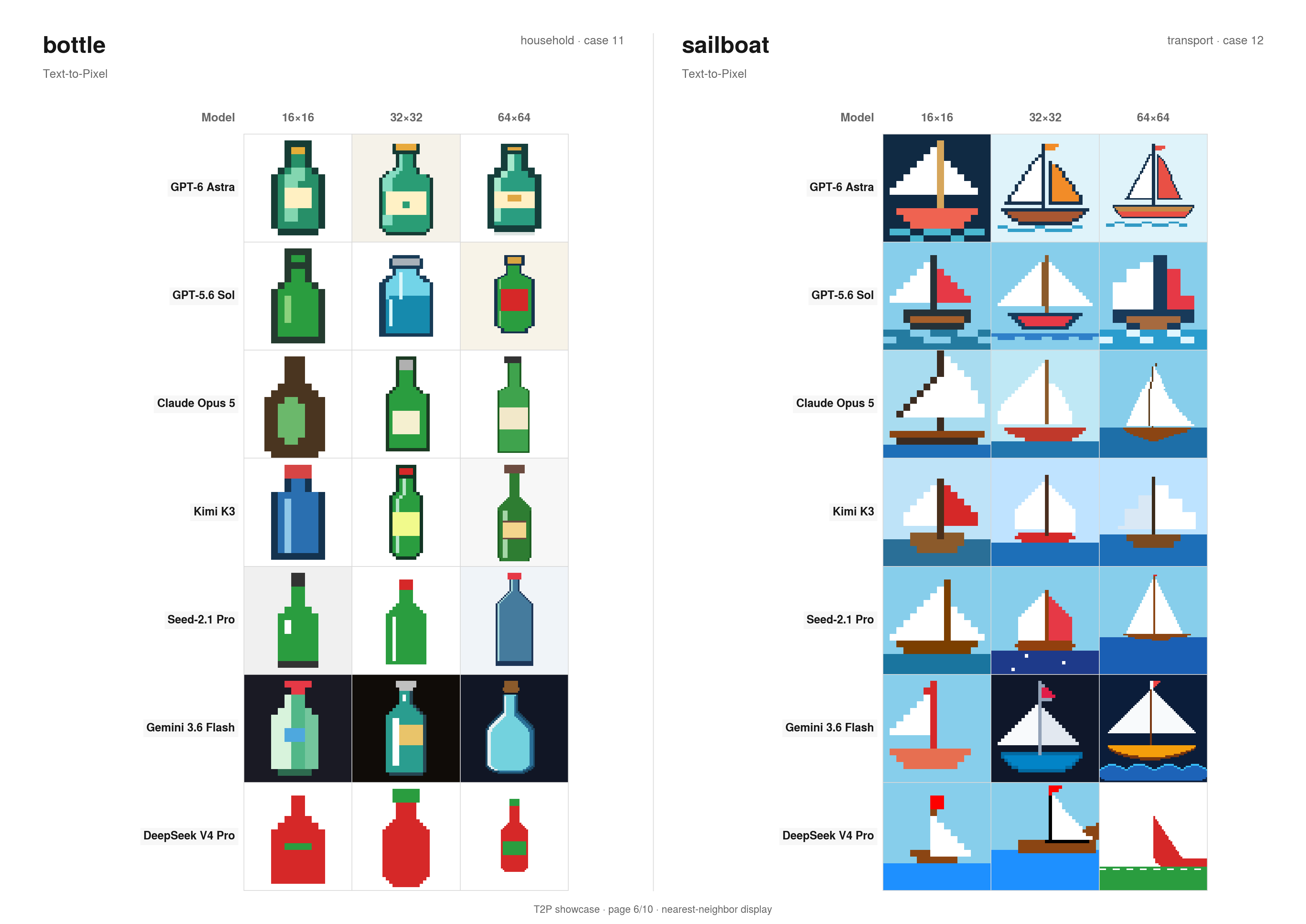}
\nativecasepair{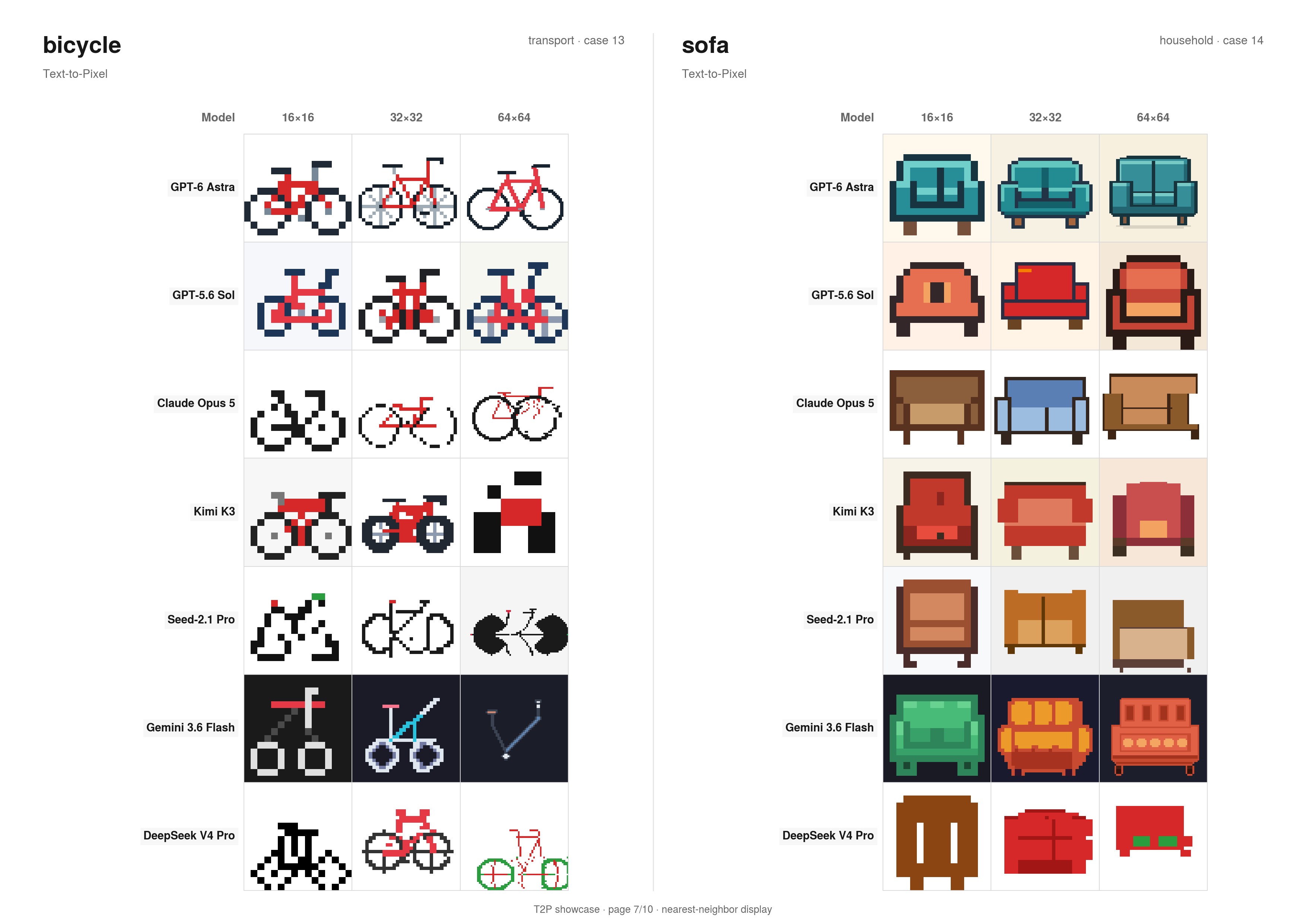}{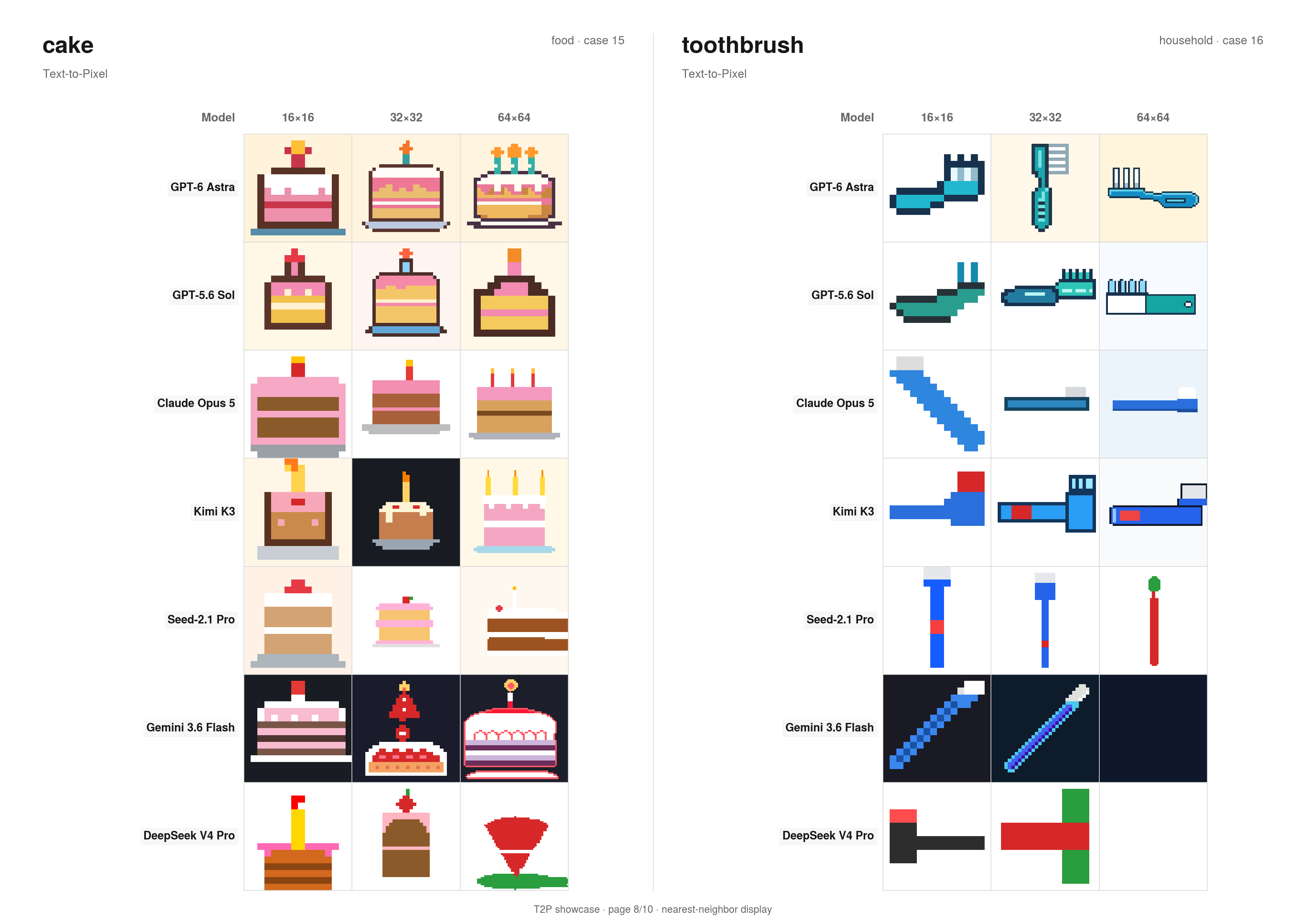}
\nativecasepair{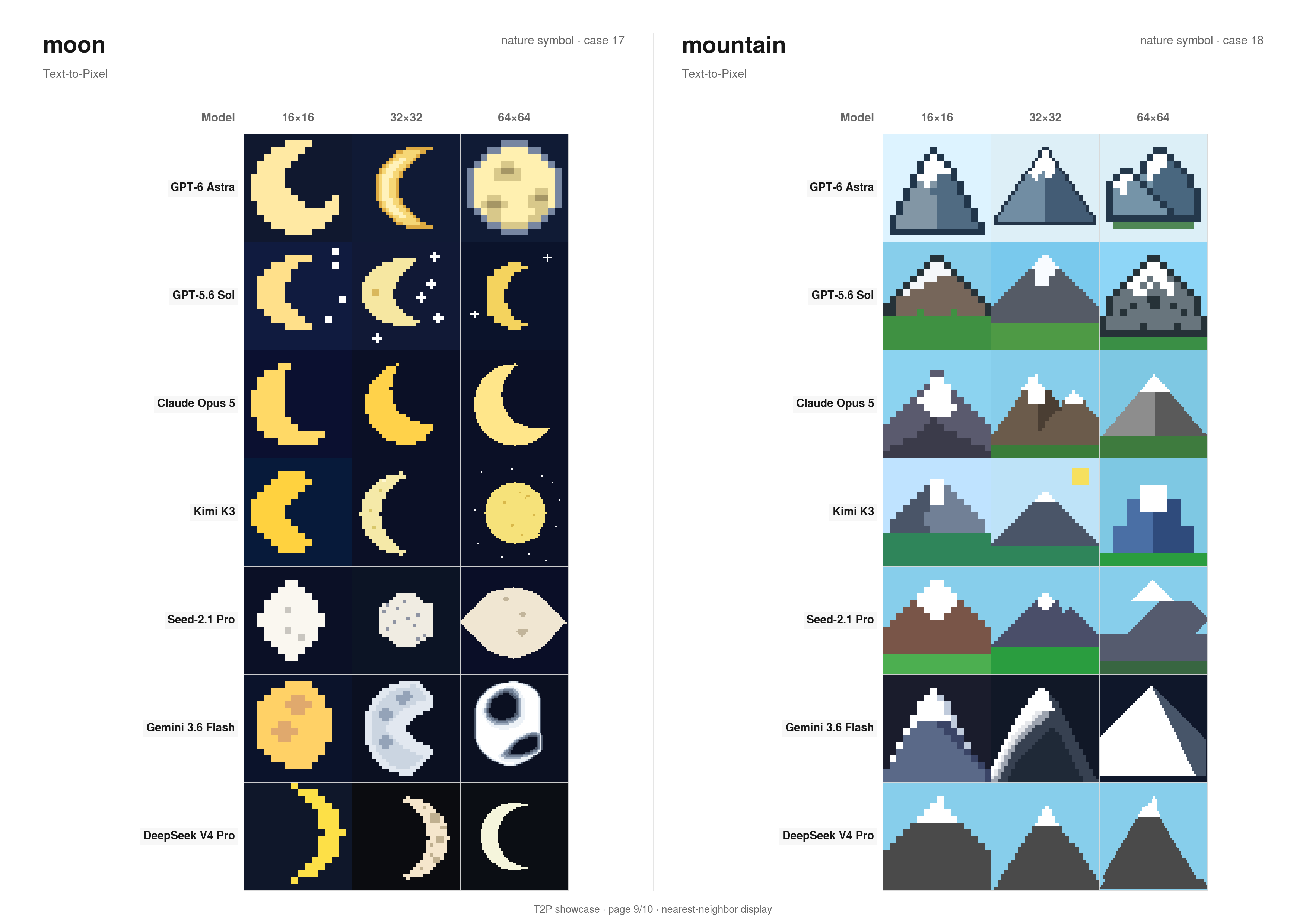}{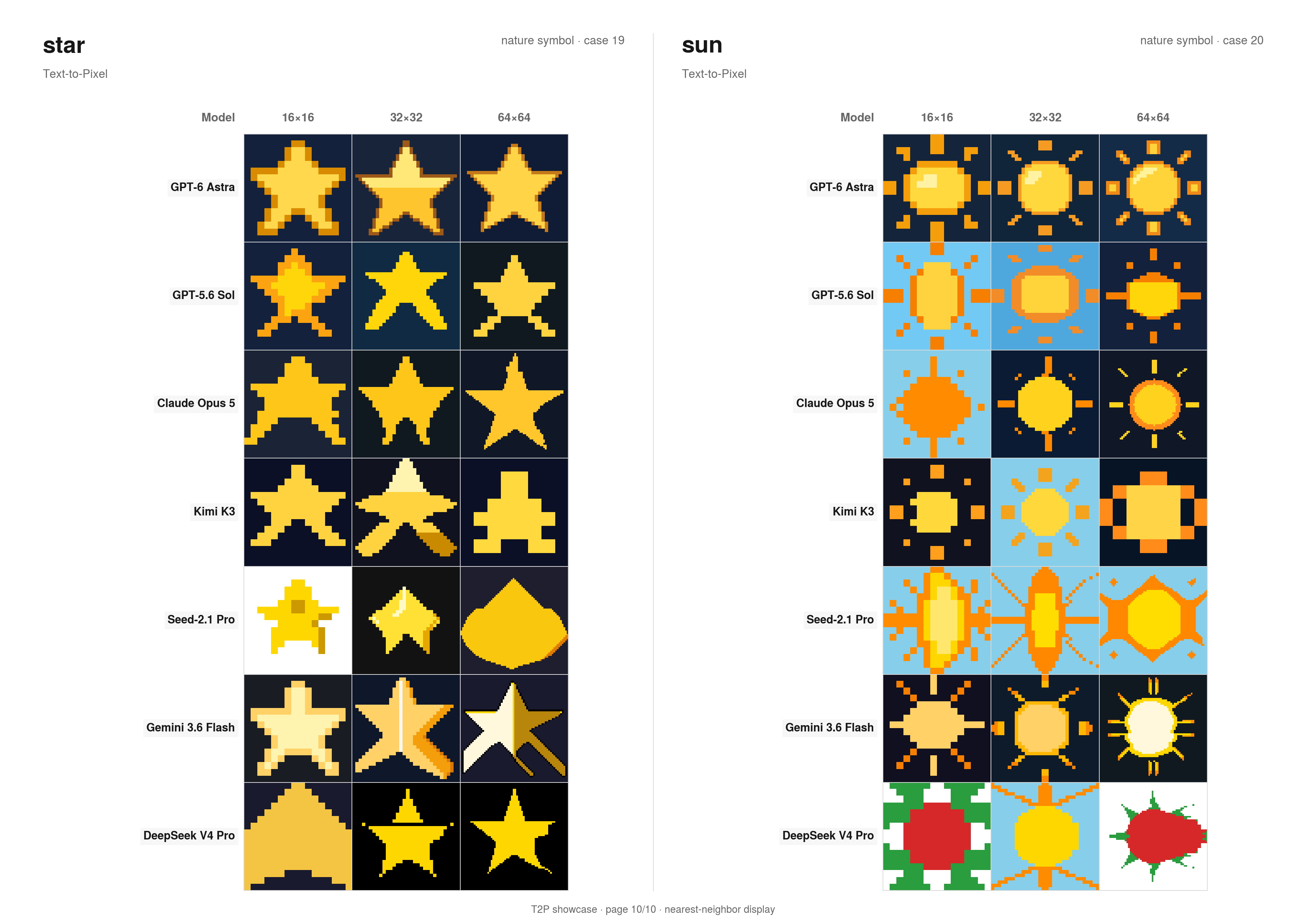}
\nativecasepair{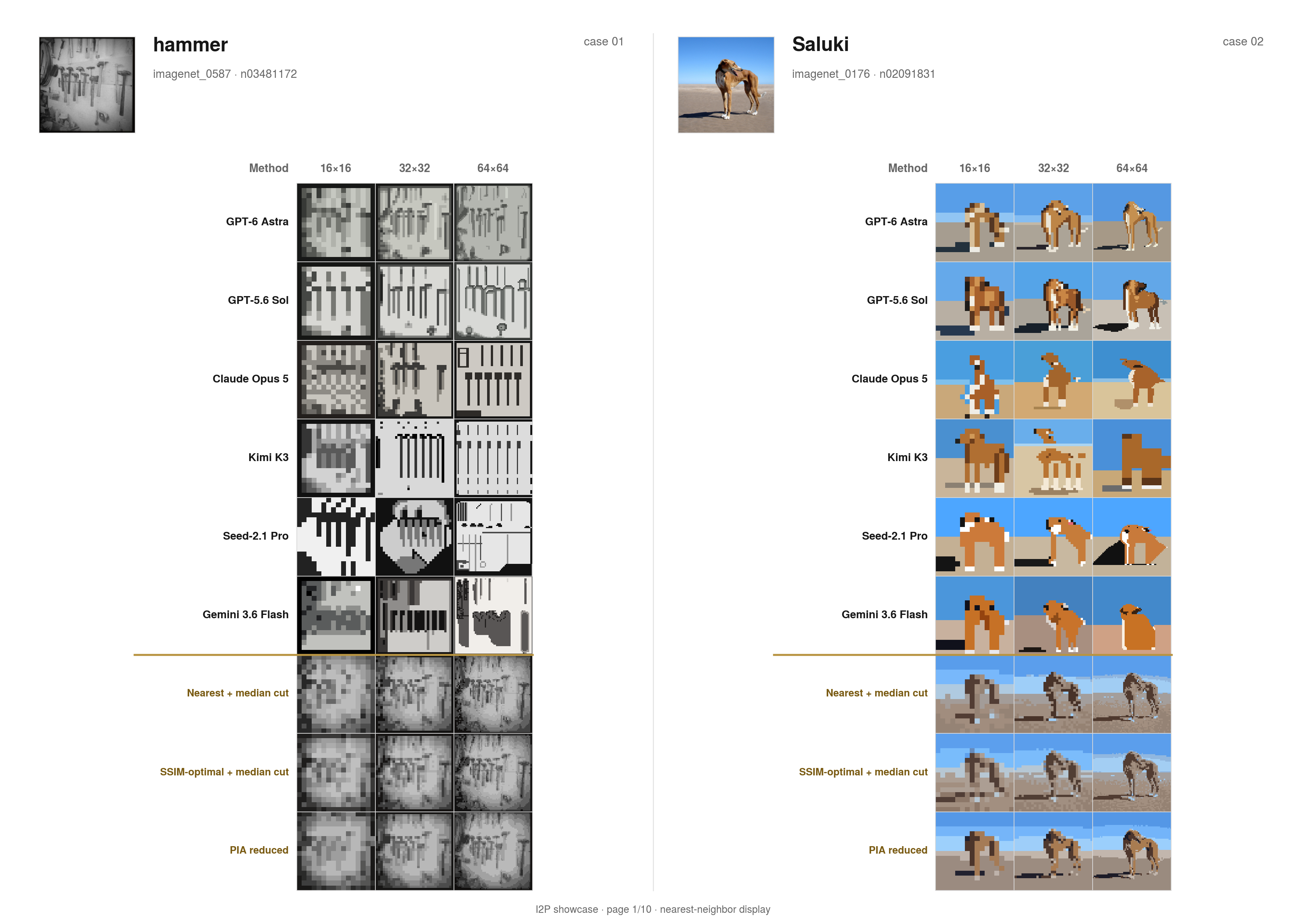}{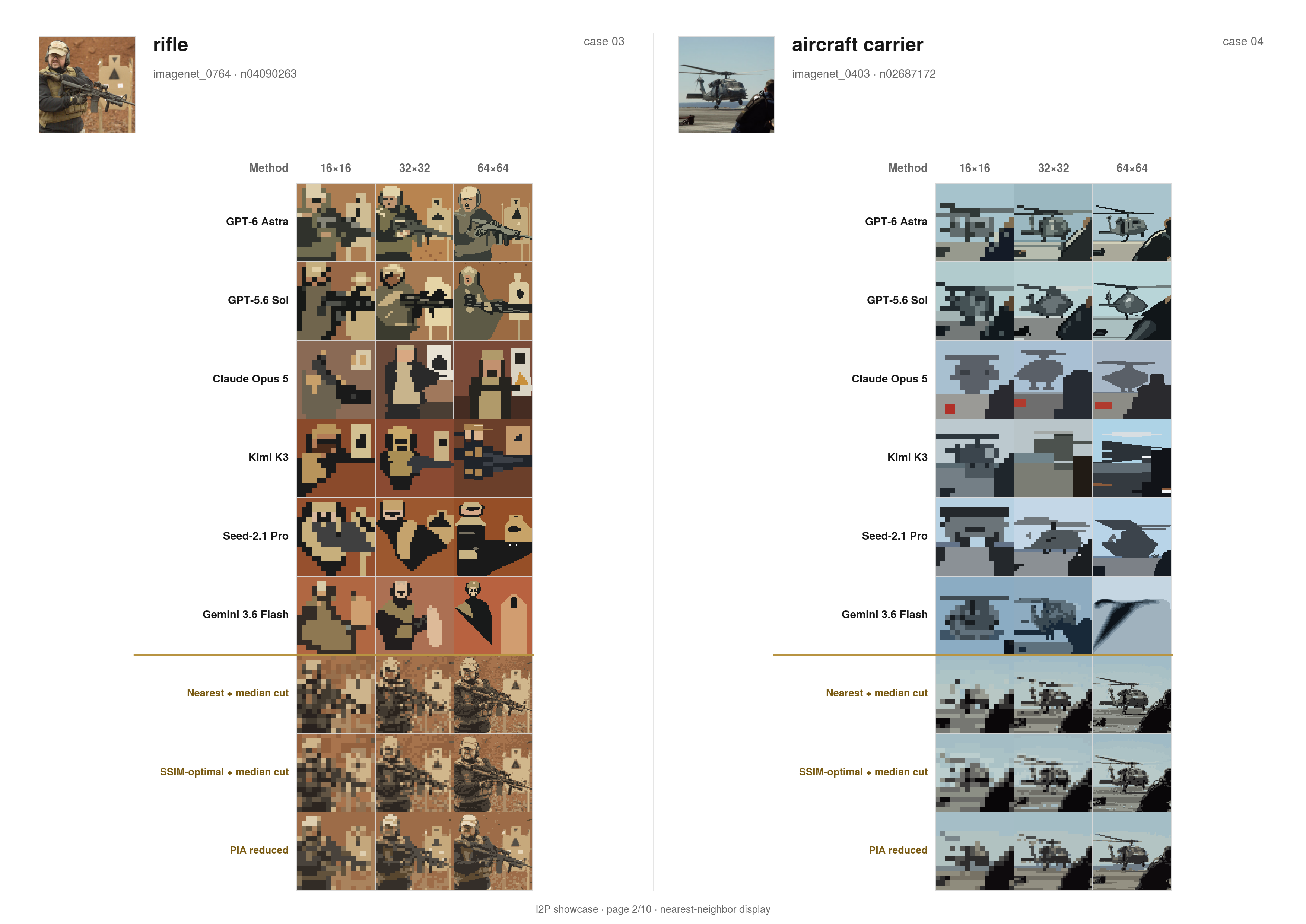}
\nativecasepair{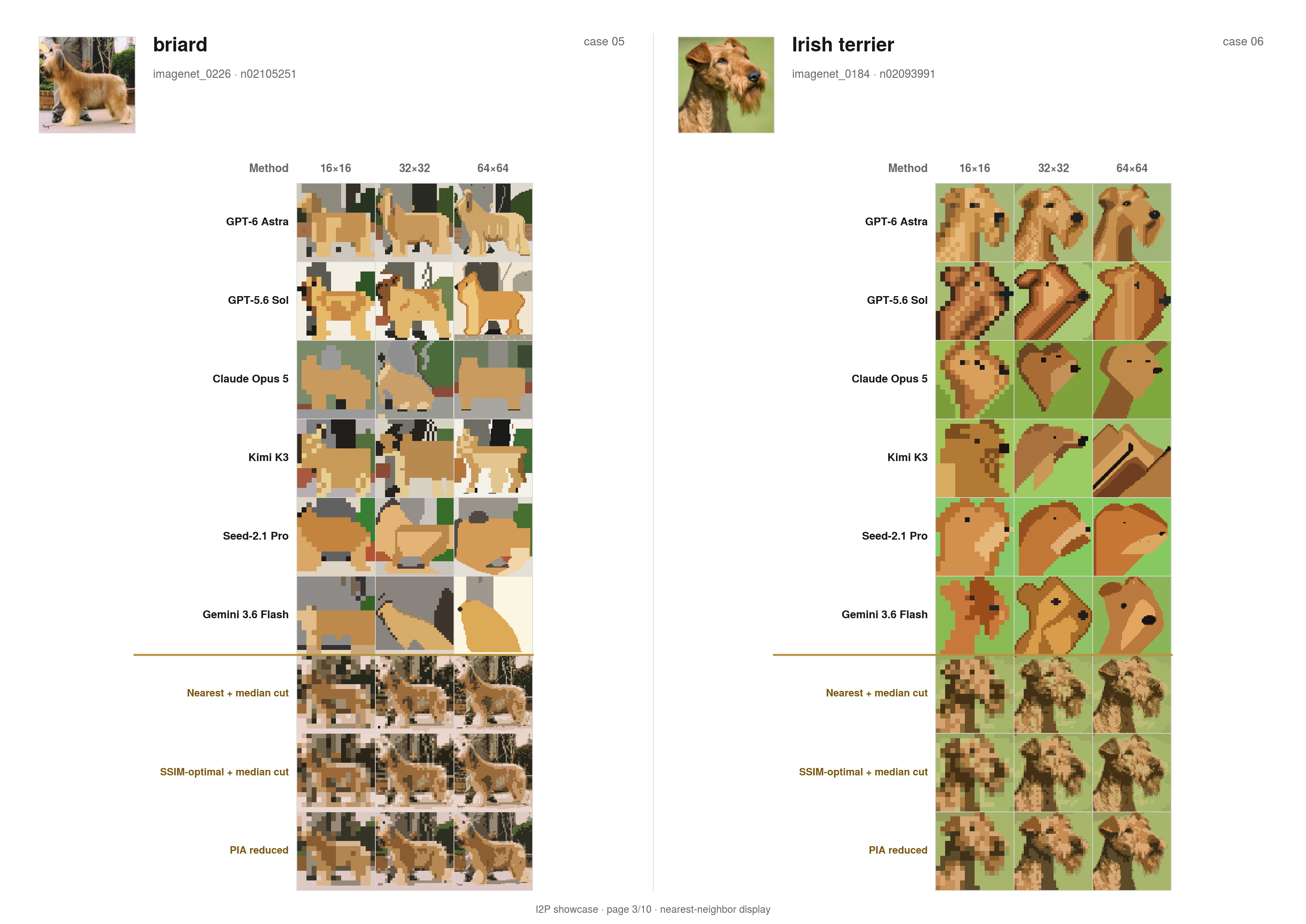}{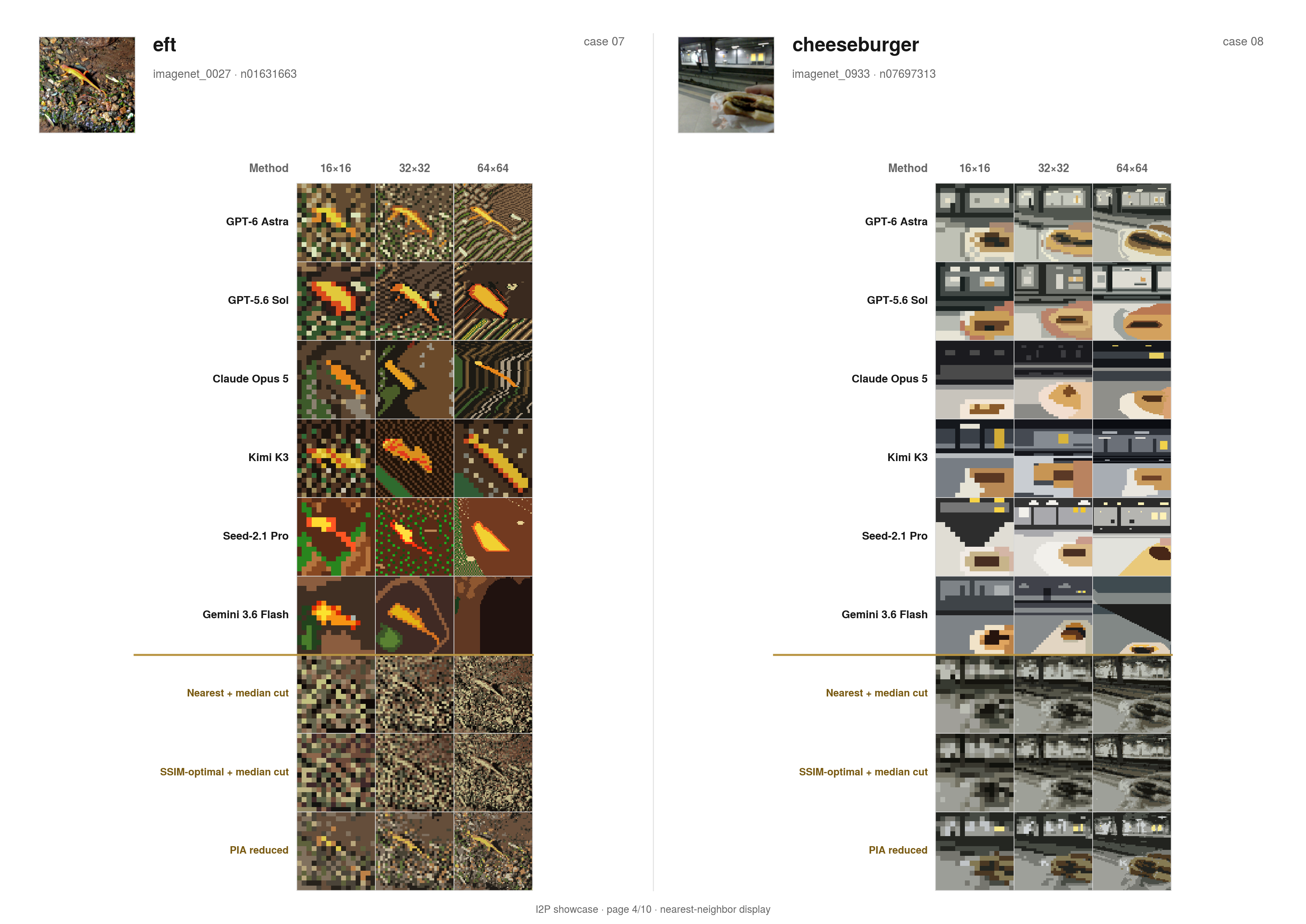}
\nativecasepair{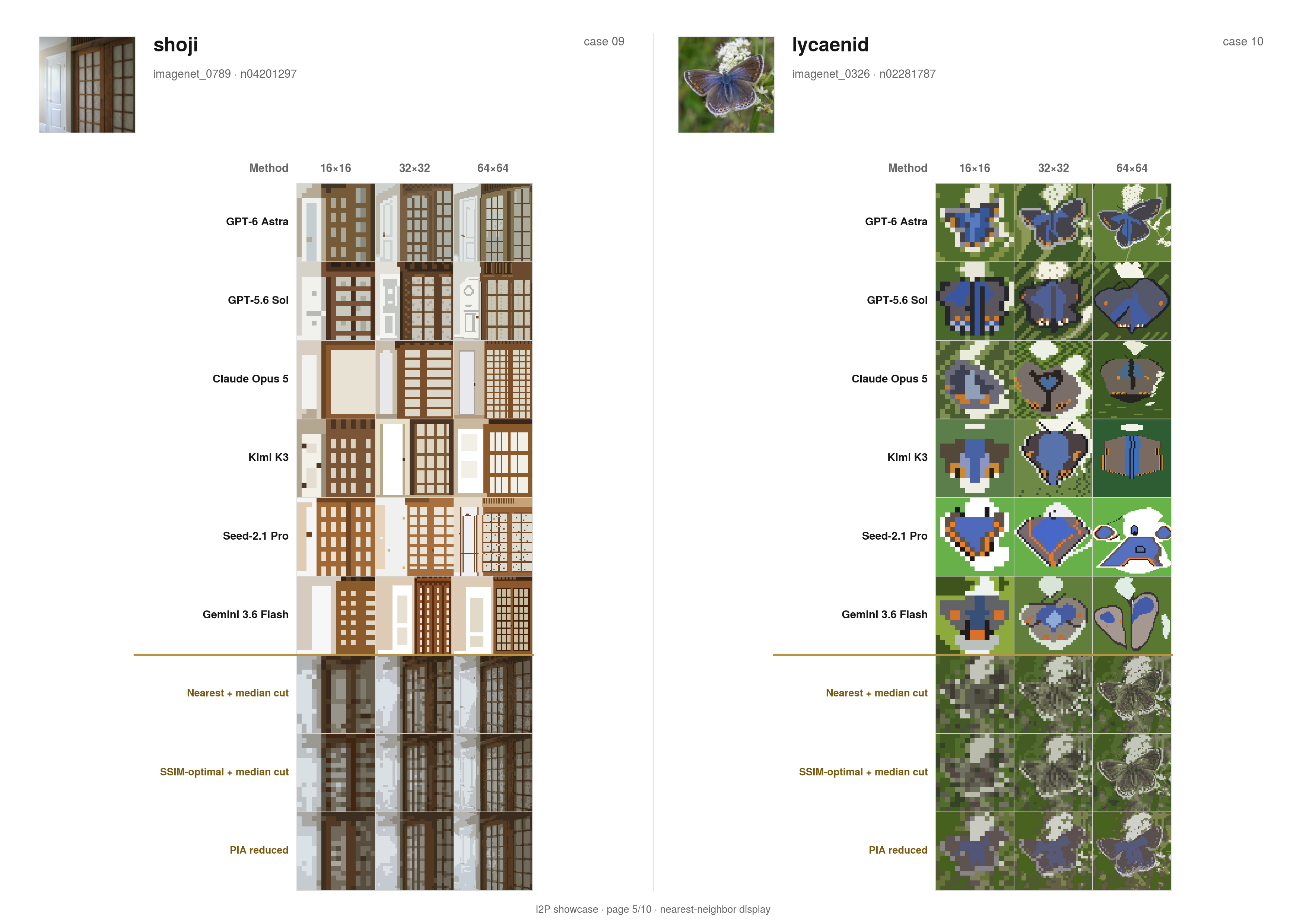}{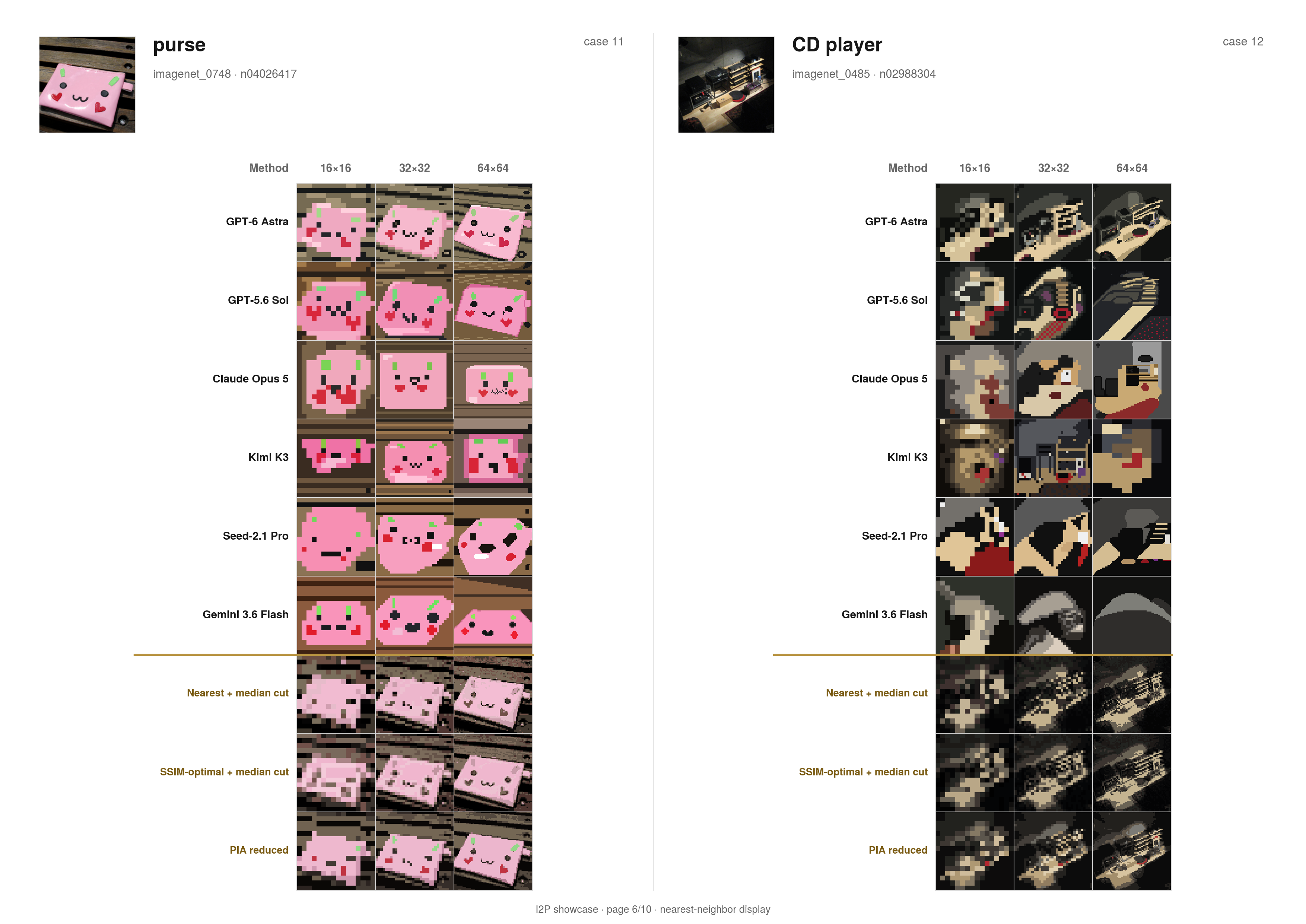}
\nativecasepair{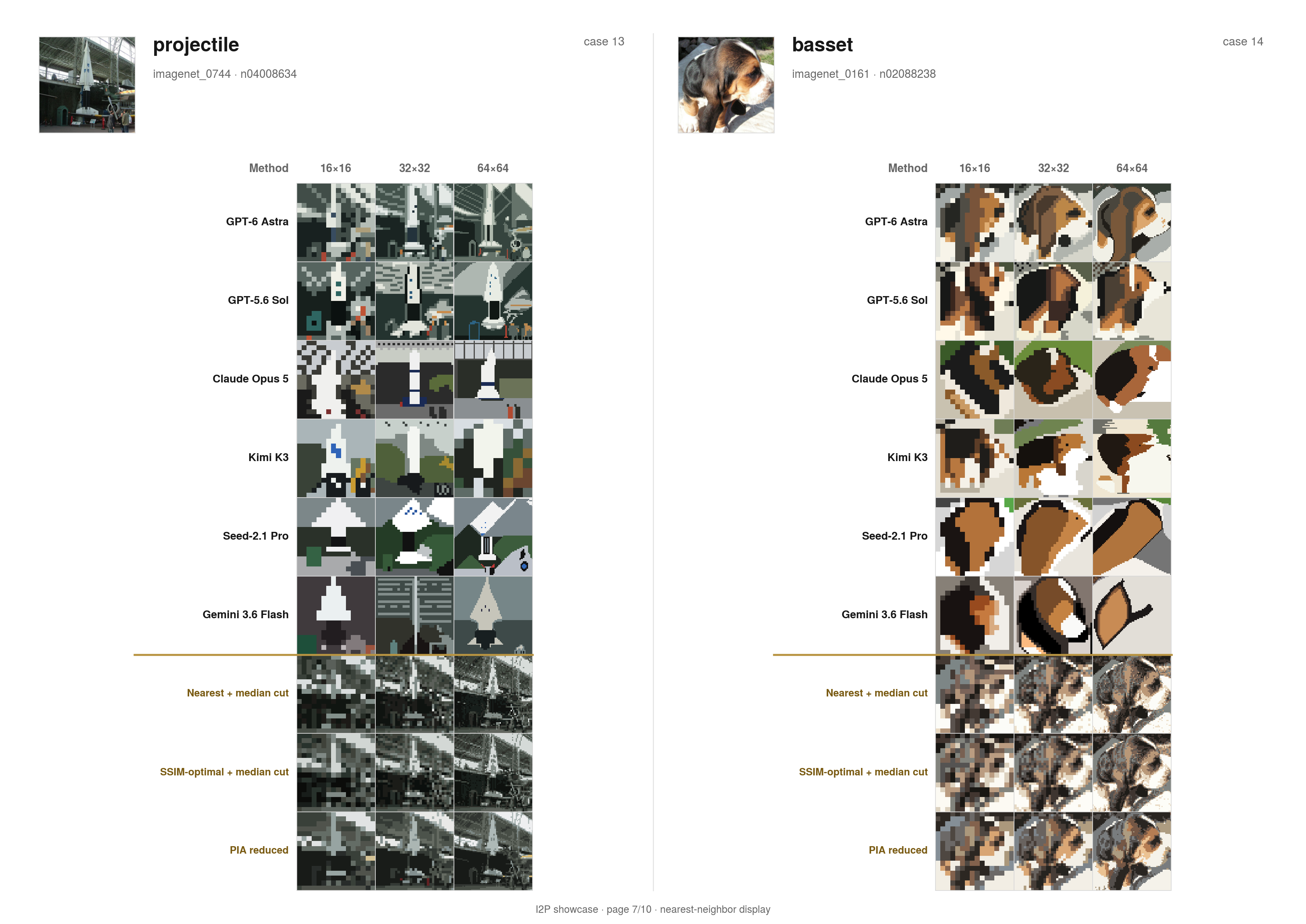}{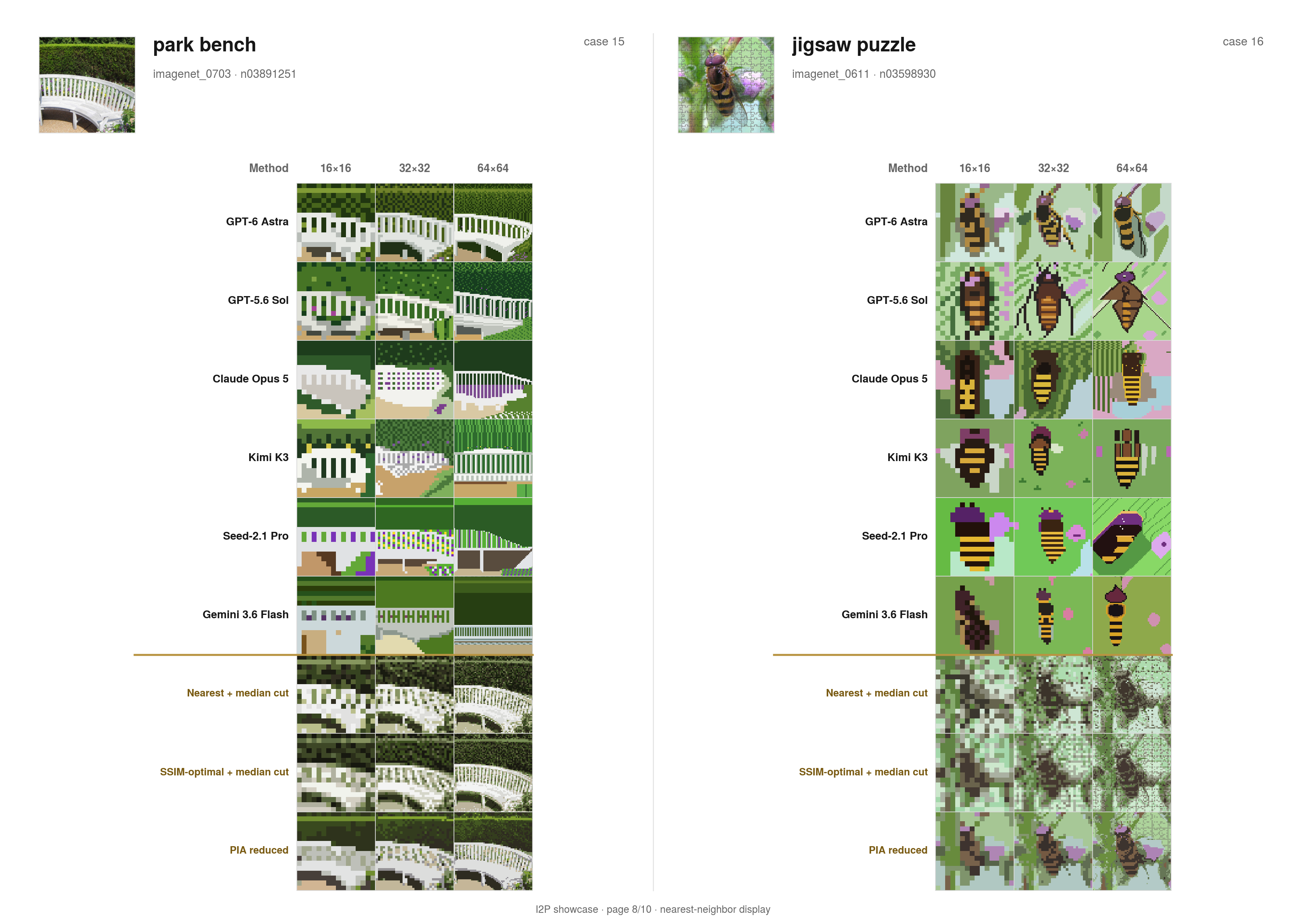}
\nativecasepair{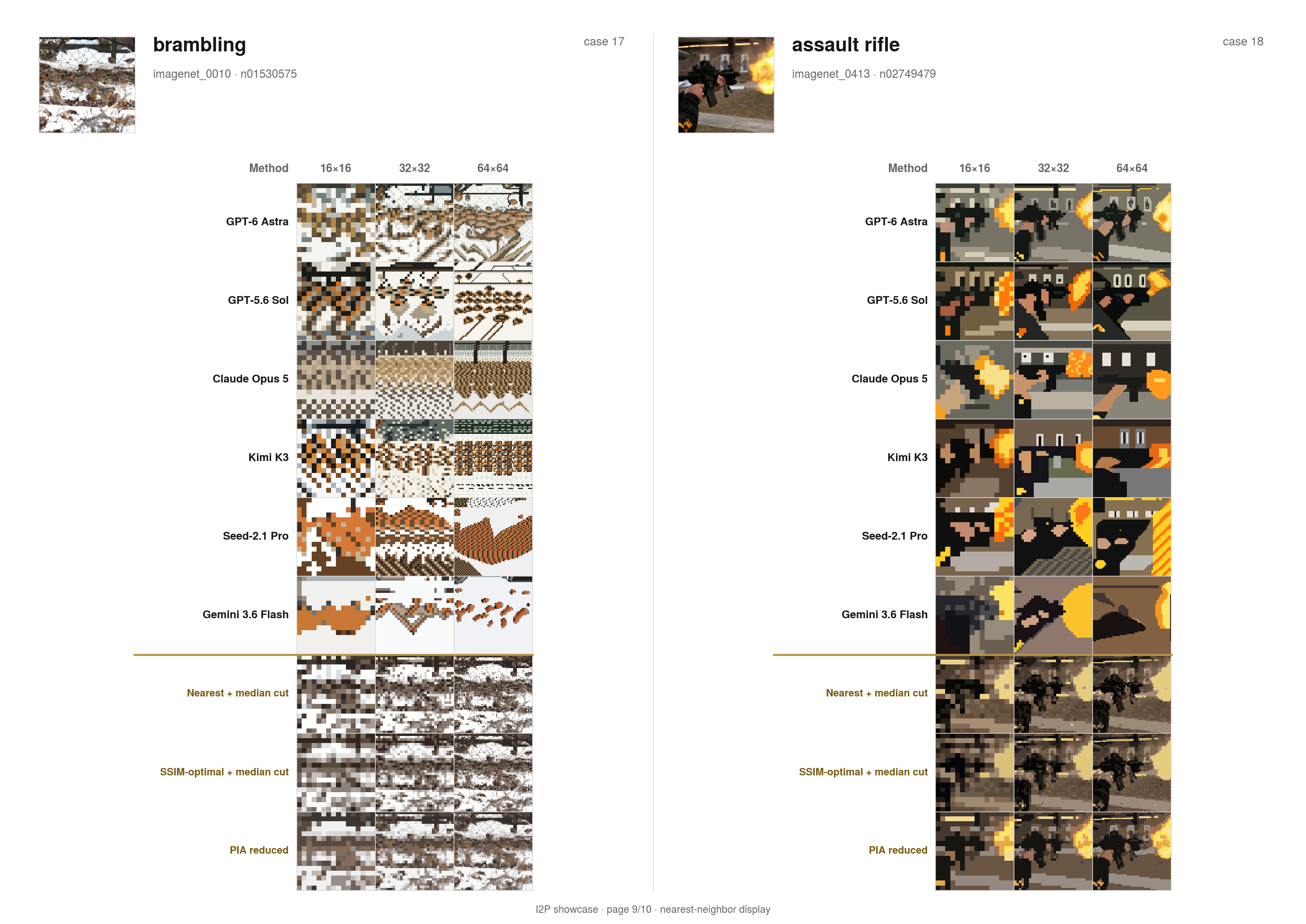}{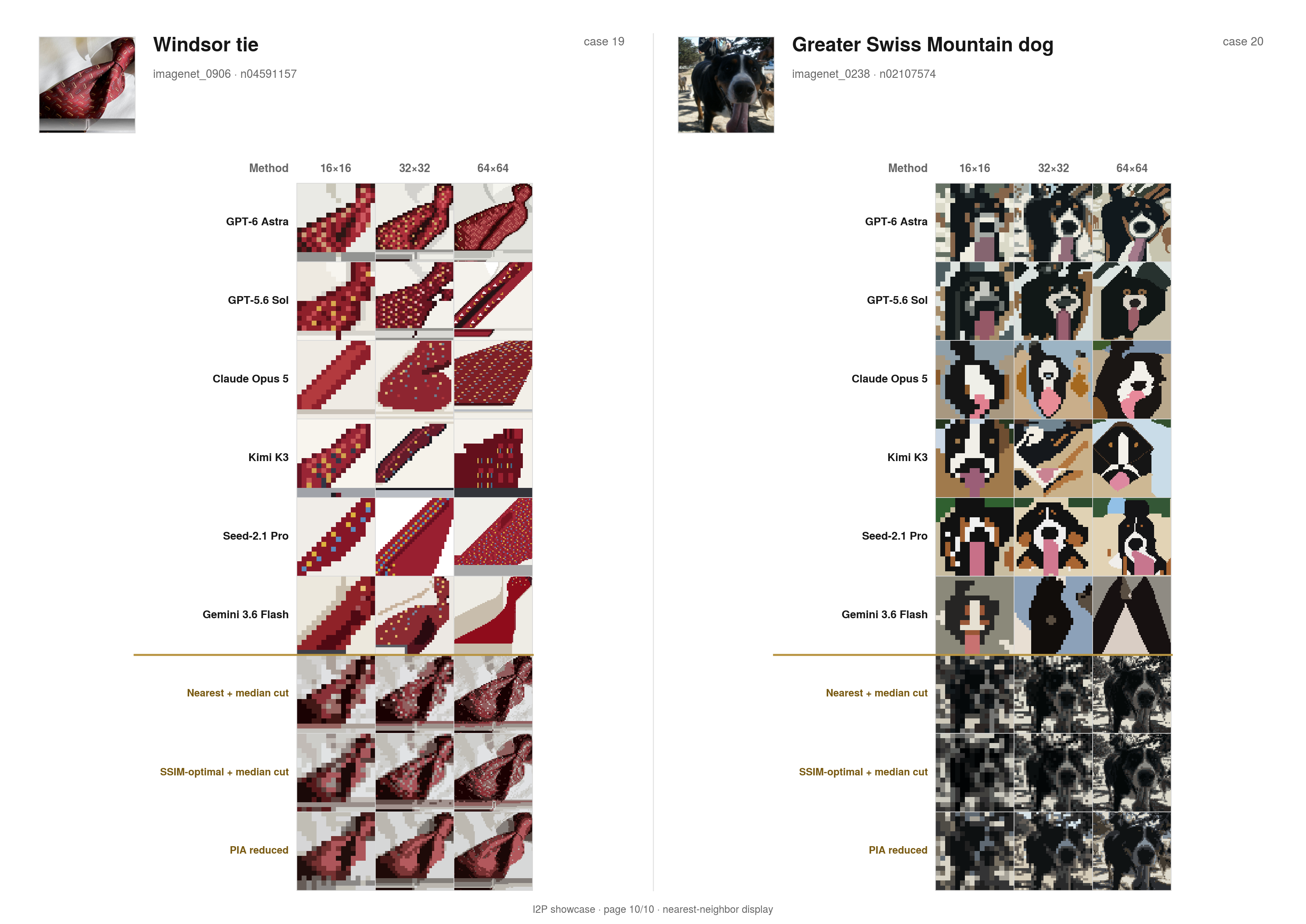}
\clearpage

\section{Additional Pixel Reconstruction Tuning Results}
\label{app:prt-additional-results}

\subsection{Molmo2 Pixel-Trajectory Construction}
\label{app:molmo-construction}

\textbf{Image pool and filtering.} Pixel trajectories
use PixMo-Cap images with human-written dense captions. We first construct a
20k-image pool and expand it with previously unused images for the 4k and 8k
sets. Exact hash matches to evaluation images are removed. We reject examples
whose captions indicate screenshots, webpages, user interfaces, documents,
posters, text, charts, diagrams, infographics, collages, or multi-panel
layouts because text-heavy and composite content cannot be faithfully
represented at $16^2$-$32^2$. Because PIA-reduced converts every source to a
square grid, we also reject images with aspect ratio above 1.33 to avoid
substantial geometric distortion.

\textbf{Recognizability and grid construction.} For each
candidate, we construct $16^2$ and $32^2$ eight-color grids using the
PIA-reduced procedure (Appendix~\ref{app:baselines}). We nearest-neighbor
upscale each grid and compute its paired cosine similarity with the source
image using SigLIP2-base-patch16-224. At each resolution, candidates are
ranked by this similarity; we retain the top 40\% at $16^2$ and the top 80\%
at $32^2$.
Images are unique within every set. The 2k set is balanced across $16^2$ and
$32^2$; 4k and 8k are nested extensions whose new rows use previously unused
images.

\textbf{Natural-language trajectory.} Gemini 3.8 Flash
receives the original image, upscaled ground-truth grid, one mask visualization
per palette letter, the PixMo caption, and measured palette/layout hints. At
temperature 0.4 and thinking budget 2,048, it writes only the caption and
palette explanation; the palette and grid remain deterministic. The response
must discuss every palette letter exactly once and in order, otherwise it is
retried up to three times. Three paraphrased user prompts are assigned
deterministically. Median target lengths are 1,693 characters at $16^2$ and
2,739 at $32^2$.

\subsection{Molmo2 Training and Evaluation}
\label{app:molmo-training}

\begin{table}[H]
\centering
\setpositionedtablecaption{\tablecaptionfont\textbf{Molmo2
post-training setup.} Both conditions start from the same released
Molmo2-4B image-pretraining checkpoint.}
{tab:molmo-training-setup}
\scriptsize
\setlength{\tabcolsep}{4pt}
\begin{tabular*}{\linewidth}{@{\extracolsep{\fill}}lp{0.73\linewidth}@{}}
\toprule
\textbf{Item} & \textbf{Setting} \\
\midrule
Initialization & Molmo2-4B-Pretrain, step 32,000; Qwen3-4B language model \\
OV mixture & 500,000 single-image rows from LLaVA-OneVision-1.5-Instruct;
answers over 2,048 tokens removed; image long side capped at 1,024 px \\
Pixel mixture & OV 500k plus 1k, 2k, 4k, or 8k unique Pixel trajectories \\
Schedule & One pass; OV 1,190 steps; OV$+$Nk uses $1{,}190+4N$ steps \\
Batching & 128 packed sequences $\times$ 4,096 tokens; one image per sample;
maximum answer length 1,800 tokens \\
Optimization & AdamW; LLM LR $10^{-5}$; ViT/connector LR $5{\times}10^{-6}$;
60-step warmup; final LR multiplier 0.1; gradient clip 1.0; bf16 \\
Hardware & Four nodes $\times$ eight NVIDIA A100-80GB GPUs per run \\
Seeds & Four matched seed pairs for OV and $+$2k; two seeds for 1k, 4k, and 8k \\
\bottomrule
\end{tabular*}
\placepositionedtablecaptionbottom
\end{table}

The primary suite comprises MMVP \citep{tong2024eyes}, BlindTest
\citep{rahmanzadehgervi2024vision}, V-STAR \citep{wu2023vstar}, BabyVision
\citep{chen2026babyvision}, HallusionBench \citep{guan2024hallusionbench},
PerceptionBench \citep{lin2026perceptionbench}, CV-Bench
\citep{tong2024cambrian}, MMStar \citep{chen2024mmstar}, Q-Bench
\citep{wu2024qbench}, MMBench \citep{liu2024mmbench}, VLMs Are Biased
\citep{vo2026biased}, BLINK \citep{fu2024blink}, ColorBench and
ColorBench-Robust \citep{liang2025colorbench}, ChartQA
\citep{masry2022chartqa}, and MathVista \citep{lu2024mathvista}.

All runs use fixed question subsets, identical prompts, and greedy decoding.
Rule-based scoring is used except for the free-form portion of BabyVision and
PerceptionBench, which use their benchmark-specific LLM judging protocols.

\begin{table}[H]
\centering
\setpositionedtablecaption{\tablecaptionfont\textbf{Molmo2
per-seed aggregate results.} Example pools the 24,850 official evaluation
units from the primary suite; Benchmark weights its 16 benchmarks equally.}
{tab:molmo-seed-results}
\scriptsize
\setlength{\tabcolsep}{7pt}
\begin{tabular}{lrr}
\toprule
\textbf{Run} & \textbf{Example avg.} & \textbf{Benchmark avg.} \\
\midrule
OV-only s0 & 54.1 & 54.3 \\
OV-only s1 & 54.7 & 54.4 \\
OV-only s2 & 53.8 & 53.6 \\
OV-only s3 & 54.2 & 54.2 \\
\arrayrulecolor{black!50}\midrule\arrayrulecolor{black}
OV$+$2k s0 & 55.1 & 55.1 \\
OV$+$2k s1 & 55.0 & 54.9 \\
OV$+$2k s2 & 54.9 & 54.3 \\
OV$+$2k s3 & 54.3 & 54.0 \\
\bottomrule
\end{tabular}
\placepositionedtablecaptionbottom
\end{table}

\begin{table}[H]
\centering
\setpositionedtablecaption{\tablecaptionfont\textbf{Molmo2
benchmark-level Pixel-quantity deltas.} Values are changes from the four-seed
OV-only mean. The 2k column averages four seeds; 1k, 4k, and 8k average two
seeds each; All Pixel pools ten runs.}
{tab:molmo-quantity-full}
\scriptsize
\setlength{\tabcolsep}{5pt}
\begin{tabular*}{0.86\linewidth}{@{\extracolsep{\fill}}lrrrrr@{}}
\toprule
\textbf{Benchmark} & $\boldsymbol{\Delta}_{\mathbf{1k}}$ &
$\boldsymbol{\Delta}_{\mathbf{2k}}$ & $\boldsymbol{\Delta}_{\mathbf{4k}}$ &
$\boldsymbol{\Delta}_{\mathbf{8k}}$ & \textbf{All Pixel} \\
\midrule
CV-Bench          & +1.8 & +0.9 & +1.5 & +1.0 & +1.2 \\
ColorBench-Robust & +1.3 & +0.8 & +1.5 & +0.4 & +1.0 \\
VLMs Are Biased   & +0.6 & +1.1 & +0.2 & +1.7 & +1.0 \\
MathVista         & +0.2 & +0.5 & +1.5 & +1.3 & +0.8 \\
MMStar            & +0.5 & +0.8 & +1.0 & +0.8 & +0.8 \\
HallusionBench    & +0.1 & +0.2 & +2.2 & +0.5 & +0.7 \\
BabyVision        & +1.3 & $-0.1$ & +0.3 & +1.0 & +0.5 \\
MMVP              & +2.0 & +0.2 & $-0.7$ & +0.7 & +0.5 \\
BlindTest         & +0.2 & +1.4 & $-0.5$ & $-0.3$ & +0.5 \\
Q-Bench           & +0.6 & +0.3 & +0.2 & +0.4 & +0.4 \\
ColorBench        & +0.4 & +0.4 & +0.3 & +0.2 & +0.3 \\
BLINK             & $-1.3$ & +0.8 & +0.3 & +0.5 & +0.2 \\
ChartQA           & +0.5 & +0.0 & $-0.1$ & +0.2 & +0.1 \\
MMBench           & +0.2 & +0.0 & $-0.2$ & +0.3 & +0.1 \\
PerceptionBench   & $-0.0$ & +0.0 & $-0.1$ & $-0.1$ & $-0.0$ \\
V-STAR Bench      & +0.5 & $-0.1$ & $-1.0$ & $-0.5$ & $-0.3$ \\
\midrule
\textbf{Example avg.}   & +0.4 & +0.6 & +0.4 & +0.5 & +0.5 \\
\textbf{Benchmark avg.} & +0.6 & +0.5 & +0.4 & +0.5 & +0.5 \\
\bottomrule
\end{tabular*}
\placepositionedtablecaptionbottom
\end{table}

\subsection{Qwen3-VL Training Setup}
\label{app:qwen-training}

\textbf{Data composition.} All PRT experiments initialize
from Qwen3-VL-Instruct checkpoints at the 2B, 4B, and 8B scales; tables
abbreviate these scales as 2B, 4B, and 8B. These
experiments use an additive design.
The FV-only baseline uses 64,000 FineVision occurrences. Each
FV$+$Pixel configuration retains those occurrences and adds 3,000
pixel-reconstruction occurrences mixed
approximately evenly across $16^2$, $32^2$, and $64^2$, for
67,000 total training examples. All examples are jointly shuffled.
FineVision responses are capped at 2,048 tokens, and pixel
targets use the deterministic eight-color PIA-reduced conversion.
Pixel source images are sampled without replacement. In the Pixel-Mix run,
each of the 3,000 unique source images is assigned to exactly one of
$16^2$, $32^2$, or $64^2$. The single-resolution ablations deliberately reuse
the same 3,000 source-image IDs across separate runs, converting each image
once at the corresponding run's target resolution; no run therefore repeats a source image
because of the resolution construction.

\textbf{Optimization.} We fully fine-tune all parameters, including the vision
transformer, using AdamW ($\beta_1=0.9$, $\beta_2=0.999$,
$\epsilon=10^{-8}$), zero weight decay, a peak learning rate of $10^{-5}$
with cosine decay and 3\% warmup, gradient clipping at 1.0, bf16 precision,
and training seed 42. The effective batch size is 64. FV-only trains for 1,000
optimizer steps, whereas FV$+$Pixel trains for 1,047 steps. Consequently, the
additive delta measures the combined effect of adding Grid data and training
for 47 additional steps.

\begin{table}[H]
\centering
\setpositionedtablecaption{\tablecaptionfont\textbf{Qwen3-VL
post-training as a regularization study.} Base is the released Instruct
checkpoint; FV is FineVision-only post-training; FV$+$Pixel retains all 64k FV
examples and adds 3k reconstruction examples. $\Delta$ is FV$+$Pixel minus
FV.}
{tab:qwen-regularization}
\scriptsize
\setlength{\tabcolsep}{6pt}
\begin{tabular*}{0.82\linewidth}{@{\extracolsep{\fill}}clrrrr@{}}
\toprule
\textbf{Model} & \textbf{Avg.} & \textbf{Base} & \textbf{FV} &
\textbf{FV$+$Pixel} & \textbf{$\Delta$} \\
\midrule
\multirow{2}{*}{2B} & Example   & 43.0 & 48.1 & 50.1 & \cellcolor{deltapos!18}+2.0 \\
                    & Benchmark & 45.2 & 47.3 & 50.3 & \cellcolor{deltapos!25}+3.0 \\
\arrayrulecolor{black!50}\midrule\arrayrulecolor{black}
\multirow{2}{*}{4B} & Example   & 48.8 & 53.9 & 56.8 & \cellcolor{deltapos!24}+2.9 \\
                    & Benchmark & 51.5 & 53.6 & 57.3 & \cellcolor{deltapos!30}+3.7 \\
\arrayrulecolor{black!50}\midrule\arrayrulecolor{black}
\multirow{2}{*}{8B} & Example   & 48.8 & 55.1 & 57.2 & \cellcolor{deltapos!18}+2.1 \\
                    & Benchmark & 53.6 & 56.3 & 57.9 & \cellcolor{deltapos!14}+1.6 \\
\bottomrule
\end{tabular*}
\placepositionedtablecaptionbottom
\end{table}

\subsection{Additive FV + Pixel Results}
\label{app:qwen-additive}

Table~\ref{tab:prt-resolution-ablation} summarizes benchmark-level
target-resolution deltas across model scales. Each panel is independently
ordered by the mixed-resolution delta, from the largest gain to the largest
drop.

\begin{table}[H]
\centering
\setpositionedtablecaption{\tablecaptionfont\textbf{Benchmark-level target-resolution deltas across
model scales.} Each entry is the accuracy change, in percentage points,
relative to the corresponding FineVision-only baseline under the additive
protocol. Mix distributes Grid examples approximately evenly across the three
resolutions. Panels are independently sorted by $\Delta_{\mathrm{Mix}}$.
Example averaging assigns equal weight to every official evaluation unit;
Benchmark averaging assigns equal weight to every benchmark. * denotes a randomly
sampled subset.}
{tab:prt-resolution-ablation}
\tiny
\setlength{\tabcolsep}{0.7pt}
\begin{minipage}[t]{0.323\linewidth}
\vspace{0pt}
\centering
\begin{tabular*}{\linewidth}{@{\extracolsep{\fill}}lccccc@{}}
\toprule
\textbf{2B} & $N$ & $\Delta_{16}$ & $\Delta_{32}$ & $\Delta_{64}$ & $\Delta_{\mathrm{Mix}}$ \\
\midrule
MMVP          &  150 & \cellcolor{white}+7.3 & \cellcolor{deltapos!30}+12.0 & \cellcolor{deltapos!13}+9.3 & \cellcolor{deltapos!26}+11.3 \\
BabyVision    &  388 & \cellcolor{deltapos!24}+4.6 & \cellcolor{deltapos!18}+2.8 & \cellcolor{deltapos!30}+7.0 & \cellcolor{deltapos!25}+4.9 \\
MMBench       & 1292 & \cellcolor{deltapos!30}+5.0 & \cellcolor{deltapos!24}+4.8 & \cellcolor{white}+4.2 & \cellcolor{deltapos!24}+4.8 \\
CV-Bench      & 2638 & \cellcolor{deltapos!20}+3.1 & \cellcolor{deltapos!21}+3.4 & \cellcolor{deltapos!21}+3.5 & \cellcolor{deltapos!19}+3.0 \\
TallyQA\subsetmark & 3000 & \cellcolor{deltapos!20}+3.3 & \cellcolor{deltapos!18}+2.8 & \cellcolor{deltapos!17}+2.4 & \cellcolor{deltapos!19}+2.9 \\
Hallusion     & 1129 & \cellcolor{deltapos!26}+5.1 & \cellcolor{deltapos!11}+1.2 & \cellcolor{deltapos!22}+3.8 & \cellcolor{deltapos!18}+2.8 \\
VStar         &  191 & \cellcolor{deltapos!10}+1.0 & \cellcolor{deltapos!20}+3.1 & \cellcolor{white}0.0 & \cellcolor{deltapos!18}+2.6 \\
Q-Bench       & 1495 & \cellcolor{deltapos!15}+1.8 & \cellcolor{deltapos!22}+3.7 & \cellcolor{deltapos!20}+3.1 & \cellcolor{deltapos!17}+2.1 \\
Perception    & 3000 & \cellcolor{deltapos!16}+2.0 & \cellcolor{deltapos!16}+2.0 & \cellcolor{deltapos!11}+1.2 & \cellcolor{deltapos!16}+1.9 \\
BlindTest     & 3434 & \cellcolor{deltapos!22}+3.6 & \cellcolor{deltapos!12}+1.3 & \cellcolor{deltapos!23}+3.9 & \cellcolor{deltapos!14}+1.7 \\
BLINK         & 1901 & \cellcolor{deltaneg!10}-0.7 & \cellcolor{deltapos!13}+1.5 & \cellcolor{deltapos!5}+0.3 & \cellcolor{deltapos!14}+1.6 \\
MMStar        & 1500 & \cellcolor{deltaneg!7}-0.5 & \cellcolor{deltapos!20}+3.3 & \cellcolor{deltapos!9}+0.9 & \cellcolor{deltapos!5}+0.3 \\
VLMBias       & 2784 & \cellcolor{deltaneg!14}-1.1 & \cellcolor{deltapos!16}+1.9 & \cellcolor{deltaneg!9}-0.6 & \cellcolor{deltaneg!10}-0.8 \\
\midrule
Example       & 22902 & \cellcolor{deltapos!8}+2.14 & \cellcolor{deltapos!30}+2.50 & \cellcolor{deltapos!16}+2.27 & \cellcolor{white}+2.01 \\
Benchmark     & 13 & \cellcolor{white}+2.66 & \cellcolor{deltapos!30}+3.37 & \cellcolor{deltapos!14}+2.99 & \cellcolor{deltapos!15}+3.02 \\
\bottomrule
\end{tabular*}
\end{minipage}\hfill
\begin{minipage}[t]{0.323\linewidth}
\vspace{0pt}
\centering
\begin{tabular*}{\linewidth}{@{\extracolsep{\fill}}lccccc@{}}
\toprule
\textbf{4B} & $N$ & $\Delta_{16}$ & $\Delta_{32}$ & $\Delta_{64}$ & $\Delta_{\mathrm{Mix}}$ \\
\midrule
MMVP          &  150 & \cellcolor{white}+18.0 & \cellcolor{deltapos!26}+22.7 & \cellcolor{deltapos!30}+23.3 & \cellcolor{deltapos!15}+20.7 \\
BlindTest     & 3434 & \cellcolor{deltapos!30}+10.7 & \cellcolor{deltapos!30}+10.6 & \cellcolor{deltapos!29}+7.3 & \cellcolor{deltapos!30}+9.6 \\
BabyVision    &  388 & \cellcolor{deltapos!18}+2.6 & \cellcolor{deltapos!23}+3.9 & \cellcolor{deltapos!22}+3.6 & \cellcolor{deltapos!26}+5.2 \\
CV-Bench      & 2638 & \cellcolor{deltapos!21}+3.4 & \cellcolor{deltapos!20}+3.2 & \cellcolor{deltapos!19}+2.9 & \cellcolor{deltapos!20}+3.0 \\
VLMBias       & 2784 & \cellcolor{deltaneg!12}-0.9 & \cellcolor{deltapos!11}+1.1 & \cellcolor{deltapos!5}+0.2 & \cellcolor{deltapos!18}+2.7 \\
MMStar        & 1500 & \cellcolor{deltapos!14}+1.6 & \cellcolor{deltapos!21}+3.4 & \cellcolor{deltapos!6}+0.4 & \cellcolor{deltapos!17}+2.3 \\
MMBench       & 1292 & \cellcolor{deltapos!30}+3.6 & \cellcolor{deltapos!22}+3.3 & \cellcolor{deltapos!10}+2.6 & \cellcolor{white}+2.2 \\
Perception    & 3000 & \cellcolor{deltapos!16}+2.0 & \cellcolor{deltapos!11}+1.2 & \cellcolor{deltapos!14}+1.6 & \cellcolor{deltapos!13}+1.5 \\
BLINK         & 1901 & \cellcolor{deltaneg!9}-0.6 & \cellcolor{deltaneg!5}-0.1 & \cellcolor{deltapos!13}+1.4 & \cellcolor{deltapos!12}+1.3 \\
Q-Bench       & 1495 & \cellcolor{deltapos!6}+0.4 & \cellcolor{deltaneg!14}-1.1 & \cellcolor{deltapos!5}+0.3 & \cellcolor{deltapos!9}+0.8 \\
TallyQA\subsetmark & 3000 & \cellcolor{deltapos!16}+2.0 & \cellcolor{deltapos!9}+0.8 & \cellcolor{deltapos!5}+0.3 & \cellcolor{deltapos!8}+0.6 \\
VStar         &  191 & \cellcolor{deltapos!26}+5.2 & \cellcolor{deltapos!18}+2.6 & \cellcolor{deltapos!18}+2.6 & \cellcolor{deltapos!7}+0.5 \\
Hallusion     & 1129 & \cellcolor{deltapos!25}+4.9 & \cellcolor{deltaneg!10}-0.8 & \cellcolor{deltaneg!15}-1.5 & \cellcolor{deltaneg!17}-2.1 \\
\midrule
Example       & 22902 & \cellcolor{deltapos!30}+3.14 & \cellcolor{deltapos!22}+2.87 & \cellcolor{white}+2.18 & \cellcolor{deltapos!24}+2.94 \\
Benchmark     & 13 & \cellcolor{deltapos!30}+4.07 & \cellcolor{deltapos!21}+3.90 & \cellcolor{white}+3.48 & \cellcolor{deltapos!11}+3.70 \\
\bottomrule
\end{tabular*}
\end{minipage}\hfill
\begin{minipage}[t]{0.323\linewidth}
\vspace{0pt}
\centering
\begin{tabular*}{\linewidth}{@{\extracolsep{\fill}}lccccc@{}}
\toprule
\textbf{8B} & $N$ & $\Delta_{16}$ & $\Delta_{32}$ & $\Delta_{64}$ & $\Delta_{\mathrm{Mix}}$ \\
\midrule
BlindTest     & 3434 & \cellcolor{deltapos!24}+4.6 & \cellcolor{deltapos!24}+4.3 & \cellcolor{deltapos!26}+5.2 & \cellcolor{deltapos!29}+7.8 \\
MMVP          &  150 & \cellcolor{white}+0.7 & \cellcolor{white}+0.7 & \cellcolor{deltapos!30}+8.0 & \cellcolor{deltapos!14}+4.0 \\
Q-Bench       & 1495 & \cellcolor{deltapos!20}+3.0 & \cellcolor{deltapos!24}+4.3 & \cellcolor{deltapos!18}+2.5 & \cellcolor{deltapos!18}+2.6 \\
TallyQA\subsetmark & 3000 & \cellcolor{deltapos!19}+2.8 & \cellcolor{deltapos!20}+3.3 & \cellcolor{deltapos!18}+2.5 & \cellcolor{deltapos!18}+2.6 \\
MMStar        & 1500 & \cellcolor{deltapos!24}+4.5 & \cellcolor{deltapos!17}+2.3 & \cellcolor{deltapos!23}+4.0 & \cellcolor{deltapos!17}+2.3 \\
BabyVision    &  388 & \cellcolor{deltaneg!16}-1.8 & \cellcolor{deltapos!13}+1.5 & \cellcolor{white}0.0 & \cellcolor{deltapos!17}+2.1 \\
BLINK         & 1901 & \cellcolor{deltapos!19}+2.9 & \cellcolor{deltapos!9}+0.8 & \cellcolor{deltapos!17}+2.4 & \cellcolor{deltapos!14}+1.7 \\
MMBench       & 1292 & \cellcolor{deltapos!30}+3.4 & \cellcolor{deltapos!2}+1.6 & \cellcolor{white}+1.5 & \cellcolor{deltapos!2}+1.6 \\
Perception    & 3000 & \cellcolor{deltapos!13}+1.4 & \cellcolor{deltapos!16}+2.0 & \cellcolor{deltapos!16}+1.9 & \cellcolor{deltapos!13}+1.3 \\
CV-Bench      & 2638 & \cellcolor{deltaneg!6}-0.3 & \cellcolor{deltapos!9}+0.8 & \cellcolor{deltaneg!5}-0.2 & \cellcolor{deltapos!10}+1.0 \\
VStar         &  191 & \cellcolor{deltapos!14}+1.6 & \cellcolor{deltaneg!7}-0.5 & \cellcolor{white}0.0 & \cellcolor{deltaneg!12}-1.0 \\
VLMBias       & 2784 & \cellcolor{deltapos!5}+0.1 & \cellcolor{deltapos!15}+1.8 & \cellcolor{deltapos!9}+0.9 & \cellcolor{deltaneg!14}-1.1 \\
Hallusion     & 1129 & \cellcolor{deltaneg!25}-4.9 & \cellcolor{deltaneg!29}-7.0 & \cellcolor{deltaneg!27}-6.1 & \cellcolor{deltaneg!24}-4.1 \\
\midrule
Example       & 22902 & \cellcolor{white}+1.88 & \cellcolor{deltapos!6}+1.92 & \cellcolor{deltapos!3}+1.90 & \cellcolor{deltapos!30}+2.07 \\
Benchmark     & 13 & \cellcolor{deltapos!9}+1.38 & \cellcolor{white}+1.23 & \cellcolor{deltapos!30}+1.73 & \cellcolor{deltapos!22}+1.60 \\
\bottomrule
\end{tabular*}
\end{minipage}
\placepositionedtablecaptionbottom
\end{table}

\begin{table}[H]
\centering
\setpositionedtablecaption{\tablecaptionfont\textbf{Qwen3-VL Grid-quantity
sweep.} Values are accuracy deltas from the corresponding FV-only baseline.
Example pools official evaluation units and Benchmark weights benchmarks
equally. Bold and underline mark the best and second-best distinct values
within each row.}
{tab:qwen-quantity-summary}
\scriptsize
\setlength{\tabcolsep}{5pt}
\begin{tabular}{clccccc}
\toprule
Model & Avg. & $\Delta_{\mathrm{1k}}$ & $\Delta_{\mathrm{2k}}$ &
$\Delta_{\mathrm{3k}}$ & $\Delta_{\mathrm{4k}}$ & $\Delta_{\mathrm{5k}}$ \\
\midrule
\multirow{2}{*}{2B} & Example
& \underline{+2.43} & \textbf{+2.68} & +2.01 & +2.31 & +2.18 \\
& Benchmark
& +3.30 & +3.20 & +3.02 & \underline{+3.31} & \textbf{+3.91} \\
\arrayrulecolor{black!50}\midrule\arrayrulecolor{black}
\multirow{2}{*}{4B} & Example
& \underline{+2.73} & +2.08 & \textbf{+2.94} & +1.92 & +2.66 \\
& Benchmark
& +3.33 & +3.01 & \textbf{+3.70} & +2.77 & \underline{+3.61} \\
\arrayrulecolor{black!50}\midrule\arrayrulecolor{black}
\multirow{2}{*}{8B} & Example
& \underline{+1.59} & +1.21 & \textbf{+2.07} & +0.59 & +0.99 \\
& Benchmark
& \textbf{+1.67} & +1.04 & \underline{+1.60} & +0.90 & +0.55 \\
\bottomrule
\end{tabular}
\placepositionedtablecaptionbottom
\end{table}

\subsection{Fixed Data Amount and Training Steps}
\label{app:fixed-data}

To compare reconstruction tuning under matched training conditions, we fix
both the amount of training data and the number of updates. FV-only contains
64,000 FineVision occurrences. Each Pixel configuration
replaces 3,000 of those occurrences with pixel-reconstruction examples, so
every fine-tuning run uses exactly 64,000 occurrences and stops at 1,000
optimizer steps. The Pixel configurations therefore contain less FineVision
supervision.

The single-resolution variants use eight-color PIA-reduced targets at
$16^2$, $32^2$, or $64^2$ in separate runs, reusing the same 3,000 unique
source-image IDs across runs. The mixed variant assigns each image from that
batch to exactly one resolution, approximately evenly, using sampling seed 0.
Source-image sampling is without replacement, so no image is repeated within
a run. All fine-tuning runs use training seed 42,
four GPUs with DeepSpeed ZeRO-2, per-GPU batch size 4, and four gradient
accumulation steps; the 4B runs use NVIDIA A100 80GB GPUs. These settings
match the number of training occurrences and optimizer steps; target sequence
length still grows with grid size.

\textbf{Training-seed sensitivity.}
For the 4B setting with 4,000 Pixel occurrences, all three seeds yield positive,
tightly clustered deltas (Table~\ref{tab:prt-seed-sensitivity}; SD 0.150 Example
and 0.197 Benchmark). This auxiliary study uses the original MMVP scoring protocol,
so the values differ from the main PRT tables; all runs and baselines here use
the same evaluation protocol.

\begin{table}[H]
\begin{minipage}[t]{0.49\linewidth}
\vspace{0pt}
\centering
\setpositionedtablecaption{\tablecaptionfont\textbf{Fixed-data-amount,
fixed-step comparison.} Every run uses 64,000 training occurrences and 1,000
optimizer steps. Values are Pixel-minus-FV-only accuracy deltas; Example pools
22,554 official evaluation units after MMVP pair-level and MMBench CircularEval
aggregation, while Benchmark weights 13 benchmarks equally.}
{tab:prt-fixed-data-summary}
\scriptsize
\setlength{\tabcolsep}{3pt}
\begin{tabular*}{\linewidth}{@{\extracolsep{\fill}}llrr@{}}
\toprule
Model & Setting & $\Delta$ Example & $\Delta$ Benchmark \\
\midrule
\multirow{4}{*}{2B}
 & Pixel-16  & \textbf{+3.152} & \textbf{+4.278} \\
 & Pixel-32  & +1.534 & +2.481 \\
 & Pixel-64  & +2.833 & \underline{+3.744} \\
 & Pixel-Mix & \underline{+2.966} & +3.577 \\
\arrayrulecolor{black!50}\midrule\arrayrulecolor{black}
\multirow{4}{*}{4B}
 & Pixel-16  & \textbf{+2.270} & \textbf{+3.129} \\
 & Pixel-32  & \underline{+1.335} & +1.828 \\
 & Pixel-64  & +1.171 & \underline{+2.306} \\
 & Pixel-Mix & +1.134 & +1.993 \\
\bottomrule
\end{tabular*}
\placepositionedtablecaptionbottom
\end{minipage}
\hfill
\begin{minipage}[t]{0.49\linewidth}
\vspace{0pt}
\centering
\setpositionedtablecaption{\tablecaptionfont\textbf{Seed sensitivity (4B,
4,000 Pixel occurrences).} Values are Pixel-minus-FV-only accuracy deltas;
Example pools examples and Benchmark weights benchmarks equally. Summary statistics
are computed across three seeds.}
{tab:prt-seed-sensitivity}
\scriptsize
\setlength{\tabcolsep}{3pt}
\begin{tabular*}{\linewidth}{@{\extracolsep{\fill}}lcc@{}}
\toprule
Run & $\Delta$ Example & $\Delta$ Benchmark \\
\midrule
Seed 42 & +1.455 & +1.789 \\
Seed 43 & +1.746 & +2.053 \\
Seed 44 & +1.534 & +1.668 \\
\midrule
Mean $\pm$ SD & +1.578 $\pm$ 0.150 & +1.837 $\pm$ 0.197 \\
SEM & 0.087 & 0.114 \\
95\% CI & [1.205, 1.952] & [1.348, 2.326] \\
\bottomrule
\end{tabular*}
\placepositionedtablecaptionbottom
\end{minipage}
\end{table}

Under the official aggregation, Pixel-16 yields the largest gains at both
model scales, and every setting---including 4B Pixel-Mix---improves both
averages.

\subsection{Limitations}
\label{app:limitations}

The main Molmo comparison adds Pixel trajectories to OV rather than replacing
OV rows, so the 2k condition receives eight additional updates. We also do not
compare against alternative auxiliary objectives on the same images, such as
caption-only trajectories. We therefore cannot fully separate the
palette-index grid from additional image exposure or its accompanying
language supervision. Pixel trajectories are longer than typical responses,
so their token share exceeds their 0.4\% share of examples.

The 2k Benchmark-average effect is small ($+0.5$ points), has a standardized
difference of $z=1.5$, and is positive in three of four matched seed pairs.
The 1k, 4k, and 8k settings have only two seeds each, and the observed gains do
not increase with quantity. These results support a small, consistent
aggregate benefit, not a data-scaling claim.

The secondary Qwen controls compare PRT against FineVision-only tuning and do
not isolate pixel targets from additional image exposure. Their additive runs
also contain 47 extra updates. The fixed-data, equal-step experiment above
addresses update count but replaces some FineVision supervision, leaving the
auxiliary-objective comparison unresolved.

\section{Model Inference Settings}
\label{app:inference-settings}

Table~\ref{tab:model-inference-settings} reports the task-specific inference
settings used for PixelBench generation. We retain one completed output per
item. A completed response that violates the output constraints is a model
error and receives zero, whereas provider-side failures and token-limit
truncations are retried; for the latter, we raise the output-token limit when
needed. Seed-2.1 Pro is the only model for which we impose an explicit
reasoning budget. We cap it at 2,500 tokens because pilot runs frequently
produced long, repetitive deliberation without improving the final pixel
output.

\begin{table}[H]
\centering
\setpositionedtablecaption{\tablecaptionfont\textbf{Model inference settings for
PixelBench generation.} Temperatures and reasoning controls are task-specific.
``Default'' denotes the provider default at the time of evaluation. The listed
output-token limit is the nominal setting and is raised only when retrying a
truncated response.}
{tab:model-inference-settings}
\scriptsize
\setlength{\tabcolsep}{4pt}
\renewcommand{\arraystretch}{1.08}
\begin{tabular*}{\linewidth}{@{\extracolsep{\fill}}lllrr@{}}
\toprule
Model & Temperature & Reasoning configuration & Max output tokens & Samples/item \\
\midrule
\multicolumn{5}{l}{\textit{\color{black!50}Text-to-Pixel (T2P)}} \\
GPT-6 Astra       & 1.0 & Default & 32,768 & 1 \\
Claude Opus 5     & 1.0 & Default (high effort) & 32,768 & 1 \\
GPT-5.6 Sol       & 1.0 & High effort & 32,768 & 1 \\
Kimi K3           & 1.0 & High effort & 32,768 & 1 \\
Gemini 3.6 Flash  & 0.7 & Default & 32,768 & 1 \\
Seed-2.1 Pro      & 0.7 & Reasoning budget 2,500 & 32,768 & 1 \\
DeepSeek V4 Pro   & 0.7 & Default & 32,768 & 1 \\
\arrayrulecolor{black!50}\midrule\arrayrulecolor{black}
\multicolumn{5}{l}{\textit{\color{black!50}Image-to-Pixel (I2P)}} \\
GPT-6 Astra       & 1.0 & Default & 32,768 & 1 \\
Claude Opus 5     & 1.0 & Default (high effort) & 32,768 & 1 \\
GPT-5.6 Sol       & 1.0 & High effort & 32,768 & 1 \\
Kimi K3           & 1.0 & High effort & 32,768 & 1 \\
Gemini 3.6 Flash  & 0.7 & Default & 32,768 & 1 \\
Seed-2.1 Pro      & 0.7 & Reasoning budget 2,500 & 32,768 & 1 \\
\bottomrule
\end{tabular*}
\placepositionedtablecaptionbottom
\end{table}

\section{SigLIP Retrieval Scores}
\label{app:siglip-retrieval}

Let \(\hat{x}\) denote a rendered pixel image and let \(\mathbf{q}\) be its
SigLIP embedding. Given a candidate set \(\mathcal{C}=\{\mathbf{c}_i\}_{i=1}^{K}\)
of the same embedding space, we score each candidate by cosine similarity,
\[
s(\mathbf{q},\mathbf{c}_i)
=
\frac{\mathbf{q}^{\top}\mathbf{c}_i}{\lVert\mathbf{q}\rVert\,\lVert\mathbf{c}_i\rVert}.
\]
The rank of the intended target \(\mathbf{c}^{\star}\) is
\[
r(\hat{x})
=
1+\sum_{\mathbf{c}\in\mathcal{C}\setminus\{\mathbf{c}^{\star}\}}
\mathbf{1}\!\left[s(\mathbf{q},\mathbf{c})>s(\mathbf{q},\mathbf{c}^{\star})\right],
\]
so \(r(\hat{x})=1\) if and only if the target is the nearest neighbor of the
query. Let \(\mathcal{V}\) index the valid outputs among \(N\) attempted
generations. The reported \emph{100-way Top-1} or \emph{500-way Top-1} score counts invalid outputs
as misses:
\[
\mathrm{Top\text{-}1}
=
\frac{1}{N}\sum_{n\in\mathcal{V}}
\mathbf{1}\!\left[r(\hat{x}_n)=1\right].
\]
The appendix MRR diagnostic is averaged over valid outputs:
\[
\mathrm{MRR}
=
\frac{1}{|\mathcal{V}|}\sum_{n\in\mathcal{V}}\frac{1}{r(\hat{x}_n)}.
\]
For T2P, \(\mathbf{q}\) comes from the SigLIP-SO400M image encoder and each
\(\mathbf{c}_i\) is the corresponding text embedding of ``a pixel art icon
depicting \{label\}''; we report \(K=100\) and \(K=500\). For I2P, both the
query and candidates are image embeddings from the same SigLIP-SO400M model,
and \(K=100\) is the set of benchmark source images.

\section{Full Evaluation Prompts}
\label{app:full-prompts}

The fields \texttt{label} (T2P) and \texttt{sample\_id} (I2P) are included only
to simplify automated parsing and pairing during evaluation; they are not part
of the visual generation objective. For pixel understanding,
\texttt{\{compact\_pixel\_json\}} is the serialized palette-and-grid input, and
\texttt{\{choice\_1\}}-\texttt{\{choice\_10\}} are the ten candidate labels.
For VLM judging, \texttt{\{label\}} and \texttt{\{size\}} are replaced with
the target concept and native grid size, and the rendered image is supplied.

\subsection{Text-to-Pixel Prompt}
\label{app:t2p-prompt}

\begin{promptbox}
\begin{Verbatim}[fontsize=\scriptsize,breaklines=true,breakanywhere=true,
breaksymbolleft={},breaksymbolright={},
listparameters={\setlength{\topsep}{0pt}\setlength{\partopsep}{0pt}}]
Create exactly N recognizable SxS pixel-art icons, one for each of these English labels in this exact order:
{labels}

Return JSON only, with no Markdown fence or commentary, using:
{"images": [{"label": "{label}",
 "palette": ["#FFFFFF", "#D62828", "#2A9D3F"],
 "pixels": ["000...", "... exactly S strings"]}]}

Requirements:
- Every icon must have exactly S rows and S columns.
- Use 2 to 8 opaque RGB colors, including a background color.
- Encode each pixel as one digit from 0 to 7.
- Favor simple silhouettes, high contrast, and centered subjects.
- Make each object recognizable at SxS resolution; use no text.
\end{Verbatim}
\end{promptbox}

\subsection{Text-to-Pixel VLM-as-Judge Prompt}
\label{app:t2p-judge-prompt}

\textbf{Judge inference and stability.}
We use Gemini~3.5 Flash with temperature 0, thinking budget 0, a maximum of
1,536 output tokens, and one sample per item. As a stability check, we re-score
a 10\% subset after the initial evaluation; the mean absolute change in total
score is below 2 points on the 100-point scale, indicating that the reported
VLM Judge scores are robust to repeated evaluation.

\begin{promptbox}
\begin{Verbatim}[fontsize=\scriptsize,breaklines=true,breakanywhere=true,
breaksymbolleft={},breaksymbolright={},
listparameters={\setlength{\topsep}{0pt}\setlength{\partopsep}{0pt}}]
Act as a strict evaluator of a text-to-pixel generation. The target concept is
"{label}". The supplied image is a nearest-neighbor enlargement of a
native {size}x{size} pixel-art image. Large square pixels are intentional.

Score the image using all six independent dimensions below. Do not reward
complexity by itself, do not infer the generator identity, and do not excuse a
wrong or unrecognizable subject merely because the image is attractive.

1. subject_identity (0-20): Does the visible subject clearly depict the target
   concept rather than a related, ambiguous, or incorrect object?
2. defining_attributes (0-25): Are the target's identity-bearing parts, shapes,
   and relations present and correct?
3. structure_proportion (0-20): Is the silhouette coherent, proportionate,
   well framed, and structurally correct rather than cropped or merged?
4. detail_quality (0-20): Are internal details coherent, informative, and
   consistent with the target at the native resolution?
5. pixel_art_execution (0-10): Are pixel clusters and edges deliberate
   and appropriate for the native resolution, without random noise or
   fragmented pseudo-texture?
6. palette_contrast (0-5): Is the limited palette coherent and does contrast
   preserve the subject and its important parts?

Return JSON only, without Markdown:
{
  "observed_subject": "short literal description of what is visibly depicted",
  "subject_identity": {"score": 0, "reason": "brief evidence"},
  "defining_attributes": {"score": 0, "reason": "brief evidence"},
  "structure_proportion": {"score": 0, "reason": "brief evidence"},
  "detail_quality": {"score": 0, "reason": "brief evidence"},
  "pixel_art_execution": {"score": 0, "reason": "brief evidence"},
  "palette_contrast": {"score": 0, "reason": "brief evidence"},
  "confidence": "low|medium|high",
  "failure_flags": []
}
\end{Verbatim}
\end{promptbox}

\subsection{Image-to-Pixel Prompt}
\label{app:i2p-prompt}

\begin{promptbox}
\begin{Verbatim}[fontsize=\scriptsize,breaklines=true,breakanywhere=true,
breaksymbolleft={},breaksymbolright={},
listparameters={\setlength{\topsep}{0pt}\setlength{\partopsep}{0pt}}]
Convert the supplied reference image into a faithful SxS pixel-art image. Preserve the main subject's identity, silhouette, pose, composition, and dominant colors. Simplify fine detail only as required by the pixel resolution.
Do not use any object name or text.

Return JSON only:
{"sample_id": "{sample_id}",
 "palette": ["#FFFFFF", "#111111"],
 "pixels": ["000...", "... exactly S strings"]}

Use 2 to 8 opaque #RRGGBB colors.
\end{Verbatim}
\end{promptbox}

\subsection{Pixel Understanding Prompt}
\label{app:understanding-prompt}

\textbf{Candidate construction.}
For each question, we exclude the correct concept from the benchmark's 100
object concepts, deterministically shuffle the remaining 99, and select the
first nine as distractors. We then deterministically reshuffle the correct
answer together with these nine distractors to assign the ten choices to
A-J. Each item therefore
contains ten distinct candidates with exactly one correct answer, and the
construction is reproducible.

\begin{promptbox}
\begin{Verbatim}[fontsize=\scriptsize,breaklines=true,breakanywhere=true,
breaksymbolleft={},breaksymbolright={},
listparameters={\setlength{\topsep}{0pt}\setlength{\partopsep}{0pt}}]
This is a {size}x{size} pixel-art icon encoded as JSON.
"palette" is 2-8 hex colors. "pixels" are {size} row strings; each character is a palette index 0-7.
Identify the depicted object. Reply with ONE letter only (A, B, C, D, E, F, G, H, I, J).

{compact_pixel_json}

A. {choice_1}
B. {choice_2}
C. {choice_3}
D. {choice_4}
E. {choice_5}
F. {choice_6}
G. {choice_7}
H. {choice_8}
I. {choice_9}
J. {choice_10}
\end{Verbatim}
\end{promptbox}

\section{Text-to-Pixel Concept List}
\label{app:t2p-concepts}

Table~\ref{tab:t2p-concepts} lists the 100 T2P concepts in the order used by
the benchmark. The set contains 18 animals, 20 food items, 30 household
objects, 15 means of transport, and 17 nature-related objects or symbols.

\begin{table}[H]
\centering
\setpositionedtablecaption{\tablecaptionfont\textbf{Complete Text-to-Pixel concept list.} Numeric identifiers
preserve the original benchmark order. Background colors denote
\colorbox{conceptanimal}{animal}, \colorbox{conceptfood}{food},
\colorbox{concepthousehold}{household},
\colorbox{concepttransport}{transport}, and
\colorbox{conceptnature}{nature/symbol}.}
{tab:t2p-concepts}
\footnotesize
\setlength{\tabcolsep}{5pt}
\begin{tabular*}{\linewidth}{l@{\extracolsep{\fill}}llll}
\toprule
\animalconcept{4.\ cat} & \animalconcept{8.\ fish} & \animalconcept{17.\ dog} & \animalconcept{18.\ bird} & \animalconcept{19.\ rabbit} \\
\animalconcept{20.\ elephant} & \animalconcept{21.\ turtle} & \animalconcept{22.\ butterfly} & \animalconcept{51.\ cow} & \animalconcept{52.\ sheep} \\
\animalconcept{53.\ pig} & \animalconcept{54.\ chicken} & \animalconcept{55.\ owl} & \animalconcept{56.\ penguin} & \animalconcept{57.\ frog} \\
\animalconcept{58.\ snake} & \animalconcept{59.\ bee} & \animalconcept{60.\ snail} & \foodconcept{1.\ apple} & \foodconcept{3.\ banana} \\
\foodconcept{23.\ orange} & \foodconcept{24.\ strawberry} & \foodconcept{25.\ watermelon} & \foodconcept{26.\ carrot} & \foodconcept{27.\ pizza} \\
\foodconcept{28.\ cake} & \foodconcept{29.\ bread} & \foodconcept{30.\ ice cream} & \foodconcept{61.\ pear} & \foodconcept{62.\ pineapple} \\
\foodconcept{63.\ grapes} & \foodconcept{64.\ cherry} & \foodconcept{65.\ lemon} & \foodconcept{66.\ corn} & \foodconcept{67.\ donut} \\
\foodconcept{68.\ hamburger} & \foodconcept{69.\ sandwich} & \foodconcept{70.\ cookie} & \householdconcept{2.\ stool} & \householdconcept{5.\ house} \\
\householdconcept{10.\ cup} & \householdconcept{14.\ key} & \householdconcept{15.\ umbrella} & \householdconcept{16.\ clock} & \householdconcept{31.\ lamp} \\
\householdconcept{32.\ spoon} & \householdconcept{33.\ scissors} & \householdconcept{34.\ toothbrush} & \householdconcept{35.\ bottle} & \householdconcept{36.\ camera} \\
\householdconcept{37.\ phone} & \householdconcept{38.\ book} & \householdconcept{39.\ hammer} & \householdconcept{71.\ fork} & \householdconcept{72.\ knife} \\
\householdconcept{73.\ plate} & \householdconcept{74.\ bowl} & \householdconcept{75.\ bed} & \householdconcept{76.\ table} & \householdconcept{77.\ sofa} \\
\householdconcept{78.\ television} & \householdconcept{79.\ computer} & \householdconcept{80.\ backpack} & \householdconcept{82.\ hat} & \householdconcept{89.\ wrench} \\
\householdconcept{90.\ screwdriver} & \householdconcept{91.\ axe} & \householdconcept{92.\ shovel} & \transportconcept{7.\ car} & \transportconcept{40.\ bus} \\
\transportconcept{41.\ bicycle} & \transportconcept{42.\ airplane} & \transportconcept{43.\ boat} & \transportconcept{44.\ train} & \transportconcept{45.\ rocket} \\
\transportconcept{46.\ truck} & \transportconcept{81.\ shoe} & \transportconcept{83.\ motorcycle} & \transportconcept{84.\ helicopter} & \transportconcept{85.\ submarine} \\
\transportconcept{86.\ tractor} & \transportconcept{87.\ skateboard} & \transportconcept{88.\ sailboat} & \natureconcept{6.\ tree} & \natureconcept{9.\ flower} \\
\natureconcept{11.\ star} & \natureconcept{12.\ heart} & \natureconcept{13.\ mushroom} & \natureconcept{47.\ sun} & \natureconcept{48.\ moon} \\
\natureconcept{49.\ cloud} & \natureconcept{50.\ mountain} & \natureconcept{93.\ rainbow} & \natureconcept{94.\ snowflake} & \natureconcept{95.\ lightning} \\
\natureconcept{96.\ fire} & \natureconcept{97.\ leaf} & \natureconcept{98.\ cactus} & \natureconcept{99.\ palm tree} & \natureconcept{100.\ castle} \\
\bottomrule
\end{tabular*}
\placepositionedtablecaptionbottom
\end{table}

\section{Image-to-Pixel Source Images}
\label{app:i2p-sources}

\begin{figure}[H]
\centering
\includegraphics[width=\linewidth]{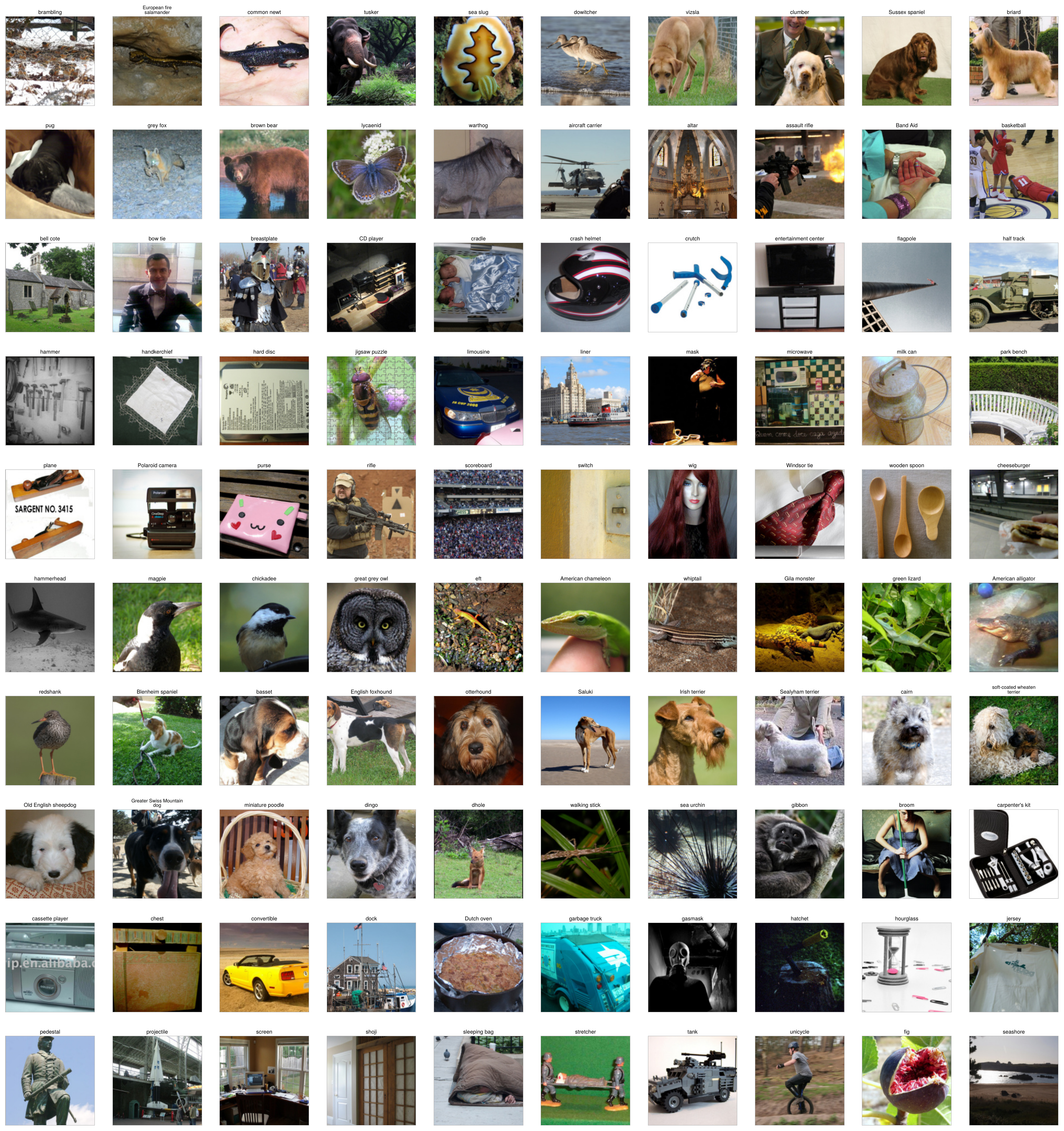}
\caption{\footnotesize\textbf{I2P source images.} The 100 ImageNet-1K
validation crops used for image-to-pixel generation, arranged in evaluation
order. Class names are shown for reference only.}
\label{fig:i2p-sources}
\end{figure}

Figure~\ref{fig:i2p-sources} shows the 100 ImageNet validation crops used as
I2P inputs. Each class contributes one center-cropped $256\times256$ image;
labels are shown for reference and are not provided to the model during
generation.

\section{500-Way Retrieval Vocabulary}
\label{app:t2p-500way}

Tables~\ref{tab:t2p-500way-short} and~\ref{tab:t2p-500way-long} list the 500
text candidates used for T2P 500-way Top-1 retrieval. Numbers are evaluation
indices: the first 100 indices are the benchmark concepts in
Table~\ref{tab:t2p-concepts}, and indices 101-500 are additional distractors.
For layout, we sort entries by label length and split at 9 characters
(271 shorter labels; 229 longer ones). Within each table, entries fill
columns from top to bottom so shorter labels occupy narrower columns.

\begin{table}[H]
\centering
\setpositionedtablecaption{\tablecaptionfont\textbf{Shorter 500-way labels ($\leq$9 characters).}
Sorted by increasing length and filled top-to-bottom across eight columns;
271 entries. Numbers are evaluation indices.}
{tab:t2p-500way-short}
\tiny
\setlength{\tabcolsep}{3pt}
\renewcommand{\arraystretch}{0.78}
\begin{tabular}{llllllll}
\toprule
500.~ox & 269.~harp & 281.~flute & 162.~buckle & 286.~teapot & 93.~rainbow & 168.~lorikeet & 421.~drumstick \\
91.~axe & 31.~lamp & 327.~goose & 98.~cactus & 375.~throne & 448.~Samoyed & 163.~Malinois & 165.~golf cart \\
75.~bed & 97.~leaf & 12.~heart & 36.~camera & 344.~tights & 154.~sawmill & 446.~marmoset & 362.~grey wolf \\
59.~bee & 333.~mask & 5.~house & 26.~carrot & 386.~tripod & 257.~snorkel & 50.~mountain & 189.~guacamole \\
204.~bra & 275.~maze & 172.~junco & 100.~castle & 21.~turtle & 424.~spatula & 13.~mushroom & 334.~gyromitra \\
40.~bus & 161.~mink & 72.~knife & 64.~cherry & 216.~tusker & 140.~sundial & 221.~odometer & 355.~hair clip \\
7.~car & 48.~moon & 65.~lemon & 498.~collie & 191.~valley & 128.~T-shirt & 229.~payphone & 68.~hamburger \\
4.~cat & 61.~pear & 252.~llama & 70.~cookie & 434.~Vizsla & 483.~thimble & 235.~platypus & 237.~harmonica \\
51.~cow & 337.~safe & 202.~macaw & 155.~coucal & 124.~wallet & 86.~tractor & 402.~plectrum & 195.~hourglass \\
10.~cup & 81.~shoe & 147.~moped & 132.~cougar & 180.~weasel & 268.~viaduct & 433.~popsicle & 30.~ice cream \\
17.~dog & 473.~sock & 37.~phone & 497.~coyote & 89.~wrench & 330.~volcano & 293.~rapeseed & 309.~lakeshore \\
484.~fig & 77.~sofa & 27.~pizza & 425.~dugong & 371.~bassoon & 137.~warthog & 88.~sailboat & 95.~lightning \\
122.~fly & 11.~star & 73.~plate & 491.~eggnog & 273.~bathtub & 192.~website & 69.~sandwich & 233.~monastery \\
82.~hat & 489.~suit & 118.~purse & 450.~eraser & 41.~bicycle & 42.~airplane & 33.~scissors & 416.~motorboat \\
116.~hay & 414.~tank & 432.~quail & 9.~flower & 54.~chicken & 80.~backpack & 389.~scorpion & 99.~palm tree \\
328.~jay & 396.~tent & 266.~quill & 179.~geyser & 135.~doormat & 350.~baseball & 205.~sea lion & 435.~partridge \\
14.~key & 404.~tram & 488.~quilt & 63.~grapes & 243.~echidna & 340.~chainsaw & 138.~stingray & 62.~pineapple \\
361.~mop & 6.~tree & 392.~rifle & 39.~hammer & 254.~go-kart & 79.~computer & 479.~sturgeon & 181.~plant pot \\
55.~owl & 303.~vase & 208.~scarf & 377.~isopod & 478.~gondola & 284.~consomme & 236.~terrapin & 321.~porcupine \\
53.~pig & 318.~acorn & 52.~sheep & 324.~jaguar & 199.~gorilla & 198.~crayfish & 15.~umbrella & 265.~projector \\
47.~sun & 381.~agama & 60.~snail & 245.~Kuvasz & 347.~hamster & 206.~dumbbell & 459.~vestment & 113.~red panda \\
212.~wok & 1.~apple & 58.~snake & 169.~langur & 406.~hatchet & 20.~elephant & 280.~armadillo & 94.~snowflake \\
18.~bird & 130.~bagel & 32.~spoon & 315.~maraca & 464.~hot dog & 436.~envelope & 193.~barometer & 150.~spoonbill \\
43.~boat & 360.~beach & 2.~stool & 261.~muzzle & 395.~hot pot & 171.~fireboat & 102.~bobsleigh & 480.~stopwatch \\
38.~book & 262.~bison & 76.~table & 23.~orange & 209.~ladybug & 143.~flagpole & 220.~brambling & 85.~submarine \\
74.~bowl & 29.~bread & 44.~train & 111.~pillow & 338.~limpkin & 294.~flamingo & 22.~butterfly & 398.~sunscreen \\
28.~cake & 339.~canoe & 46.~truck & 240.~poncho & 394.~macaque & 218.~flatworm & 276.~car wheel & 244.~tarantula \\
66.~corn & 219.~chain & 454.~vespa & 194.~prison & 227.~Maltese & 313.~fountain & 114.~centipede & 123.~tiger cat \\
443.~dome & 447.~cloak & 354.~zebra & 19.~rabbit & 364.~marimba & 177.~fur coat & 401.~chickadee & 152.~trash can \\
96.~fire & 16.~clock & 481.~baboon & 410.~racket & 305.~minibus & 167.~goldfish & 141.~cockroach & 115.~tree frog \\
8.~fish & 456.~clogs & 3.~banana & 45.~rocket & 210.~monitor & 299.~hair wig & 282.~corkscrew & 190.~turnstile \\
71.~fork & 49.~cloud & 387.~beaker & 431.~sandal & 108.~padlock & 238.~hornbill & 475.~crossword & 153.~water jug \\
57.~frog & 311.~dingo & 290.~beaver & 160.~shield & 384.~pelican & 278.~knee pad & 157.~Dalmatian & 366.~wild boar \\
419.~gown & 67.~donut & 35.~bottle & 92.~shovel & 56.~penguin & 246.~lacewing & 291.~dishcloth &  \\
\bottomrule
\end{tabular}

\placepositionedtablecaptionbottom
\end{table}

\begin{table}[H]
\centering
\setpositionedtablecaption{\tablecaptionfont\textbf{Longer 500-way labels ($\geq$10 characters).}
Sorted by increasing length and filled top-to-bottom across four columns;
229 entries. Numbers are evaluation indices.}
{tab:t2p-500way-long}
\tiny
\setlength{\tabcolsep}{4pt}
\renewcommand{\arraystretch}{0.80}
\begin{tabular}{llll}
\toprule
260.~Arctic fox & 499.~coffeemaker & 249.~Cairn Terrier & 119.~steam locomotive \\
263.~bald eagle & 442.~dung beetle & 365.~china cabinet & 196.~computer keyboard \\
397.~black swan & 382.~fire screen & 490.~digital clock & 186.~impala (antelope) \\
477.~Bloodhound & 264.~freight car & 336.~garbage truck & 482.~little blue heron \\
383.~box turtle & 207.~grand piano & 166.~grocery store & 295.~monarch butterfly \\
461.~bridegroom & 452.~great egret & 158.~howler monkey & 373.~Redbone Coonhound \\
213.~can opener & 234.~green mamba & 427.~Irish Terrier & 285.~rhinoceros beetle \\
156.~car mirror & 139.~hermit crab & 106.~Japanese Chin & 356.~Shetland Sheepdog \\
439.~chiffonier & 495.~hummingbird & 105.~manhole cover & 107.~sulphur butterfly \\
277.~chimpanzee & 259.~hunting bow & 201.~measuring cup & 405.~threshing machine \\
296.~coral reef & 385.~leaf beetle & 148.~messenger bag & 426.~Bluetick Coonhound \\
463.~crane bird & 272.~mixing bowl & 465.~movie theater & 469.~Christmas stocking \\
451.~dishwasher & 185.~pencil case & 487.~red king crab & 485.~hand-held computer \\
370.~fire truck & 357.~Persian cat & 331.~reflex camera & 493.~Labrador Retriever \\
121.~frying pan & 230.~pirate ship & 378.~rocking chair & 453.~Scottish Deerhound \\
242.~greenhouse & 223.~planetarium & 467.~silver salmon & 367.~semi-trailer truck \\
255.~guinea pig & 173.~rain barrel & 174.~upright piano & 444.~smooth green snake \\
462.~hair dryer & 250.~salt shaker & 215.~velvet fabric & 149.~Alaskan tundra wolf \\
144.~hair spray & 90.~screwdriver & 412.~American robin & 248.~black-footed ferret \\
274.~hammerhead & 496.~sea anemone & 126.~Carolina anole & 267.~electric locomotive \\
84.~helicopter & 241.~Siamese cat & 353.~cliff dwelling & 297.~Irish Water Spaniel \\
317.~hoop skirt & 300.~tape player & 253.~Cocker Spaniel & 486.~kite (bird of prey) \\
428.~lawn mower & 326.~tennis ball & 117.~Dungeness crab & 307.~through arch bridge \\
142.~leafhopper & 363.~tiger shark & 103.~filing cabinet & 476.~Cardigan Welsh Corgi \\
146.~lighthouse & 214.~titi monkey & 379.~French Bulldog & 359.~entertainment center \\
83.~motorcycle & 332.~toilet seat & 449.~graduation cap & 306.~Old English Sheepdog \\
415.~mud turtle & 164.~trench coat & 420.~great grey owl & 289.~southern black widow \\
349.~neck brace & 322.~triceratops & 372.~ping-pong ball & 125.~swim trunks / shorts \\
423.~Otterhound & 298.~wheelbarrow & 182.~shower curtain & 400.~European green lizard \\
352.~paintbrush & 256.~Windsor tie & 380.~soap dispenser & 348.~red admiral butterfly \\
325.~park bench & 178.~wine bottle & 101.~spiral or coil & 492.~rotary dial telephone \\
251.~piggy bank & 224.~wolf spider & 310.~tandem bicycle & 323.~small white butterfly \\
329.~pipe organ & 351.~Afghan Hound & 127.~American dipper & 188.~yellow lady's slipper \\
239.~Pomeranian & 109.~barber chair & 104.~baseball player & 374.~Curly-coated Retriever \\
418.~pool table & 292.~black grouse & 225.~chocolate syrup & 222.~loupe magnifying glass \\
151.~promontory & 403.~Brittany dog & 358.~cocktail shaker & 258.~ram (adult male sheep) \\
455.~pufferfish & 466.~coral fungus & 393.~common redshank & 302.~red wolf or maned wolf \\
312.~rotisserie & 440.~Egyptian Mau & 314.~electric guitar & 228.~sidewinder rattlesnake \\
369.~safety pin & 399.~fox squirrel & 226.~fire salamander & 429.~Welsh Springer Spaniel \\
197.~sea urchin & 316.~green iguana & 422.~football helmet & 304.~black-and-white colobus \\
200.~shoe store & 131.~handkerchief & 458.~four-poster bed & 136.~Australian Silky Terrier \\
460.~shower cap & 368.~home theater & 211.~Groenendael dog & 417.~automated teller machine \\
87.~skateboard & 271.~killer whale & 471.~laptop computer & 170.~Bouvier des Flandres dog \\
391.~snoek fish & 335.~mosquito net & 145.~longhorn beetle & 301.~Chesapeake Bay Retriever \\
407.~snowmobile & 231.~patas monkey & 270.~mashed potatoes & 283.~English Springer Spaniel \\
110.~spider web & 438.~picket fence & 346.~Norfolk Terrier & 232.~sulphur-crested cockatoo \\
345.~stone wall & 287.~punching bag & 376.~Standard Poodle & 279.~German Shorthaired Pointer \\
24.~strawberry & 247.~railroad car & 472.~aircraft carrier & 288.~product packet / packaging \\
474.~table lamp & 494.~refrigerator & 184.~bulletproof vest & 112.~Staffordshire Bull Terrier \\
468.~teddy bear & 183.~sleeping bag & 430.~butternut squash & 470.~shoji screen / room divider \\
78.~television & 445.~sliding door & 441.~chain-link fence & 129.~Soft-coated Wheaten Terrier \\
34.~toothbrush & 342.~slot machine & 159.~desktop computer & 320.~American Staffordshire Terrier \\
25.~watermelon & 187.~snow leopard & 120.~English foxhound & 413.~desert grassland whiptail lizard \\
203.~worm snake & 134.~swimming cap & 457.~high-speed train & 437.~military hat (bearskin or shako) \\
133.~barn spider & 411.~toilet paper & 390.~magnetic compass & 343.~Mexican hairless dog (xoloitzcuintli) \\
409.~bell pepper & 341.~Affenpinscher & 217.~military uniform &  \\
175.~block plane & 308.~baby pacifier & 388.~Miniature Poodle &  \\
319.~brain coral & 408.~bighorn sheep & 176.~proboscis monkey &  \\
\bottomrule
\end{tabular}

\placepositionedtablecaptionbottom
\end{table}

\section{Full Text-to-Pixel Results}
\label{app:t2p-full-results}

Tables~\ref{tab:t2p-main} and~\ref{tab:t2p-mrr} report the per-resolution
Top-1 and MRR scores corresponding to Table~\ref{tab:t2p-main-results}.
Table~\ref{tab:t2p-judge-dimensions} decomposes the VLM Judge score into its
six rubric dimensions.

\begin{table}[H]
\centering
\setpositionedtablecaption{\tablecaptionfont\textbf{Full Text-to-Pixel semantic recognition results.}
SigLIP-SO400M 100-way and 500-way Top-1 under text-candidate pools.
``Gem'' and ``F'' denote Gemini and Flash. Invalid generations are counted as
retrieval misses. All values are proportions; higher is better.}
{tab:t2p-main}
\small
\setlength{\tabcolsep}{4pt}
\resizebox{\linewidth}{!}{%
\begin{tabular}{@{}llccccccc@{}}
\toprule
Resolution & Pool & Gem 3.6 F & GPT-5.6 Sol & GPT-6 Astra &
Opus 5 & Kimi K3 & DeepSeek V4 Pro & Seed-2.1 Pro \\
\midrule
$16^2$ & 100-way & .910 & .910 & \textbf{.950} & .910 & .900 & .580 & .810 \\
$16^2$ & 500-way & \textbf{.870} & .840 & \textbf{.870} & .810 & .810 & .490 & .700 \\
\addlinespace
$32^2$ & 100-way & .830 & .920 & \textbf{.980} & .950 & .930 & .680 & .820 \\
$32^2$ & 500-way & .780 & .890 & \textbf{.920} & .870 & .870 & .590 & .720 \\
\addlinespace
$64^2$ & 100-way & .810 & .890 & \textbf{.980} & .970 & .860 & .630 & .830 \\
$64^2$ & 500-way & .720 & .830 & \textbf{.940} & .890 & .770 & .510 & .740 \\
\bottomrule
\end{tabular}%
}
\placepositionedtablecaptionbottom
\end{table}

\begin{table}[H]
\centering
\setpositionedtablecaption{\tablecaptionfont\textbf{Full Text-to-Pixel mean reciprocal rank results.}
SigLIP-SO400M MRR under 100-way and 500-way candidate pools. ``Gem'' and ``F'' denote
Gemini and Flash. GPT-6 Astra and Claude Opus 5 have 100 valid generations at every resolution.
DeepSeek V4 Pro has 100, 99, and 97 valid generations at
$16^2$, $32^2$, and $64^2$, respectively. $^\dagger$Evaluated on 99 valid
generations; ``--'' denotes an unavailable result.}
{tab:t2p-mrr}
\small
\setlength{\tabcolsep}{4pt}
\resizebox{\linewidth}{!}{%
\begin{tabular}{@{}llccccccc@{}}
\toprule
Resolution & Pool & Gem3.6 F & GPT-5.6 Sol & GPT-6 Astra &
Opus 5 & Kimi K3 & DeepSeek V4 Pro & Seed-2.1 Pro \\
\midrule
$16^2$ & 100-way & .945 & .945 & \textbf{.972} & .949 & .941 & .697 & .874 \\
$16^2$ & 500-way & .913 & .899 & \textbf{.927} & .884 & .881 & .599 & .792 \\
\addlinespace
$32^2$ & 100-way & .887 & .956 & \textbf{.990} & .972 & .965 & .769 & .892 \\
$32^2$ & 500-way & .837 & .931 & \textbf{.960} & .925 & .932 & .681 & .815 \\
\addlinespace
$64^2$ & 100-way & .843 & .933 & \textbf{.990} & .985 & .896$^\dagger$ & .718 & .898 \\
$64^2$ & 500-way & .785 & .893 & \textbf{.970} & .945 & .826$^\dagger$ & .614 & .832 \\
\bottomrule
\end{tabular}%
}
\placepositionedtablecaptionbottom
\end{table}

\begin{table}[H]
\centering
\setpositionedtablecaption{\tablecaptionfont\textbf{Text-to-Pixel VLM Judge
dimension scores.} Gemini~3.5 Flash scores each rendered output on subject
identity (ID), defining attributes and key parts (Parts), structure and
proportion (Struct.), detail quality (Detail), pixel-art execution (Pixel), and
palette and contrast (Color). Entries are mean scores at each native
resolution; Total is out of 100. Bold and underline mark the best and
second-best values within each resolution.}
{tab:t2p-judge-dimensions}
\scriptsize
\setlength{\tabcolsep}{3.2pt}
\renewcommand{\arraystretch}{0.96}
\begin{tabular*}{\linewidth}{@{\extracolsep{\fill}}lrrrrrrr@{}}
\toprule
Model & ID/20 & Parts/25 & Struct./20 & Detail/20 & Pixel/10 & Color/5 & Total/100 \\
\midrule
\multicolumn{8}{l}{\textit{\color{black!50}$16^2$}} \\
GPT-6 Astra      & \textbf{18.83} & \textbf{22.89} & \textbf{18.16} & \textbf{16.95} & \textbf{8.94} & \textbf{4.98} & \textbf{90.75} \\
GPT-5.6 Sol      & 16.11 & \underline{19.31} & 15.77 & \underline{13.98} & \underline{8.21} & \underline{4.75} & \underline{78.13} \\
Claude Opus 5    & 15.77 & 18.52 & 15.61 & 12.60 & 7.88 & \underline{4.75} & 75.13 \\
Kimi K3          & 15.48 & 18.21 & 15.27 & 12.90 & 7.99 & 4.69 & 74.54 \\
Gemini 3.6 Flash & \underline{16.22} & 19.23 & \underline{15.79} & 13.66 & 8.17 & 4.72 & 77.79 \\
Seed-2.1 Pro     & 14.12 & 16.57 & 14.07 & 11.61 & 7.64 & 4.56 & 68.57 \\
DeepSeek V4 Pro  & 10.99 & 12.34 & 11.47 & 8.17 & 6.68 & 4.00 & 53.65 \\
\arrayrulecolor{black!50}\midrule\arrayrulecolor{black}
\multicolumn{8}{l}{\textit{\color{black!50}$32^2$}} \\
GPT-6 Astra      & \textbf{19.16} & \textbf{23.18} & \textbf{18.36} & \textbf{16.94} & \textbf{8.95} & \textbf{4.99} & \textbf{91.58} \\
GPT-5.6 Sol      & \underline{16.57} & \underline{19.52} & \underline{16.06} & \underline{13.80} & \underline{8.11} & \underline{4.80} & \underline{78.86} \\
Claude Opus 5    & 16.04 & 18.39 & 15.49 & 11.92 & 7.59 & 4.69 & 74.12 \\
Kimi K3          & 15.80 & 18.27 & 15.35 & 12.68 & 7.92 & 4.65 & 74.67 \\
Gemini 3.6 Flash & 14.77 & 17.05 & 14.47 & 12.33 & 7.99 & 4.60 & 71.21 \\
Seed-2.1 Pro     & 13.56 & 15.55 & 13.22 & 10.77 & 7.10 & 4.36 & 64.56 \\
DeepSeek V4 Pro  & 11.76 & 13.13 & 11.50 & 8.51 & 6.17 & 3.95 & 55.02 \\
\arrayrulecolor{black!50}\midrule\arrayrulecolor{black}
\multicolumn{8}{l}{\textit{\color{black!50}$64^2$}} \\
GPT-6 Astra      & \textbf{18.74} & \textbf{22.63} & \textbf{18.04} & \textbf{16.38} & \textbf{8.92} & \textbf{4.95} & \textbf{89.66} \\
GPT-5.6 Sol      & 15.14 & \underline{17.95} & \underline{15.05} & \underline{12.23} & \underline{7.87} & \underline{4.72} & \underline{72.96} \\
Claude Opus 5    & \underline{15.55} & 17.54 & 14.58 & 10.95 & 6.97 & 4.48 & 70.07 \\
Kimi K3          & 12.95 & 14.69 & 12.82 & 9.68 & 7.05 & 4.34 & 61.53 \\
Gemini 3.6 Flash & 12.67 & 14.30 & 12.30 & 9.87 & 7.05 & 4.16 & 60.35 \\
Seed-2.1 Pro     & 12.89 & 14.24 & 12.27 & 9.32 & 6.45 & 4.16 & 59.33 \\
DeepSeek V4 Pro  & 10.16 & 10.77 & 9.58 & 6.54 & 5.20 & 3.42 & 45.67 \\
\bottomrule
\end{tabular*}
\placepositionedtablecaptionbottom
\end{table}

\raggedbottom
\section{Algorithmic Image-to-Pixel Baselines in Full}
\label{app:baselines}

Each method either combines a spatial downscaling stage with a subsequent color
quantization stage or jointly determines the output regions and palette. We set
the palette size to eight for all constrained methods and evaluate them on the
same 100 ImageNet validation crops used for the VLMs.

\textbf{Tier 0: simple resampling.} Nearest-neighbor sampling selects the
source-cell center, while box filtering averages the corresponding source
patch. Both are followed by median-cut quantization
\citep{heckbert1982color}.

\textbf{Tier 1: linear filters.} Bicubic and Lanczos interpolation use
standard fixed resampling kernels before median-cut quantization. These methods
allocate the output resolution uniformly across the image.

\textbf{Tier 2: content-adaptive downscaling.} DPID
\citep{weber2016dpid} weights input pixels according to their deviation from a
local guidance color; we evaluate $\lambda\in\{0.5,1.0\}$.
SSIM-optimal downscaling \citep{oztireli2015perceptual} minimizes structural
dissimilarity using local image statistics. Both methods are followed by
median-cut quantization.

\textbf{Tier 3: pixel-art-specific methods.}
Pyxelate\footnote{\url{https://github.com/sedthh/pyxelate}} uses Sobel-gradient
information to merge local cells and estimates an eight-color palette with a
Bayesian Gaussian mixture. We use version 2.1 or later with
\texttt{sobel=3}, \texttt{depth=1}, and dithering disabled. Pixelated Image
Abstraction (PIA) \citep{gerstner2012pixelated} jointly constructs output
regions and a reduced palette. The reference implementation uses modified SLIC
mapping, mass-constrained deterministic annealing
\citep{rose1998deterministic}, bilateral superpixel smoothing, and a saturation
factor of $1.1$. PIA (reduced) retains SLIC with compactness $m=45$, replaces
annealing with weighted eight-means over superpixel colors, and omits bilateral
smoothing and saturation adjustment.

\subsection{Representative Implementations}
\label{app:baseline-implementations}

The following procedures specify the three representatives used in the main
comparison. All receive a $256\times256$ RGB crop $I$ and target side
length $N\in\{16,32,64\}$. Median cut uses $K=8$ without dithering. SLIC and
$k$-means in PIA (reduced) operate in CIELAB; the other operations use RGB.

\begin{algorithmbox}{Nearest + median cut}
\textit{Input:} image $I$ and target side length $N$.
\begin{enumerate}[leftmargin=1.6em,itemsep=1pt,topsep=3pt]
    \item For each output cell $(y,x)$, sample $I$ at the center of the
    corresponding source cell to obtain $Y[y,x]$.
    \item Apply $\mathrm{MedianCut}(Y,K=8)$ and return the quantized grid.
\end{enumerate}
\end{algorithmbox}

\begin{algorithmbox}{SSIM-optimal + median cut}
\textit{Input:} image $I$ of side length $H$ and target side length $N$.
\begin{enumerate}[leftmargin=1.6em,itemsep=1pt,topsep=3pt]
    \item Set $s=H/N$. For each output cell, compute the aligned
    $s\times s$ block mean $L[y,x]$ and squared-value mean $Q[y,x]$ for each
    RGB channel.
    \item For each $2\times2$ output neighborhood, compute its mean $\mu$,
    the variance $\sigma_L^2$ of $L$, and
    $\sigma_H^2=\operatorname{mean}(Q)-\mu^2$.
    \item Set $r=\sqrt{\sigma_H^2/\sigma_L^2}$ when
    $\sigma_L^2\geq\varepsilon$, and $r=0$ otherwise. Form the local proposal
    $\hat{Y}=\mu+r(L-\mu)$.
    \item Average overlapping proposals, clip values to $[0,1]$, and return
    $\mathrm{MedianCut}(\hat{Y},K=8)$.
\end{enumerate}
\end{algorithmbox}

\newpage
\begin{algorithmbox}{PIA (reduced)}
\textit{Input:} image $I$ and target side length $N$.
\begin{enumerate}[leftmargin=1.6em,itemsep=1pt,topsep=3pt]
    \item Compute $S=\mathrm{SLIC}(I;n=N^2,m=45)$ in CIELAB.
    \item For each superpixel $u$, compute its mean color $c_u$ and size
    $w_u=|u|$.
    \item Fit eight palette colors $\{p_k\}_{k=1}^{8}$ and assignments $a$
    using weighted $k$-means on $\{c_u\}$ with weights $\{w_u\}$.
    \item For each output cell, identify the superpixel at the corresponding
    source-cell center and assign the RGB conversion of $p_{a(u)}$.
\end{enumerate}
\end{algorithmbox}

\subsection{Complete Algorithmic Results}
\label{app:baseline-results}

\begin{table}[H]
\centering
\setpositionedtablecaption{\tablecaptionfont\textbf{All palette-matched algorithmic baselines on I2P.}
Every method is restricted to eight colors. DINO is paired DINOv2-Large CLS
cosine, SigLIP cosine uses pooled SigLIP-SO400M features, and 100-way Top-1
is measured over the 100 source images. Higher is better, with $n=100$ for every
cell. Bold values mark the best result in each column, and colored bands
distinguish the four method families.}
{tab:baselines-fair}
\small
\setlength{\tabcolsep}{4pt}
\resizebox{\linewidth}{!}{%
\begin{tabular}{lccccccccc}
\toprule
& \multicolumn{3}{c}{DINO CLS cosine}
& \multicolumn{3}{c}{SigLIP cosine}
& \multicolumn{3}{c}{100-way Top-1 (\%)} \\
\cmidrule(lr){2-4}\cmidrule(lr){5-7}\cmidrule(lr){8-10}
Method & $16^2$ & $32^2$ & $64^2$ & $16^2$ & $32^2$ & $64^2$
& $16^2$ & $32^2$ & $64^2$ \\
\midrule
\rowcolor{tierzerohead}
\multicolumn{10}{l}{\tiny\textbf{Tier 0\quad Simple resampling}} \\
\rowcolor{tierzerobody}
Nearest + median cut & .012 & .090 & .480 & .523 & .568 & .665 & 9 & 28 & 62 \\
\rowcolor{tierzerobody}
Box + median cut & .012 & .099 & .475 & .523 & .577 & .673 & 10 & 27 & 65 \\
\addlinespace[2pt]
\rowcolor{tieronehead}
\multicolumn{10}{l}{\tiny\textbf{Tier 1\quad Linear filters}} \\
\rowcolor{tieronebody}
Bicubic + median cut & .013 & .093 & .459 & .524 & .573 & .672 & 9 & 26 & 64 \\
\rowcolor{tieronebody}
Lanczos + median cut & .013 & .099 & .478 & .524 & .576 & .673 & 10 & 29 & 64 \\
\addlinespace[2pt]
\rowcolor{tiertwohead}
\multicolumn{10}{l}{\tiny\textbf{Tier 2\quad Content-adaptive downscaling}} \\
\rowcolor{tiertwobody}
DPID $\lambda=0.5$ + median cut & .011 & .103 & .509 & .523 & .578 & .679 & 10 & 27 & 68 \\
\rowcolor{tiertwobody}
DPID $\lambda=1.0$ + median cut & .011 & .104 & .527 & .522 & .580 & .678 & 9 & 32 & 65 \\
\rowcolor{tiertwobody}
SSIM-optimal + median cut & .013 & .119 & \textbf{.543} & .524 & .579 & .686 & 9 & 34 & 75 \\
\addlinespace[2pt]
\rowcolor{tierthreehead}
\multicolumn{10}{l}{\tiny\textbf{Tier 3\quad Pixel-art-specific methods}} \\
\rowcolor{tierthreebody}
Pyxelate & .013 & .085 & .377 & .525 & .569 & .655 & 12 & 32 & 65 \\
\rowcolor{tierthreebody}
PIA (reduced) & .014 & \textbf{.135} & .490 & \textbf{.546} & \textbf{.608} & \textbf{.697} & \textbf{18} & \textbf{50} & \textbf{78} \\
\rowcolor{tierthreebody}
PIA (reference) & \textbf{.015} & .123 & .485 & .541 & .599 & .692 & 17 & 40 & 71 \\
\bottomrule
\end{tabular}%
}
\placepositionedtablecaptionbottom
\end{table}

Table~\ref{tab:baselines-fair} shows that standard resampling filters and the
two DPID settings yield similar results, while the strongest method depends on
the metric. SSIM-optimal obtains the highest DINOv2 fidelity at $64^2$, whereas
PIA (reduced) leads both SigLIP metrics across resolutions. We therefore use
one representative for simple resampling, fidelity-oriented downscaling, and
joint geometry-palette optimization.

\end{document}